\documentclass[10pt,twocolumn,letterpaper]{article}

\usepackage{cvpr}              % To produce the CAMERA-READY version
\newcommand{\gradient}[1]{#1} % Placeholder definition for \gradient
\newcommand{\gradienttwo}[1]{#1} % Placeholder definition for \gradienttwo
\usepackage{graphicx}
\usepackage{amsmath}
\usepackage{amssymb}
\usepackage{booktabs}
\usepackage{multirow}
\usepackage{bbm}
\usepackage{amsmath}
\usepackage{enumitem}

\usepackage{pifont}
\usepackage{xcolor}
\usepackage{colortbl}
\usepackage{arydshln}
\usepackage{tikz}

\usepackage{makecell}

\usepackage{algorithm,algcompatible} % for algorithm table
\algnewcommand\INPUT{\item[\textbf{Input:}]}%  for algorithm table
\algnewcommand\OUTPUT{\item[\textbf{Output:}]}%for algorithm table

\definecolor{darkgreen}{rgb}{0.2, 0.7, 0.1}

\definecolor{Gray}{gray}{0.8}
\definecolor{LG}{gray}{.92}

\newcommand*\colourcheck[1]{%
  \expandafter\newcommand\csname #1check\endcsname{\textcolor{#1}{\ding{51}}}%
}
\colourcheck{green}
\colourcheck{darkgreen}

\usepackage{multirow}
\usepackage{adjustbox}
\usepackage{array}
\usepackage{wrapfig,lipsum,booktabs}
\usepackage{amsmath,amssymb}
\usepackage{enumitem}
\usetikzlibrary{decorations.pathmorphing}
\definecolor{darkgreen}{rgb}{0.2, 0.7, 0.1}
\newcommand*\colourxmark[1]{%
  \expandafter\newcommand\csname #1xmark\endcsname{\textcolor{#1}{\ding{55}}}%
}
\colourxmark{red}
\usepackage{arydshln}

\usepackage{stmaryrd}
\usepackage{trimclip}
\usepackage{stfloats} 

\makeatletter
\DeclareRobustCommand{\shortto}{%
  \mathrel{\mathpalette\short@to\relax}%
}

\newcommand{\short@to}[2]{%
  \clipbox{{.3\width} 0 0 0}{$\m@th#1\vphantom{+}{\shortrightarrow}$}%
  }
\makeatother

\definecolor{cvprblue}{rgb}{0.21,0.49,0.74}
\usepackage[pagebackref,breaklinks,colorlinks,allcolors=cvprblue]{hyperref}

\def\paperID{6479} % *** Enter the Paper ID here
\def\confName{CVPR}
\def\confYear{2025}

\title{Ev-3DOD: Pushing the Temporal Boundaries of 3D Object Detection \\ with Event Cameras}

\author{Hoonhee Cho$^{*}$, Jae-young Kang$^{*}$, Youngho Kim, and Kuk-Jin Yoon  \\
KAIST\\
{\tt\small \{gnsgnsgml,mickeykang,kmax2001,kjyoon\}@kaist.ac.kr}
}

\begin{document}

\twocolumn[{%
\renewcommand\twocolumn[1][]{#1}%
\maketitle
\begin{center}
    \vspace{-10pt}
    \centering
    \captionsetup{type=figure}
    \includegraphics[width=.98\textwidth]{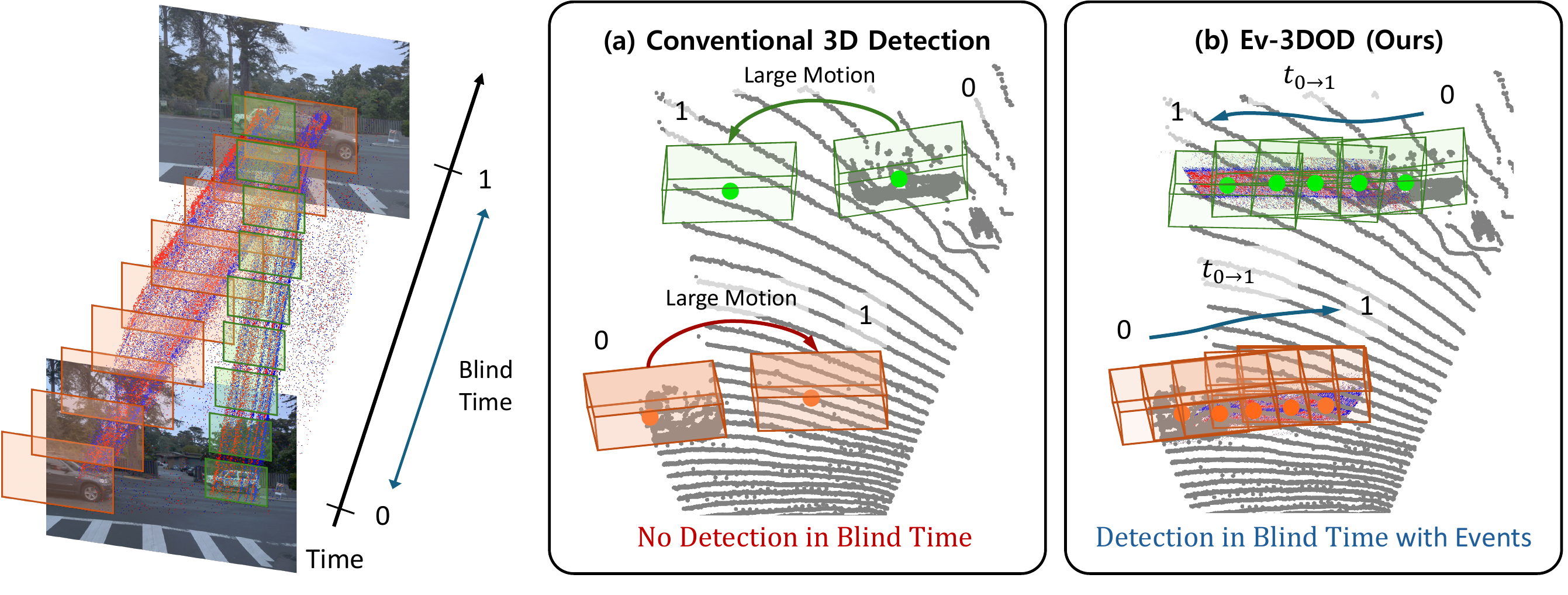}
    \vspace{-12pt}
    % \captionof{figure}{TBD. $t_0$, $t_0^0,t_0^1,t_0^2$
    % $T_0$, $T_0^0$, $T_0^1$, $\dots$, , $T_0^9$, $T_1$
    % }
    % % \vspace{5pt}
    \captionof{figure}{Comparison between the proposed Ev-3DOD and conventional 3D detection methods. Fixed frame rate sensors (\eg,~LiDAR, frame camera) inevitably have periods of ``blind time" ($t_{0\rightarrow1}$) when they lack data between active timestamps (0 and 1). Consequently, as shown in (a), conventional methods cannot detect objects during blind time, while other objects like vehicles can undergo significant motion within these gaps. In contrast, as illustrated in (b), our method, Ev-3DOD, leverages the high temporal resolution of the event camera, enabling accurate object detection even during the blind times of other sensors.}
    \label{fig:teaser}
\end{center}%
}]

\begin{abstract}
Detecting 3D objects in point clouds plays a crucial role in autonomous driving systems. Recently, advanced multi-modal methods incorporating camera information have achieved notable performance. For a safe and effective autonomous driving system, algorithms that excel not only in accuracy but also in speed and low latency are essential. However, existing algorithms fail to meet these requirements due to the latency and bandwidth limitations of fixed frame rate sensors, \textit{e.g.}, LiDAR and camera. To address this limitation, we introduce asynchronous event cameras into 3D object detection for the first time. We leverage their high temporal resolution and low bandwidth to enable high-speed 3D object detection. Our method enables detection even during inter-frame intervals when synchronized data is unavailable, by retrieving previous 3D information through the event camera. Furthermore, we introduce the first event-based 3D object detection dataset, DSEC-3DOD, which includes ground-truth 3D bounding boxes at 100 FPS, establishing the first benchmark for event-based 3D detectors. The code and dataset are available at \url{https://github.com/mickeykang16/Ev3DOD}.
\end{abstract}

\section{Introduction}
% \begin{figure*}[h!]
% \begin{center}
% \includegraphics[width=.9\linewidth]{figures/teaser_v1.pdf}
% % \vspace{-10pt}
% \caption{\textbf{TBD}}
% \label{fig:teaser}
% \end{center}
% % \vspace{-10pt}
% \end{figure*}
\def\thefootnote{*}\footnotetext{The first two authors contributed equally.}\def\thefootnote{\arabic{footnote}} 

\label{sec:intro}
3D object detection is a crucial task in autonomous driving. Multi-modal sensors like LiDAR and cameras are commonly used to understand scenes and handle complex road scenarios. This sensor setup consists of synchronized sensors with a fixed frame rate and has limited speed due to bandwidth limitations.
As shown in Fig.~\ref{fig:teaser}~(a), the blind time, which is the period when sensor information is unavailable (\eg,~100 milliseconds), is critical as cars can experience large motions as 3 meters during this time. Therefore, temporal resolution is a crucial factor that can directly influence accident risk.

% 센서의 정보가 부재한 시간인 blind 시간 (\eg, 100ms) 동안 차와 같은 object들은 3 meter 이상을 움직일 수 있기에, temporal resolution은 굉장히 중요한 문제이다.

% 최근에는 neuromorphic sensor인 이벤트 카메라를 이용하여 perceptual research에 새로운 페러다임을 제시하였다. 이벤트 카메라는 brightness changes를 stream 형태의 데이터로 제공하여, sub-milisecond latency를 가져서 motion blur~\cite{yang2024latency}에 강건하고, 낮은 bandwidth를 가진다~\cite{}.
% 이벤트 카메라는 bandwith와 blind time에 대한 trade-off를 극복한 대신, intensity에 대한 정보를 희생한다는 단점이 있으며, 최근 work들은 이러한 이벤트 카메라의 단점을 보완하기 위해 fixed  camera와 함께 사용하여 low-level vision과 autonomous driving에서 이벤트 카메라의 sparsity를 보완함과 동시에 센서의 temporal limit을 극복하였다.

Recently, event cameras~\cite{gallego2020event} have introduced a new paradigm in perceptual research. Event cameras provide brightness changes as a stream of data in sub-millisecond latency, which makes them robust to motion blur~\cite{yang2024latency} and have low bandwidth~\cite{Gehrig2022RecurrentVT, tulyakov2022time}. Event cameras overcome the trade-off between bandwidth and blind time but have the drawback of sacrificing intensity information. To address this, recent work has combined event cameras with frame cameras to complement the sparsity of event cameras while overcoming the fixed frame rate sensor's limits. 
% Apart from this line of research, event cameras are 2D image-based sensors, and as such, they are not inherently optimized for performing 3D perception on their own.
% 이러한 흐름과는 별개로 이벤트 카메라는 이미지 기반의 센서이기 때문에 그 자체만으로 3D perception을 수행하는데 최적이지는 않다. 

In this paper, we aim for accurate 3D object detection in blind time through a multi-modal approach that combines LiDAR and frame-based sensors with an event camera, leveraging its high temporal resolution. % Since LiDAR and cameras have a fixed frame rate, blind time inevitably occurs between the active timestamps, which refer to the moments when sensor data is acquired. 
% 이러한 세팅에서는 blind time 동안에는 이벤트 데이터만 존재하고, LiDAR와 image data는 존재하지 않으며, 가장 최근 active timestamp의 LiDAR와 이미지 데이터만 존재한다.
% 즉, 이미지와 LiDAR 데이터와 현재의 시점은 timestamp가 어긋나 있으며, 우리는 이벤트 카메라를 통해 현재 시점에 맞게 과거 정보를 옮겨주는 전략이 필요하다.
As shown in Fig.~\ref{fig:teaser}~(b), event data is available during blind times, while the LiDAR and image data from the most recent active timestamp are accessible. Therefore, there is a misalignment in timestamps between the fixed frame sensor data and the current blind time. Then, this raises the question: \textit{how can we estimate 3D motion using only event data, especially in cases where data from other sensors aligned with the current time is unavailable?}

To address this, we propose a virtual 3D event fusion (V3D-EF) that transfers past 3D information to align with the present time using the event camera. 
% 이벤트 카메라는 2D image plane 상에서 만의 pixel 단위의 움직임을 가지고 있기에, 3D motion을 추정하기 위해 우리는 3D voxel 공간에 이벤트를 project하여 가상의 3D event feature를 생성한다. 그 과정에서, point cloud가 존재하는 공간에 대해서만 event를 project하여 point cloud와 align하는 과정을 거친다. 다른 기존 연구에서 많이 다루었뜻이, 이벤트 데이터로 생성된 feature는 그 자체로 motion을 포함하지만, 우리의 연구에서는 이벤트로 3D motion을 추정해야한다.
As extensively covered in previous research~\cite{kim2024cross, shiba2022secrets, gehrig2024dense}, event data inherently contain motion information on an image plane. Based on this observation, we hypothesized that providing a 3D cue associated with the initial timestamp to the event information would enable accurate estimation of the 3D motion vector.
% Since the event camera captures only pixel-level motion on the 2D image plane, to estimate 3D motion, 
To this end, we project event features into a 3D voxel space to create virtual 3D event features. In this process, events are projected only onto regions where point cloud data exists, aligning with the point cloud data. 
% Then, we apply ROI pooling [9, 48] to generate the ROI-grid features.
Then, we aggregate the event features with the voxel features to estimate an implicit motion field that captures the motion inherent within the grid. The generated implicit motion field encapsulates both the 3D spatial information from voxels and the motion information from events, enabling the calculation of a 3D motion vector from the most recent active timestamp to the present blind time.

Another crucial factor in 3D object detection is the score for each bounding box. The model considers both box proposal scores at the active timestamp and the motion score of each predicted box in blind time.
In blind time, the model aligns 3D information of voxel features and motion information of events to learn the confidence of box motion.

% event camera의 3D perception 연구의 장벽(허들)은 데이터셋의 부재이며, 이를 해결하기 위해 우리는 이 work에서 두 가지 데이터셋을 제시한다 synthetic event data를 포함하는 Ev-Waymo와 real event data를 포함하는 DSEC-3DOD 데이터셋을. 
A major hurdle in 3D detection research using event cameras is the lack of suitable datasets. To address this, we present two datasets: Ev-Waymo, which includes synthetic event data, and DSEC-3DOD, which contains real event data. Unlike existing public 3D object detection datasets, ours includes annotations not only for active timestamps with 3D LiDAR sensor data but also for blind times, providing annotation in 100 FPS frame rate.

\section{Related Works}
\label{sec:related_work}

\noindent
\textbf{Single-modality 3D Object Detection.}
% Camera-based 3D object detection has seen significant developments, categorized into image-view-based~\cite{chen2017multi, ku2018joint, chen2016monocular, mousavian20173d, qin2019monogrnet} and bird-eye-view (BEV)-based approaches~\cite{}. 
% Recent advancements, including DETR3D and PETR, employ transformer models to convert 2D features into 3D space, with BEVDet and BEVDepth predicting depth distributions to map image features to a 3D frustum meshgrid~\cite{}. 
Given that cameras offer significant cost advantages over LiDAR sensors, many researchers have focused on developing methods that leverage camera-based systems for 3D object detection using image-only inputs~\cite{guo2021liga, chen2022pseudo, he2019mono3d++, liu2020smoke, lu2021geometry, you2019pseudo, wang2022detr3d, liu2022petr, li2023bevdepth, huang2021bevdet, jiang2023polarformer, li2019stereo, sun2020disp, chen2020dsgn, peng2020ida, wang2024bevspread, Liu_2024_CVPR}. 
% In image-based 3D object detection, the absence of direct depth information poses a challenge. To address this, several studies~\cite{philion2020lift, reading2021categorical, wang2019pseudo, you2019pseudo, park2021pseudo} have employed depth estimation techniques to create pseudo 3D point cloud representations or lift 2D features into the 3D space for subsequent object detection. 
% Recently, transformer-based architectures have been introduced to exploit 3D object queries and 3D-2D correspondences in the detection process~\cite{}. However, due to the inherent difficulty of accurately estimating depth from images, the performance of image-based approaches still lags behind that of LiDAR-based methods.
Image-based 3D object detection faces challenges due to the lack of direct depth information. To overcome this challenge, several studies~\cite{philion2020lift, reading2021categorical, wang2019pseudo, you2019pseudo, park2021pseudo, chen2020dsgn} have employed depth estimation techniques to generate pseudo 3D point clouds or elevate 2D features into 3D space for object detection, alongside proposals for transformer-based architectures~\cite{huang2022monodtr, zhang2023monodetr, wang2022detr3d} that utilize 3D object queries and 3D-2D correspondences. Nonetheless, accurately estimating depth from images remains difficult, resulting in image-based methods performing worse than LiDAR-based approaches.

% Despite these advancements, the information provided by a single modality remains constrained, impacting overall detection performance.

LiDAR-based 3D object detection can reliably estimate 3D bounding boxes using point clouds captured from LiDAR sensors. Current LiDAR-based detection methods can be divided into three categories based on different point cloud encoding formats: point-based methods~\cite{shi2020point, shi2019pointrcnn, qi2017pointnet, qi2017pointnet++, zhou2020joint, yin2022proposalcontrast, yang20203dssd, hu2022point, qi2019deep, feng2024interpretable3d}, voxel-based methods~\cite{ he2022voxel, lang2019pointpillars, li2024di, wang2023ssda3d, yin2022semi, ho2023diffusion, zhou2018voxelnet, chen2022mppnet, deng2021voxel, li2021lidar, mao2021pyramid, xu2022int, fan2022embracing, guan2022m3detr, he2022voxel, mao2021voxel, sheng2021improving, Zhang_2024_CVPR}, and point-voxel fusion networks~\cite{chen2019fast, miao2021pvgnet, shi2020pv, yang2019std, he2020structure}. While LiDAR detection is advantageous in various conditions, it often struggles in areas with sparse LiDAR data. Therefore, integrating the geometric benefits of point clouds with the semantic richness of images remains crucial for improving performance.

\begin{figure*}[t]
\begin{center}
\vspace{-2pt}
\includegraphics[width=.95\linewidth]{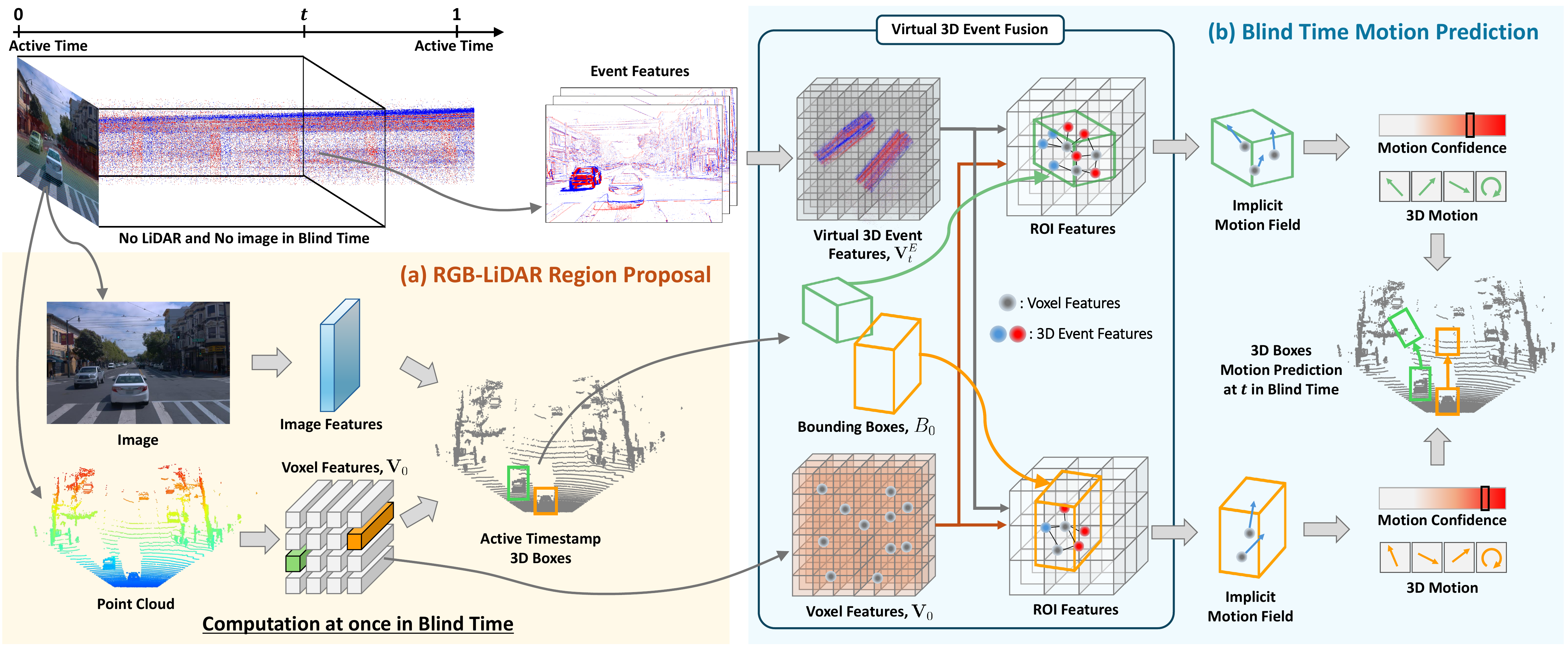}
\vspace{-7pt}
\caption{The overall pipeline for event-based 3D object detection (Ev-3DOD). (a): At active timestamp 0, LiDAR and image data are available. Therefore, we utilize an RGB-LiDAR Region Proposal Network (RPN) to extract voxel features and 3D bounding boxes at the active timestamp. (b): To predict the bounding box during blind time, $0\leq t<1$, we estimate the 3D motion and confidence score using event features. For computational efficiency, we design the process to compute (a) only once before the next active timestamp while performing iterative computations solely for (b).
}
\label{fig:overall_method}
\end{center}
\vspace{-10pt}
\end{figure*}
% For this purpose, we propose Virtual 3D Event Fusion (V3D-EF), which projects and fuses event data into 3D space.

\noindent
\textbf{Multi-modality 3D Object Detection.} Multi-modal sensor fusion~\cite{liang2019multi, liang2018deep, xu2018pointfusion, huang2021makes, li2023logonet, Chen2022FUTR3DAU, Xie2023SparseFusionFM, li2024gafusion, yin2024fusion, lin2024rcbevdet, huang2025detecting, song2024graphbev}, where different sensors complement each other, typically utilizes a combination of cameras and LiDAR. Multimodal 3D object detection has gained attention for improving accuracy over unimodal methods. Fusion approaches are categorized into early~\cite{chen2022deformable, vora2020pointpainting, xu2021fusionpainting, chen2017multi, ku2018joint, qi2018frustum}, middle~\cite{li2022unifying, li2022voxel, li2022deepfusion, liang2022bevfusion, liu2023bevfusion, piergiovanni20214d, yang2022deepinteraction, zhang2022cat, prakash2021multi}, and late fusions~\cite{pang2020clocs, bai2022transfusion, li2023logonet}, with middle fusion now preferred due to its robustness to calibration errors and enhanced feature interaction. Cross-modal fusion has achieved significant performance improvements by enhancing the sparse features of point clouds with semantic information from images. However, there are still unexplored areas concerning safety in autonomous driving. LiDAR and cameras have limited time resolution (\ie,~10-20 Hz) due to the relatively high bandwidths. To enhance safety, algorithms that operate at high speeds are essential. To address this, we introduce an asynchronous event camera with high temporal resolution for 3D object detection for the first time. The proposed method performs multi-modal fusion between synchronized sensors and asynchronous events to increase the operational frequency of 3D detection.

% cross-modal fusion은 point cloud의 부족한 sparse한 feature를 이미지의 semantic으로 부터 enhance할 수 있어 큰 성능 향상을 달성하였지만, 자율주행의 safety 측면에서 아직 덜 탐구된 영역이 있다. 라이다와 카메라는 상대적으로 높은 bandwidth로 time resolution이 높지 않으며, 안전성을 높이려면 고속도로 작동하는 알고리즘이 요구된다. 이를 위해, 우리는 처음으로 high temporal resolution을 가지는 asynchronous한 event camera를 3D object detection에 도입하여, 다른 센서와 multi-modal fusion을 하는 방식을 제안하여 모델 작동의 frequency를 향상한다.

\noindent
\textbf{Event-based Object Detection.} Event cameras break through the limitations of bandwidth requirements and perceptual latency with the trade-off of sparse data and loss of intensity information. To make the most of these characteristics, prior work has designed models to adapt to event data using graph neural networks~\cite{Schaefer22cvpr, Gehrig2024LowlatencyAV, Gehrig2022PushingTL, Bi2019GraphBasedOC, Li2021GraphbasedAE}, spiking neural networks~\cite{Yao2021TemporalwiseAS, Zhang2022SpikingTF}, and sparse neural networks~\cite{Peng2024SceneAS, Messikommer2020EventbasedAS}. Recently, effective dense neural networks~\cite{Li2022AsynchronousSM, Perot2020LearningTD, Iacono2018TowardsEO, Hamaguchi2023HierarchicalNM, Peng2023BetterAF, Wang2023DualMA, Zubic2023FromCC} and transformer architectures~\cite{Peng2024SceneAS, Gehrig2022RecurrentVT, Peng2023GETGE, Li2023SODFormerSO} have been developed to leverage event cameras, aiming to achieve both low latency and high performance. These advances enable real-time operation with impressive accuracy, demonstrating the practical applicability of event cameras in real-world scenarios. To the best of our knowledge, we are the first to extend event-based object detection from 2D to 3D space. By employing a multi-modal fusion approach of the event camera and 3D sensors (\ie,~LiDAR), we aim to overcome the latency and persistent interval time barriers of 3D object detections.

% 우리가 알기로, 우리는 처음으로 이러한 2D 상에서 event-based object detection을 3D space로 확장하고자 한다. 우리는 라이다와 같은 3D 센서와 함께 멀티 모달 퓨전 접근을 통해, 기존 3D object detection의 latency와 고질적인 interval time 장벽을 부수고자 한다.

\section{Method}

% \textbf{Probl}

\subsection{Problem Definition.}
% 우리의 goal은 LiDAR와 Image가 존재하는 active timestamp에 해당하는 0으로부터 다음 active timestamp인 1에 도달하기 전인 blind time $0 \leq t < 1$에서 3D bounding box를 추론하는 것이다.
LiDAR and camera data are only available at active timestamps 0 and 1, where 1 represents invalid future information in an online process. 
% Our goal is to predict 3D bounding boxes during \(0 \leq t < 1\), including blind time, when synchronized sensors are unavailable. 
Our goal is to predict 3D bounding boxes during \(0 \leq t < 1\), including the blind time when synchronized sensors are unavailable.
Given a pair of synchronized 3D points $P_0=\{(x_l, y_l, z_l)\}_{l=1}^N$ and image $I_0 \in \mathbb{R}^{3\times H \times W}$ at active timestamp 0, we leverage event stream $\mathcal{E}_{0 \rightarrow t}=\{(i, \ j, \ p, \ \tau) | 0 \leq \tau < t\}$ to detect 3D bounding box set $B_t$ at an arbitrary time $0 \leq t < 1$. 
% 본 연구는 fixed frame rate 라이다와 카메라 데이터가 부재한 interframe interval에서 aynchronous event를 이용하여 3d object를 detect 하는 것을 목적으로 한다. 우리의 모델은 

\subsection{Framework Overview}
As illustrated in (a) of Fig.~\ref{fig:overall_method}, at active timestamp 0, both LiDAR and image data are available. RGB-LiDAR Region Proposal Network (RPN) generates bounding box proposals from rich 3D information in active time. We begin by voxelizing the point cloud and generating voxel features, denoted as $\mathbf{V}_0 \in \mathbb{R}^{D_X \times D_Y \times D_Z \times C}$, where $D_X$, $D_Y$, and $D_Z$ represent the grid size, and $C$ denotes the channel dimension, using a 3D backbone~\cite{zhou2018voxelnet}. Then, using existing region proposals~\cite{yin2021center} and refinement networks~\cite{li2023logonet} with image features, we generate bounding box proposals, $B_0 = \{ B_0^1, B_0^2, \ldots, B_0^n \}$ and box proposal confidence $p_0=\{p^1_0, p^2_0, \ldots ,p^n_0\} \in [0, 1]$ for the active timestamp 0. RGB-LiDAR-based multi-modal networks have been widely studied, leading to many advanced models capable of accurate estimation at active timestamps. We tuned the model with fewer parameters and a reduced channel dimension to improve efficiency.

Our emphasis is on achieving precise estimation throughout the blind time, spanning the interval from active timestamp 0 to the subsequent timestamp 1. To infer bounding boxes at an arbitrary inter-frame time \( t \), we first convert the event stream, $\mathcal{E}_{0 \rightarrow t}$ into voxel grids~\cite{zhu2019unsupervised} as $\mathbb{E}_{t}$. Then, we use a feature encoder to produce event features, $\mathbf{E}_{t} \in \mathbb{R}^{H/4 \times W/4 \times C}$, where $H$ and $W$ is the spatial dimensions of the events. Our objective is to estimate the 3D motion of objects during blind time. To achieve this, we utilize the bounding box set \( B_0 \) and voxel features \( \mathbf{V}_0 \) from the active timestamp 0. During blind times, we share the voxel features, $\mathbf{V}_{0}$, generated at the active timestamp, ensuring that voxel features are computed only once to maintain a cost-effective structure. Through the Virtual 3D Event Fusion (V3D-EF), we generate an implicit motion field for each bounding box by integrating event and voxel features, encapsulating motion information. Using this field, we compute the 3D motion and confidence score, enabling the estimation of 3D boxes at the blind time \( t \).

% During blind times, we reuse the voxel features, $\mathbf{V}_{0}$, generated at the active timestamp to maintain a cost-effective structure.

% \subsection{Event-based Motion Prediction Module}

\subsection{Virtual 3D Event Fusion for Motion Prediction}
\label{Sec:v3d_ef}

% 이벤트 feature는 2D image plane 상에서 각 픽셀의 motion 정보를 시간에 대해 연속적으로 담고 있어, 2D motion을 추정하는데 굉장히 특화되어 있다. 
The event feature continuously captures each pixel's motion information over time on the 2D image plane. However, estimating 3D motion using only events is a challenge. We hypothesize that providing each event feature with adequate 3D information would allow for effective 3D motion estimation. To achieve this, we aim to incorporate a 3D cue into the event features. 
% To match the sparse event features with sparse voxel features, it is crucial to fully utilize the information from both modalities.
Since both event features and voxel features are sparse, it is crucial to align each voxels with the corresponding events accurately.
Directly using the voxel features would cause misalignment with the actual 3D geometry. Thus, we adopt the voxel point centroid method employed in previous works~\cite{hu2022point}.

% 우리는 sparse한 event feature를 sparse한 voxel feature와 matching을 해야하기 때문에, 두 modality의 information에서 나오는 정보를 적극적으로 활용해야 한다. 단순히, voxel feature를 그대로 사용하면, 실제 geometric한 coordinate와 다르기 때문에, 우리는 기존 work에서 사용한 voxel point centorid 방법을 채택한다.

% Inspired by grid-subsampling in KPConv~\cite{31, 32}, the voxel point centroid localization module spatially locates non-empty voxel features for aggregation in density-aware RoI grid pooling. 

\begin{figure}[t]
\begin{center}
\includegraphics[width=.98\linewidth]{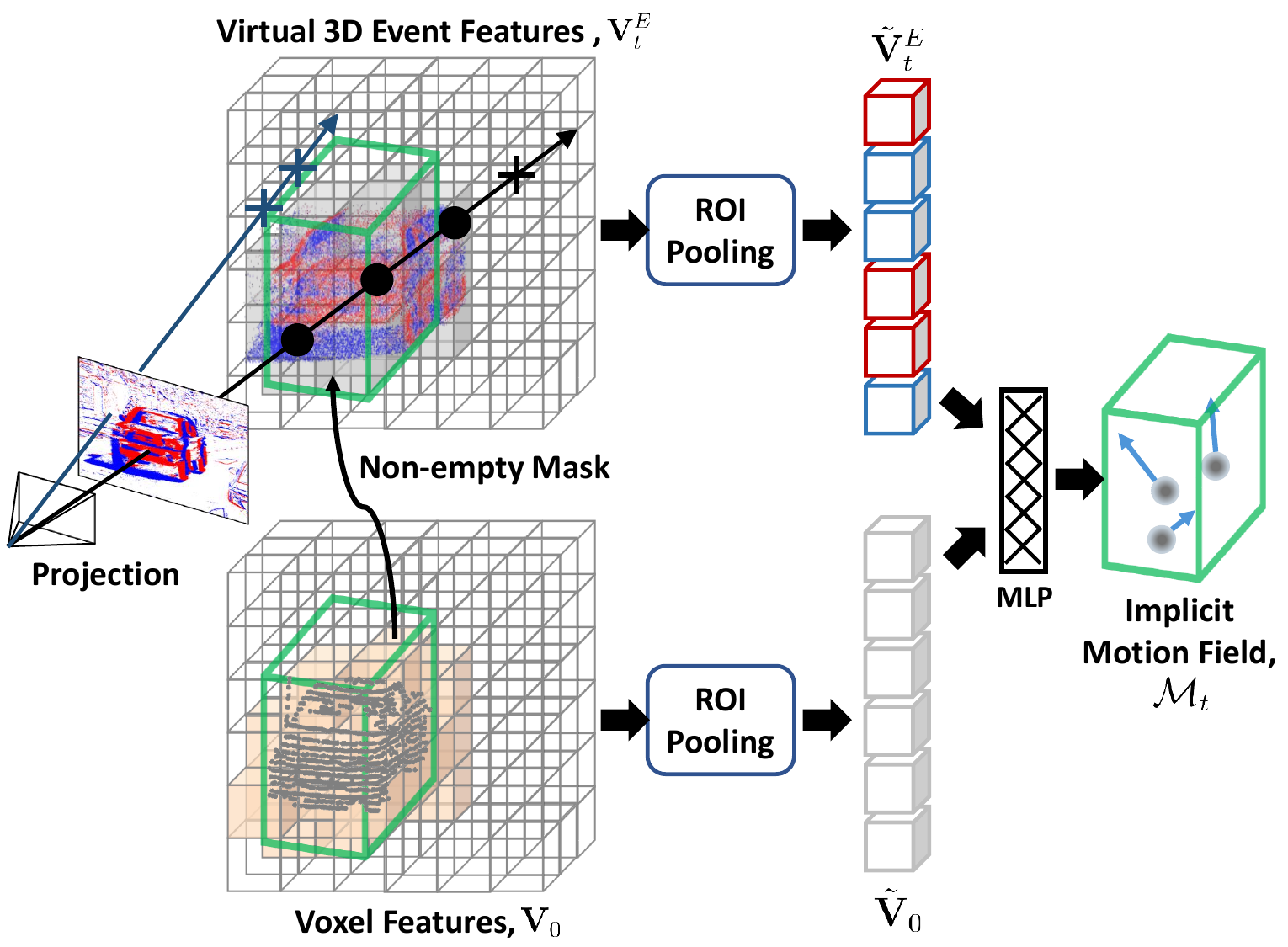}
\vspace{-2pt}
\caption{Virtual 3D Event Fusion (V3D-EF). We generate an implicit motion field for each bounding box by applying the ROI pooling  separately to the virtual 3D event features, which represent the 3D projection of events, and the voxel features.}
\label{fig:v3d_ef}
\end{center}
\vspace{-16pt}
\end{figure}

Specifically, we first consider the non-empty voxel features as $\mathbf{V}_0 = \{ V^k_0 = \{ h^{k}_0, f^{k}_0 \} \ | \ k = 1, \ldots, N_\mathcal{V} \}$, where $h_0^k$ is the 3D voxel index, $f_0^k$ is the associated voxel feature vector, and $N_\mathcal{V}$ is the number of non-empty voxels. Then, points within each voxel are grouped into a set \( \mathcal{N}(V_0^k) \) by determining their voxel index \( h_0^k \) based on their spatial coordinates \( X_i=(x_i, y_i, z_i) \) of points. The centroid of each voxel feature is subsequently computed as:
\begin{equation}
\begin{aligned}
c_0^k = \frac{1}{|\mathcal{N}(V_0^k)|} \sum_{X_i \in \mathcal{N}(V_0^k)} X_i.
\label{equ:cost}
\end{aligned}
\end{equation}
The centroid \( c_0^k \) is calculated for non-empty voxels, ensuring it corresponds only to indices with existing points, with the goal of aligning these centroids to the event data. As shown in the top of Fig.~\ref{fig:v3d_ef}, we propose a method to project events into a virtual voxel space based on the coordinates of the centroids. The projection onto the image plane from 3D space can be obtained as:
$\mathbf{p}_0^k = K \cdot E \cdot c_0^k,$
% \begin{equation}
% \begin{aligned}
% \label{equ:project}
% \end{aligned}
% \end{equation}
where $K$ and $E$ denote the intrinsic and extrinsic parameters, $\cdot$ operation is matrix multiplication.
Therefore, we calculate the event coordinates corresponding to each centroid, $c_0^k$, and gather the event features, $\mathbf{E}_{t}$, at these coordinates, $\mathbf{p}_0^k$, to create virtual 3D event features, $\mathbf{V}_t^E = \{\mathbf{E}_t(\mathbf{p}_0^k) \}_{k=1}^{N_\mathcal{V}} \in \mathbb{R}^{N_\mathcal{V} \times C}$.

% 이러한 centroid는 non-empty voxel에 대해 고려하였으므로, 모두 point가 존재하는 index에 해당하며, 우리는 해당 centroid에 대하여 이벤트와 matching하고자 한다.

% 우리는 centroid의 coordinate를 기반으로 가상의 voxel 공간에 이벤트를 프로젝션하는 방식을 제안한다.

% 우리는 active timestamp 0에 추정된 bounding box에 대한 3D motion을 추정하고자 하기 때문에, box proposal에 대해서만 feature를 fusion 한다.

% ROI-grid feature를 생성한다.
Since we aim to estimate the 3D motion of bounding boxes from the active timestamp 0 up to the current time \( t \), we consider fusion only for the features corresponding to the box proposal. Given the bounding box proposals, $B_0$, at active timestamp 0, we divide box proposals into $S \times S \times S$ regular sub-voxels. 
% Then, we perform the ROI pooling on 
% Then, 우리는 각 sub-voxel에 해당하는 virtual 3D event features와 voxel features에 대해 ROI pooling~\cite{} operation을 각각 적용하여, bounding box들에 해당하는 feature를 생성한다.
% 그리고 생성한 feature들을 concatenate하고, MLP를 거쳐 implicit motion field를 생성해낸다.
% Then, we apply the ROI pooling~\cite{deng2021voxel} operation separately to the virtual 3D event features, $\mathbf{V}_t^E$, and voxel features, $\mathbf{V}_0$, corresponding to each sub-voxel, generating features for the bounding boxes as $\tilde{\mathbf{V}}_t^E, \tilde{\mathbf{V}}_0 \in \mathbb{R}^{n \times S^3 \times C}$, respectively, where $n$ is the number of box proposals.
Then, we independently perform the ROI pooling~\cite{deng2021voxel} on the virtual 3D event features $\mathbf{V}_t^E$, and the voxel features $\mathbf{V}_0$, corresponding to each sub-voxel. This generates grid features for the bounding boxes, denoted as $\tilde{\mathbf{V}}_t^E \in \mathbb{R}^{n \times S^3 \times C} $ and $ \tilde{\mathbf{V}}_0 \in \mathbb{R}^{n \times S^3 \times C}$, where $n$ is the number of box proposals. The generated features are then concatenated and passed through an MLP to produce the implicit motion field, $\mathcal{M}_t=\{\mathcal{M}_t^0,\mathcal{M}_t^1,\dots,\mathcal{M}_t^n\}$. The implicit motion field is created by aligning 3D events with their corresponding voxel features, encapsulating motion information for the bounding box. Using an MLP, we estimate the explicit 3D motion. The explicit box motion $\mathbf{M}_{0 \rightarrow t}^i=(dx^i, dy^i, dz^i, d\alpha^i)$, defined in box local coordinate of $B_0^i$, contains spatial shift and yaw rotation. The final box location is calculated as $B_t^i= g(B_0^i, \mathbf{M}_{0 \rightarrow t}^i)$, where $g$ is a box shifting operation.

% $ 

% 우리는 non-empty voxel에 대한 

% Since the voxels in the convolution layers are in a sparse format, an intermediate hash table is used to efficiently map each calculated voxel point centroid to its corresponding feature vector. As shown in Figure~3, both voxel point centroids and sparse voxel features are associated with a shared voxel index. The intermediate hash table links the centroid \( c^{V^l}_k \) with \( V^l_k \) using the matching voxel index \( h^{V^l}_k \).

% centroid~\cite{hu2022point}

%EMP module

% ray, virtual, voxel 이런게 넣고 싶긴 한 욕구가 있음.

% \subsection{Dynamic Region Proposal Module}
% new object handling 해주는 모듈

% \subsection{Confidence Refinement}

% \subsection{Persistence Scoring Module}
% removal이 강조되면 좋을듯

\subsection{Motion Confidence Estimator}
\label{sec:mce}
% Our framework estimates the motion of box proposals predicted at the active timestamp 0 throughout the blind time. 
As discussed in previous studies~\cite{zhou2018voxelnet, yin2021center}, self-awareness of the confidence of predicted bounding boxes is a key factor. The simplest choice of final confidence score in our setting would be the score of active timestamp 0. However, the final score should take into account the object's motion prediction during the blind time. If the object experiences challenging motion during the blind time, a reliable bounding box at the active timestamp 0  could become less reliable. Therefore, the final score should consider two different sources. The first aspect is the score estimated at the active timestamp, based on LiDAR and images. The second aspect is the confidence score based on motion during the blind time using events.

% Sec.~\ref{sec:}에서 다룬, implicit motion field는 point cloud와 event 사이에 align을 통해, 3D motion 정보를 담고 있어, 우리는 confidence만을 위한 MLP를 추가로 사용하여 motion confidence를 계산한다.

The implicit motion field from Sec.~\ref{Sec:v3d_ef} contains 3D motion information through alignment between the point cloud and events. We additionally employ an MLP specifically for calculating motion confidence. Box-wise motion confidence is estimated and normalized, resulting in $p_{0 \rightarrow t}=\{p^1_{0 \rightarrow t}, p^2_{0 \rightarrow t}, \ldots ,p^n_{0 \rightarrow t}\} \in [0, 1]$. The final confidence score for box proposal $i$ is calculated by considering both the score at the active timestamp and the motion confidence:
\begin{equation}
\begin{aligned}
p_t^i=p_0^i \cdot p_{0 \rightarrow t}^i.
% \label{equ:final_score}
\end{aligned}
\end{equation}

\subsection{Training Objectives}
% \noindent
% \textbf{Motion Estimation.}

For RGB-LiDAR RPN in the active timestamp, we follow RPN loss from ~\cite{Yan2018SECONDSE, li2023logonet, yin2021center}.
For blind time prediction, box regression loss from \cite{yin2021center} is applied between the predicted 3D box $B_t^i$ and the ground-truth $\hat{B}_t^i$. 
% \noindent
% \textbf{Motion Confidence.}
The target assigned to the motion confidence branch is an IoU-related value, as:
\[
\hat{p}^i_{0 \rightarrow t}(\text{IoU}_i) = 
\begin{cases} 
0 & \text{if IoU}_i < \theta_L, \\
\frac{\text{IoU}_i - \theta_L}{\theta_H - \theta_L} & \text{if } \theta_L \leq \text{IoU}_i < \theta_H, \\
1 & \text{if IoU}_i \geq \theta_H,
\end{cases}
\]
 $\text{IoU}_i$ is the IoU between the $i$-th prediction and the corresponding ground truth box, while $\theta_H$ and $\theta_L$ are foreground and background IoU thresholds. Binary Cross Entropy Loss is exploited here for confidence prediction. The losses of our detect head are computed as:

 \begin{equation}
\begin{aligned}
\mathcal{L}= \mathcal{L}_{RPN}
+\lambda_1 \cdot \mathcal{L}_{reg}
+\lambda_2 \cdot \mathcal{L}_{score}
\end{aligned}
\end{equation}

\section{Event-based 3D Object Detection Datasets}

This study is the first to explore the utilize event camera for 3D detection. To validate our model's effectiveness, we introduce two event 3D detection datasets: Ev-Waymo, which consists of synthetic events, and DSEC-3DOD with real events. We first obtained accurate 3D bounding box annotations for the active time, when both LiDAR and image data are available, and then processed the blind time.

%본 연구는 event camera를 3d detection에 사용하는 첫 번째 시도임. 따라서 우리는 두 개의 multi-modality event 3d detection dataset을 introduce 하여, 우리 모델의 effectiveness를 증명함. Interframe interval에서의 3d detection을 실험하기 위해, 라이다와 이미지가 부재한 시간에서의 3D box annotation을 제작하였다. 

\subsection{Annotation for Active Time at 10 FPS}

% \subsection{Event Waymo Open Dataset}
\noindent
\textbf{Ev-Waymo Dataset} is a synthetic event dataset based on the Waymo Open Dataset (WOD)~\cite{Sun2019ScalabilityIP}. WOD provides sequential pairs of synchronized LiDAR, image data, and 3D box annotations at 10 FPS. Event streams are generated using a widely-adopted event simulator~\cite{Gehrig_2020_CVPR}.
% We added annotations at blind time divided evenly into 10 points between consecutive frames, resulting in temporally dense annotations at 100 FPS. 
% Ev-Waymo evaluation uses the same metric as WOD. Average precision weighted by heading (APH) and average precision (AP) are calculated for 3 classes: vehicle, pedestrian, and cyclist. All 100 FPS GT boxes are used for evaluation. Also, objects are categorized into two difficulty levels depending on the number of LiDAR points inside the bounding box.
The Ev-Waymo evaluation follows the same metrics as WOD, calculating both average precision (AP) and average precision weighted by heading (APH).
% for three classes: vehicle, pedestrian, and cyclist. 
% Additionally, objects are classified into two difficulty levels (Level 1 (L1), Level 2 (L2)) based on the number of LiDAR points within each bounding box.
% Furthermore, objects are divided into two difficulty levels, Level 1 (L1) and Level 2 (L2), determined by the number of LiDAR points within each bounding box.
% DSEC setting과 동일하게 하기 위해, 우리는 front camera와 해당 field-of-view의 LiDAR를 crop하여 사용한다.
% To align with the DSEC-3DOD setting, Ev-waymo uses the front camera and crop the LiDAR to the corresponding field of view.
% To conform to the DSEC-3DOD setting, Ev-Waymo utilizes the front camera and crops the LiDAR to match the corresponding field of view.
Adhering to the DSEC-3DOD setting, Ev-Waymo employs the front camera and a cropped LiDAR aligned to the corresponding field of view.

% Event Waymo Open Dataset (Ev-WOD)는 Waymo Open Dataset 기반의 synthetic event 3d detection datset임. WOD는 10HZ의 싱크로나이즈된 라이다와 이미지, 그리고 3D box ground truth를 제공함. 우리는 연속된 프레임 사이 interval을 10개로 균일하게 분할한 시점에서의 annotation을 추가하였으며, 결과적으로 100fps의 temporally dense한 high quality annotaion을 제공함. 

%라이다와 이미지가 없는 시점에서 정확한 bounding box ground truth를 제공하기 위헤, state-of-the art video frame interpolation과 pointcloud interpolation 모델을 사용하여 interval time에서의 pseudo sensor data를 생성하였음. 10fps의 bounding box를 interpolation한 후, data annotation expert를 고용하여 pseudo sensor data를 기반으로 bounding box를 refine하여 높은 퀄리티의 100fps event 3d detection dataset을 제작함. Ev-Waymo dataset은 총 16k개의 labeled scene으로 이루어진 80개의 sequence로 이루어져있음.

\begin{figure}[t]
\begin{center}
\includegraphics[width=.99\linewidth]{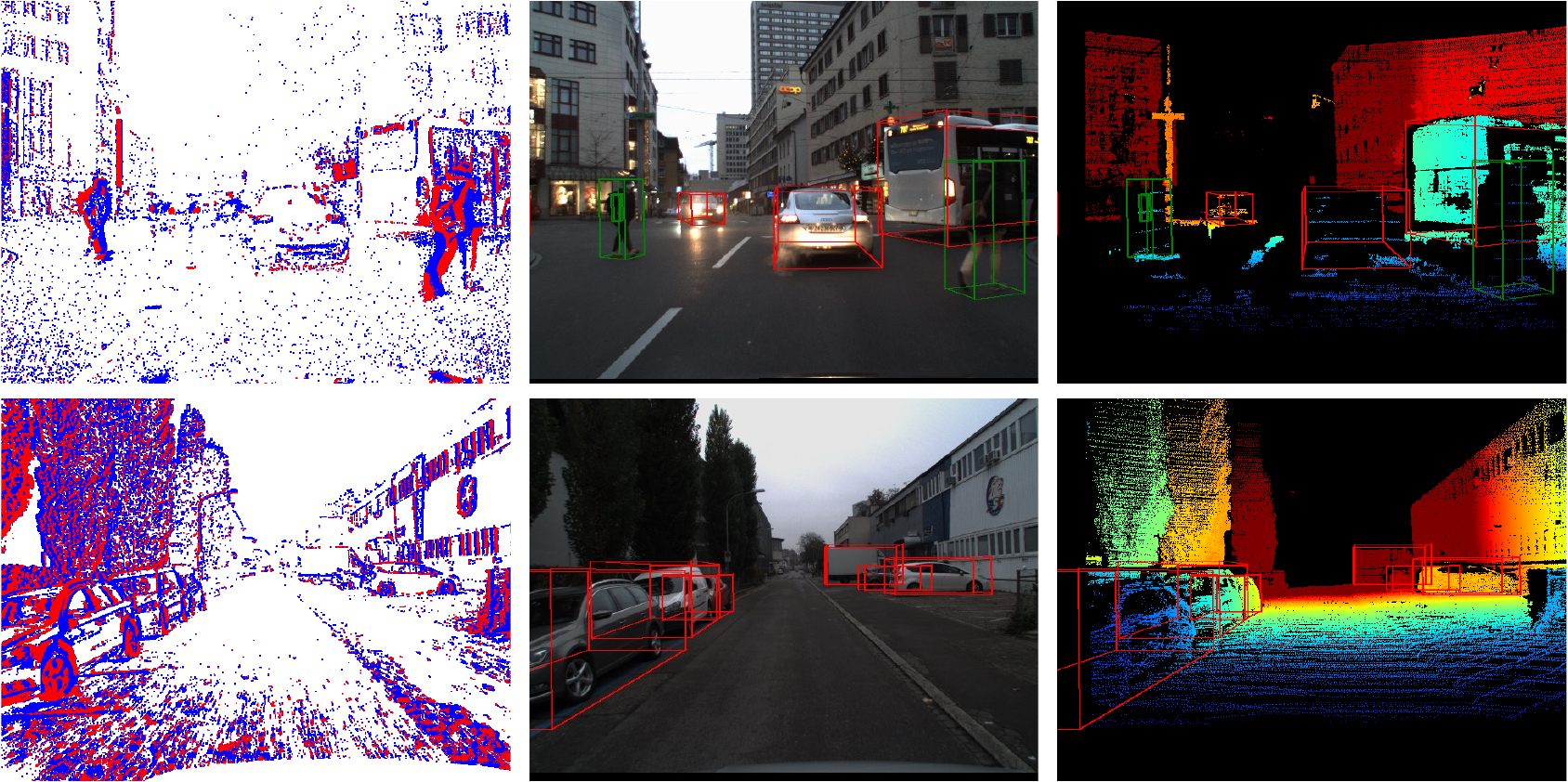}
\vspace{-6pt}
\caption{DSEC 3D Object Detection Dataset samples. Event data, label overlaid image, and accumulated LiDAR from left to right.}
\label{fig:data_sample}
\end{center}
\end{figure}

\setlength{\aboverulesep}{-1.5pt}
\setlength{\belowrulesep}{0pt}
\setlength{\tabcolsep}{2.4pt}
\renewcommand{\arraystretch}{1.1}
\begin{table*}[t]
\begin{center}
\caption{Performance comparison on the Ev-Waymo dataset for 3D vehicle (IoU = 0.7), pedestrian (IoU = 0.5) and cyclist (IoU = 0.5) detection following previous protocols. Evaluated on 100 FPS ground-truth annotation for $t=\{0, 0.1, \ldots, 0.9 \}$. Offline represents results obtained by interpolating data using sensor data from active timestamps 0 and 1. Online represents results obtained using data accessible at present. Specifically, online evaluation of previous works uses only the data from active timestamp 0, while we also utilize event data from active timestamp 0 up to the present time.}
\vspace{-7pt}
\label{tab:main_waymo}
\resizebox{.99\linewidth}{!}{
\begin{tabular}{c|c|c|c|c|cc|cc|cc}
\toprule
% \hline
% \hline
% & & \multicolumn{14}{c}{Target Sequences} \\
% \cline { 3 - 16 }
& \multirow{3}{*}{\thead{Interpolation\\Methods}} & \multirow{3}{*}{Methods} & \multirow{3}{*}{\thead{3D Detection\\Modality}} &  ALL & 
\multicolumn{2}{c|}{\multirow{2}{*}{VEH (AP/APH)}} & \multicolumn{2}{c|}{\multirow{2}{*}{PED (AP/APH)}} & \multicolumn{2}{c}{\multirow{2}{*}{CYC (AP/APH)}} \\
% \hline
&  &  &  &  (mAP/mAPH) & & & & & & \\
\cline{5-11}
 & &  &  & L2 & L1 & L2 & L1 & L2 & L1 & L2 \\
\hline
\multirow{6}{*}{Offline} & \multirow{2}{*}{N-PCI~\cite{zheng2023neuralpci}} & 
VoxelNeXt~\cite{chen2023voxelnext} & L & 53.61/50.67 & 62.96/62.51 & 59.10/58.67 & 68.43/60.77 & 62.85/55.69 & 44.55/43.15 & 38.88/37.65 \\
& & HEDNet~\cite{zhang2024hednet} & L & 50.52/47.51 & 59.57/59.09 &	55.90/55.45 & 63.28/55.32 & 58.03/50.62 & 43.14/41.80 & 37.63/36.46 \\
\cline{2-11}
& \multirow{2}{*}{N-PCI~\cite{zheng2023neuralpci} + EMA~\cite{zhang2023extracting}} & Focals Conv~\cite{chen2022focal} & L + I & 42.53/40.41 & 55.49/55.02 & 54.45/53.99 & 48.07/43.06 & 46.54/41.69 & 27.70/26.60 & 26.59/25.54 \\
& & LoGoNet~\cite{li2023logonet} & L + I & 53.29/50.49 & 63.48/63.05 & 60.15/59.73 & 67.85/60.28 & 62.43/55.37 & 42.40/41.37 & 37.29/36.38 \\
\cline{2-11}
& \multirow{2}{*}{N-PCI~\cite{zheng2023neuralpci} + CBM~\cite{kim2023event}} & Focals Conv~\cite{chen2022focal} & L + I & 42.51/40.39 & 55.48/55.01 & 54.44/53.98 & 48.07/43.06 & 46.54/41.68 & 27.67/26.57 &	26.56/25.50 \\
& & LoGoNet~\cite{li2023logonet} & L + I & 53.57/50.75 & 63.60/63.16 & 60.27/59.85 &	68.14/60.53 & 62.77/55.67 &	42.85/41.81 & 37.66/36.74 \\
\noalign{\vskip -0.6pt}
\hline
\hline
\noalign{\vskip 1pt}
\rowcolor{LG}
 &  & VoxelNeXt~\cite{chen2023voxelnext} & L & \underline{33.32}/31.70 & \underline{44.40}/\underline{44.10} & \underline{41.78}/\underline{41.49} &	\underline{40.52}/\underline{36.23} & \underline{36.93}/\underline{32.96} &	24.40/23.73 & 21.24/20.66 \\
 \rowcolor{LG}
 &  & HEDNet~\cite{zhang2024hednet} & L & 31.57/29.90 & 42.03/41.71 & 39.32/39.02 &	38.86/34.43 & 35.67/31.53 &	22.64/21.99 & 19.72/19.14 \\
\rowcolor{LG}
& & Focals Conv~\cite{chen2022focal} & L + I & 26.27/25.01 & 37.31/37.01 & 36.60/36.31 &	29.20/26.16 & 28.41/25.44 &	14.30/13.78 & 13.79/13.29  \\
\rowcolor{LG}
& & LoGoNet~\cite{li2023logonet} & L + I  & 33.27/\underline{31.75} & 44.14/43.87 & 41.73/41.47 &	39.98/35.84 & 36.48/32.67 &	\underline{24.71}/\underline{24.15} & \underline{21.59}/\underline{21.10}\\
\rowcolor{LG}
\multirow{-5}{*}{Online}& \multirow{-5}{*}{N/A} & Ev-3DOD (Ours) &  L + I + E &\textbf{48.06}/\textbf{45.60} & \textbf{60.30}/\textbf{59.95} & \textbf{59.19}/\textbf{58.85} & \textbf{57.40}/\textbf{50.78} & \textbf{55.30}/\textbf{48.93} & \textbf{31.08}/\textbf{30.38} & \textbf{29.69}/\textbf{29.03} \\
\bottomrule
\end{tabular}
}
\end{center}
\vspace{-15pt}
\end{table*}

% \subsection{DSEC 3D Object Detection Dataset}
\noindent
\textbf{DSEC-3DOD.} 
The DSEC dataset~\cite{gehrig2021dsec} is a stereo event dataset that includes LiDAR, image, and real event data captured in challenging environments. We sampled a portion of the ``zurich\_city" sequence, providing  LiDAR and image data at 10 FPS. We hired annotation experts to label 3D bounding boxes at 10 FPS, ensuring alignment with the sensor data.
% From this dataset, we extracted LiDAR and image data at 10 FPS and hired annotation experts to label 10 FPS 3D bounding boxes aligned with the sensor data. 
% Ground-truth 3D bounding boxes were generated based on the accumulated LiDAR and image data.
% The ground-truth 3D bounding boxes were generated using accumulated LiDAR and image data.
% To achieve 100 FPS annotation, we interpolated the 10 FPS annotations to 100 FPS and refined them based on the results of event frame interpolation~\cite{Kim2023EventbasedVF} and point cloud interpolation~\cite{zheng2023neuralpci}. The DSEC 3D dataset comprises 178 scenes and includes 54K labeled scenes.
As shown in Fig.~\ref{fig:data_sample}, in the annotation process, we used accumulated LiDAR points to enable the precise generation of ground-truth 3D bounding boxes.
% DSEC 3D detection shares evaluation metrics with the Ev-Waymo dataset. However, it contains vehicle and pedestrian classes as the cyclist objects are insufficient. Also, all objects are considered as a single difficulty level. 
For evaluation on DSEC-3DOD, we used the same evaluation metrics as Ev-Waymo but include only vehicle and pedestrian classes due to a lack of cyclist data, with all objects categorized under a single difficulty level.

% DSEC dataset은 challeging한 환경에서 취득한 라이다, 이미지, 그리고 real 이벤트 데이터를 포함하고 있는 stereo event dataset임. "zurich_city" sequence 중 일부를 sampling하여 사용함. 우리는 데이터셋에서 10fps의 라이다와 이미지 데이터를 추출한 후, annotation 전문가를 고용하여 센서와 align된 10fps 3D bounding box를 annotation함. Accumulated LiDAR와 이미지 데이터를 기반으로 3d bounding box ground truth를 생성함.

%100fps annoatation을 위해서 10fps annotation을 100fps로 interpolation한 후, event frame interpolation과 pointcloud interpolation 결과를 기반으로 refine함. DSEC 3D dataset은 178개의 scene과 54k개의 label된 scene을 포함하고 있음.

\begin{figure*}[t]
\begin{center}
\includegraphics[width=.99\linewidth]{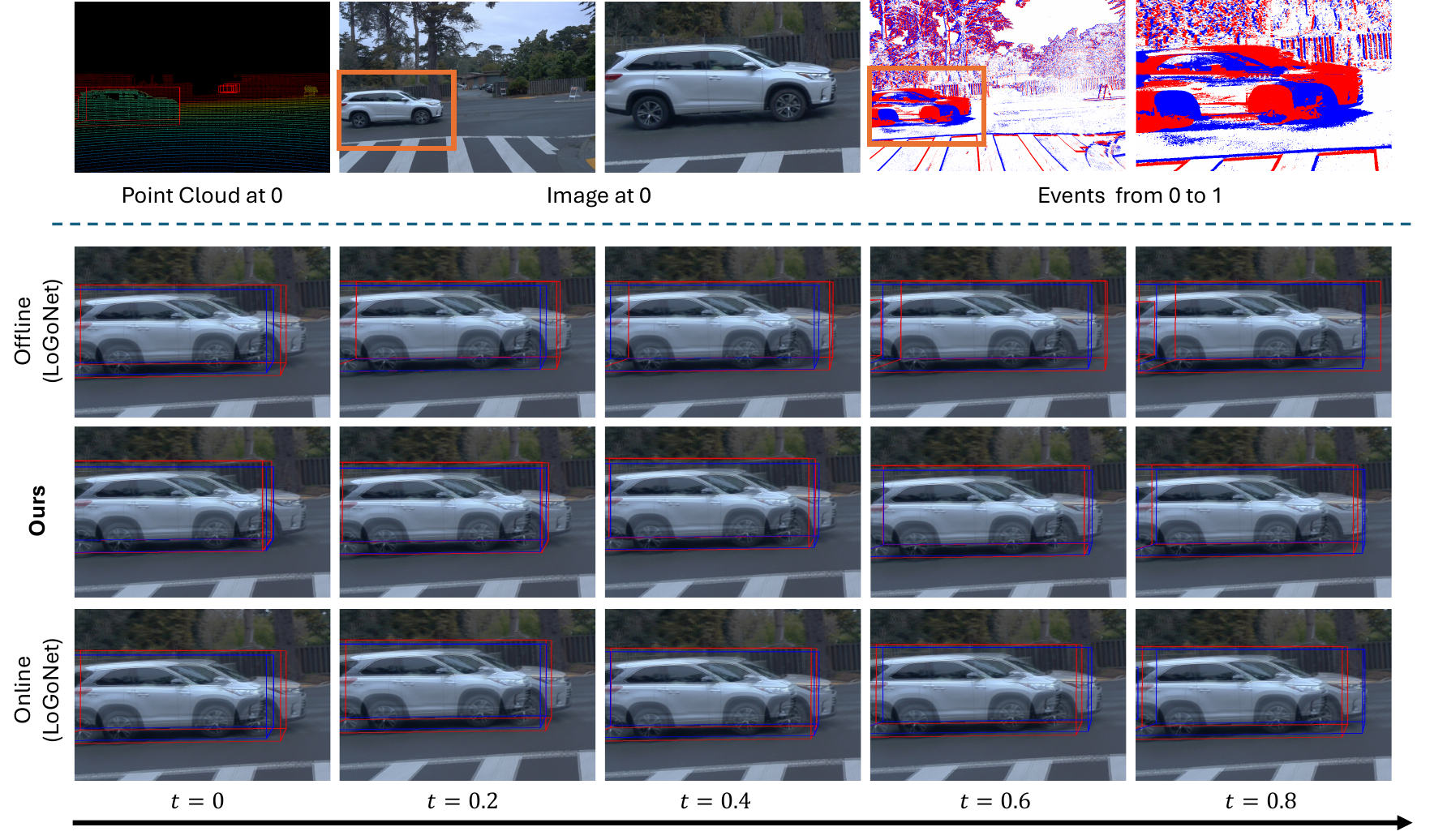}
\vspace{-5pt}
\caption{Qualitative comparisons of our method with other offline and online evaluations on the Ev-Waymo dataset. $t=0$ represents the active time, while $t = 0.2, 0.4, 0.6, 0.8$ denotes the blind times. The \textcolor{blue}{blue} box represents the ground truth, while the \textcolor{red}{red} box shows the prediction results of each method. For easier understanding, images at active timestamps 0 and 1 are overlaid.}
\label{fig:waymo_qual}
\end{center}
\vspace{-13pt}
\end{figure*}

% \subsection{Generation of Annotations for Blind Time}
\subsection{Annotation for Blind Time at 100 FPS}

For training and evaluation on 3D detection during blind time, we created 3D box annotations for periods when LiDAR and images were absent.
% We added annotations at blind times evenly divided into 10 intervals between consecutive frames, resulting in temporally dense annotations at 100 FPS.
% For 100 FPS annotation, we interpolated the original 10 FPS annotations to 100 FPS
% We linearly interpolated annotations at blind times, evenly dividing them into 10 intervals between consecutive frames, resulting in temporally dense annotations at 100 FPS. 
We linearly interpolated the original 10 FPS annotations, dividing blind times into 10 evenly spaced intervals between consecutive frames, resulting in temporally dense annotations at 100 FPS. To ensure accurate bounding box ground-truths in instances where LiDAR and image data are unavailable, we employed event-based video frame interpolation~\cite{kim2023event} and point cloud interpolation~\cite{zheng2023neuralpci} methods to generate sensor data for blind time. Data annotation experts refined interpolated bounding boxes based on the generated sensor data. The Ev-Waymo dataset consists of 80 sequences comprising a total of 157K labeled scenes, providing 100 FPS detection ground-truth. The DSEC 3D dataset consists of 178 sequences and includes 54K labeled scenes. For evaluation in blind time, we utilize all ground truth boxes at 100 FPS. More details of the datasets are described in the supplementary material.

\begin{table*}[t]
\setlength{\tabcolsep}{12.0pt}
\renewcommand{\arraystretch}{1.0}
\begin{center}
\caption{Results on the DSEC-3DOD dataset for 3D vehicle (IoU = 0.7), and pedestrian (IoU = 0.5) detection.}
\vspace{-8pt}
\label{tab:main_dsec}
\resizebox{.99\linewidth}{!}{
\begin{tabular}{c|c|c|c|cc|cc|cc}
\toprule
% \hline
% \hline
% & & \multicolumn{14}{c}{Target Sequences} \\
% \cline { 3 - 16 }
& Interpolation & \multirow{2}{*}{Methods} & 3D Detection &  \multicolumn{2}{c|}{ALL} & \multicolumn{2}{c|}{VEH} & \multicolumn{2}{c}{PED} \\
\cline{5-10}
 & Methods &  & Modality & mAP & mAPH & AP & APH & AP & APH \\
\hline
\multirow{6}{*}{Offline} & \multirow{2}{*}{N-PCI~\cite{zheng2023neuralpci}} & 
VoxelNeXt~\cite{chen2023voxelnext} & L & 36.29&	32.43&	45.35&	44.93&	27.23&	19.93\\
& & HEDNet~\cite{zhang2024hednet} & L &37.73&	30.73&	41.82&	41.35&	33.64&	20.1\\
\cline{2-10}
& \multirow{2}{*}{N-PCI~\cite{zheng2023neuralpci} + EMA~\cite{zhang2023extracting}} & Focals Conv~\cite{chen2022focal} & L + I & 30.29&	25.85&	35.52&	35.04&	25.05&	16.65\\
& & LoGoNet~\cite{li2023logonet} & L + I & 37.30&	31.39&	44.64&	44.06&	29.96&	18.72\\
\cline{2-10}
& \multirow{2}{*}{N-PCI~\cite{zheng2023neuralpci} + CBM~\cite{kim2023event}} & Focals Conv~\cite{chen2022focal} & L + I & 30.28&	25.86&	35.56&	35.08&	24.99&	16.64\\
& & LoGoNet~\cite{li2023logonet} & L + I &37.35&	31.45&	44.76&	44.17&	29.94&	18.73 \\
\noalign{\vskip -0.6pt}
\hline
\hline
\noalign{\vskip 1pt}
\rowcolor{LG}
 &  & VoxelNeXt~\cite{chen2023voxelnext} & L & \underline{13.94}&	 \underline{12.90} &   \underline{21.42} &  \underline{21.23} &	6.46&	4.57 \\
 \rowcolor{LG}
 &  & HEDNet~\cite{zhang2024hednet} & L & 13.75&	11.68&	18.48&	18.29&	\underline{9.02} &	 \underline{5.07} \\
\rowcolor{LG}
& & Focals Conv~\cite{chen2022focal} & L + I & 10.81&	9.61&	15.74&	15.55&	5.88&	3.67 \\
\rowcolor{LG}
& & LoGoNet~\cite{li2023logonet} & L + I  & 13.29&	11.91&	19.50&	19.32&	7.07&	4.49\\
\rowcolor{LG}
\multirow{-5}{*}{Online}& \multirow{-5}{*}{N/A} & Ev-3DOD (Ours) &  L + I + E & \textbf{31.17}  & \textbf{26.54} & \textbf{41.65} & \textbf{41.20} & \textbf{20.68} & \textbf{11.88}\\
\bottomrule
\end{tabular}
}
\end{center}
\vspace{-14pt}
\end{table*}

\begin{figure*}[t]
\begin{center}
\includegraphics[width=.92\linewidth]{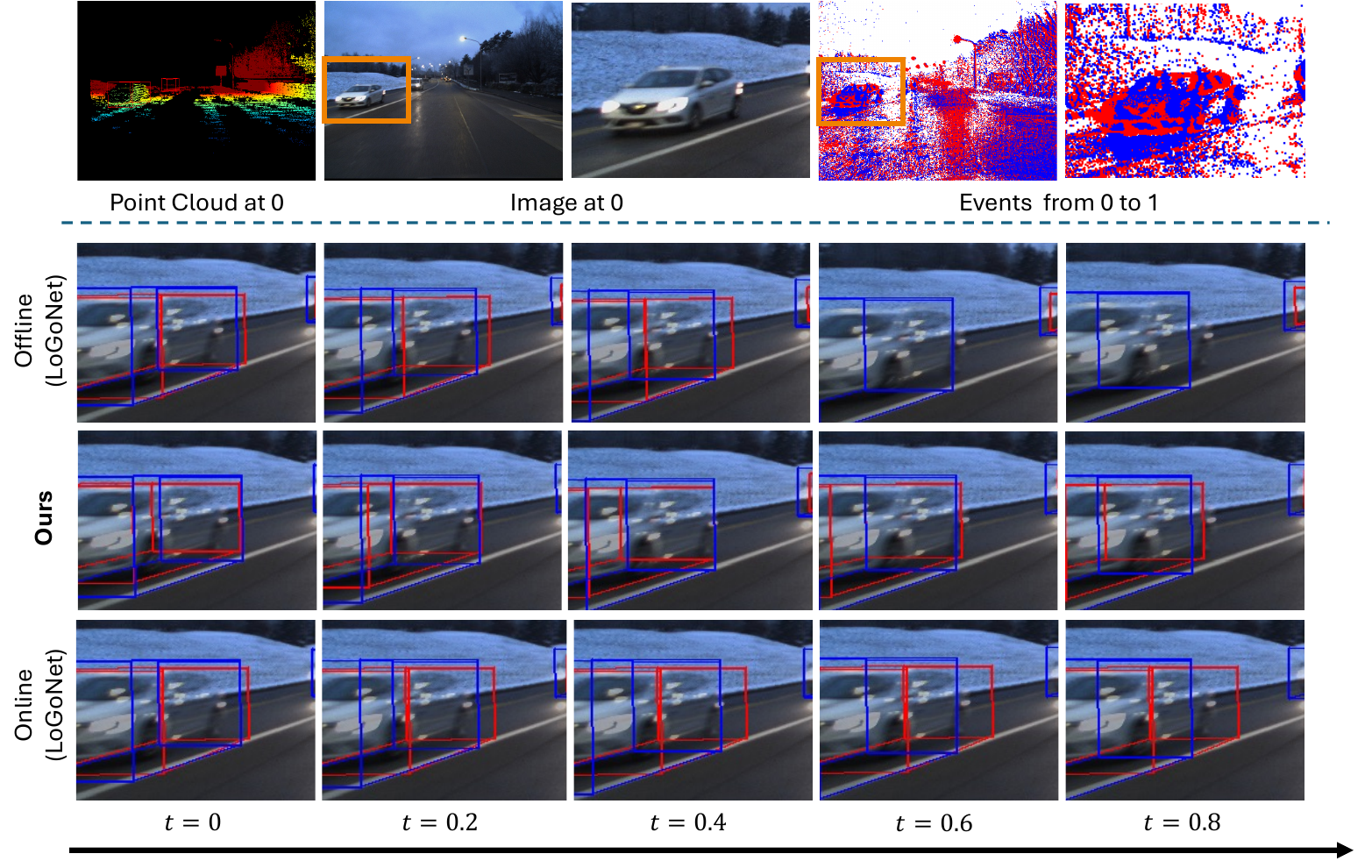}
\vspace{-11pt}
\caption{Qualitative comparisons of our method with other offline and online evaluations on the DSEC-3DOD dataset. $t=0$ represents the active time, while $t = 0.2, 0.4, 0.6, 0.8$ denotes the blind times. The \textcolor{blue}{blue} box represents the ground truth, while the \textcolor{red}{red} box shows the prediction results of each method. For easier understanding, images at active timestamps 0 and 1 are overlaid.}
\label{fig:dsec_qual}
\end{center}
\vspace{-19pt}
\end{figure*}

\section{Experiments}

% \subsection{Datasets}
% \noindent
% \textbf{Ev-Waymo.}

% \noindent
% \textbf{DSEC-3DOD.}

\subsection{Experiment Setup} % Setup?

% \noindent
% \textbf{Evaluation Setup.}

We target causal 3D detections to detect objects during blind time.
% Although direct comparison is not feasible, 
For comparison, we present two types of evaluation methods for conventional models. First is the online evaluation, where the network is inferred at any given blind time using only accessible data. Conventional methods can only process the most recent sensor data at active timestamp 0. In contrast, our approach additionally incorporates event data from active timestamp 0 up to the current blind time.
% to this end, 우리는 두 종류의 평가 방식을 present 한다. 첫번째로, online result로, straight forward하게 임의의 blind 시점에서 accesible한 데이터를 이용하여 네트워크가 inference하여 평가하는 것이다. 기존 방식의 경우, 가장 최근의 active timestmap 0의 센서 데이터만 사용이 가능하고, 우리의 방식은 추가적으로 active timestamp 0에서 현재 blind 시점까지의 이벤트 데이터도 함께 사용이 가능하다. 
% 다른 하나는, 기존 방식의 경우 fixed frame rate를 가지고 있어, direct하게 apply하여 blind time에 적용하면 어려움을 겪기 때문에, future data를 함께 사용하여 추론 및 평가하는 방식이다. In other words, data for the blind time is interpolated using active time stamps 0 and 1, which is the future data.
Second, offline evaluation addresses the challenge of applying conventional fixed-frame-rate approaches directly to blind time by utilizing future data. In other words, data for the blind time is interpolated using active timestamps 0 and 1, where timestamp 1 represents future data. Consequently, these offline evaluations serve as a performance oracle for online methods. More implementation details are provided in the supplementary material.

% Existing methods based on fixed frame rate sensors face difficulties in directly applying detection to these blind times. 

% To this end, we present offline evaluation results using future data as a performance reference. 

% For offline results, 

% 우리는 라이다와 이미지가 존재하지 않은 blind time에서 event data를 이용하여 3D detection이 가능함. 지금까지의 fixed frame rate sensor를 이용하는 방법론들은 blind time detection을 직접적으로 적용하기에 어려움이 있음. 비록 직접적으로 비교해야 하는 결과는 아니지만, 성능적인 참고를 위해 future 데이타를 사용한 offline evaluation 결과를 제시함. Offline 결과는 미래의 데이터를 이용하여 blind time의 pseudo data를 생성한 후 평가함. 따라서 offline evaluation들이 online 방법론의 성능적인 oracle임.

% \noindent
% \textbf{Implementation Details}

% Batch size 4.
% $S=6$ in ROI pooling
% event voxel grid bin size = 5

\subsection{Experimental Results}
% 우리는 기존 방법들과 비교하기 위해 최근 state-of-the-art 방법들인, camera-based와 camera-LiDAR 기반의 multi-modal approach들을 사용한다. 
We employ recent state-of-the-art approaches, including LiDAR-based~\cite{chen2023voxelnext, zhang2024hednet} and LiDAR-camera-based multi-modal methods~\cite{chen2022focal, li2023logonet}. For the offline image interpolation, we utilize frame-based, EMA~\cite{zhang2023extracting}, and the event-based method, CBM~\cite{kim2023event}.  NeuralPCI~\cite{zheng2023neuralpci} is adopted for point cloud interpolation.

\noindent
\textbf{Ev-Waymo.}
Table~\ref{tab:main_waymo} shows the quantitative result of Ev-Waymo. 
% Previous methods cannot detect new objects during blind time, resulting in reduced performance when evaluated with 100 FPS ground truth as they fail to capture scene changes. 
Comparing methods in the online setting, previous methods~\cite{li2023logonet, chen2022focal, zhang2024hednet, chen2023voxelnext}, which rely on fixed-frame-rate sensors (\ie,~LiDAR and Image), cannot detect object motion during blind times. This limitation leads to reduced performance when evaluated with 100 FPS ground-truth. In contrast, our approach, leverages high-temporal resolution of event data to estimate object motion during blind time, significantly outperforming existing online methods. Notably, Ev-3DOD is comparable to offline oracles and even surpasses the performance of a certain approach~\cite{chen2022focal}.

% Additionally, our method demonstrates comparable performance even when compared to offline oracle methods that use sensors interpolated with future data.

Figure~\ref{fig:waymo_qual} shows comparisons between the results of online, offline, and our method during blind time. Existing online methods fail to align prediction with the ground truth during blind time. In contrast, Ev-3DOD, despite being an online approach, aligns well with the ground-truth and demonstrates performance comparable to offline methods.

% The offline method, by interpolating sensor data, accurately predicts the actual object location during blind time. In contrast, e

% 이전 방법론들은 blind time에서 새로운 object를 detection 하는 것이 불가능하다. 따라서 100 FPS ground truth에서 평가를 했을 때 scene의 변화를 반영하지 못해 성능이 떨어진다. 반면, 우리 방법론은 이벤트 데이터를 이용하여 blind time에서 object motion을 추정하고 3D detection이 가능하며, 기존 online 방법론들을 큰 폭으로 outperform했다. 또한 미래의 데이터로 interpolation된 sensor를 사용한 offline oracle 방법론들과 비교했을 때도 comparable한 성능인 것을 확인할 수 있다.
% 아래 그림은 blind time에서 online method, offline method, 그리고 ours를 비교한 결과이다. 이해의 편의를 위해 0번과 1번 이미지를 겹쳐 표현하였다. 파란 박스는 ground truth, 빨간 박스는 prediction 결과이다. Offline 방법은 sensor data를 interpolation 하여 blind time에서 실제 물체 위치를 잘 prediction 하였다. 반면 기존 online 방법론은 active timestamp의 데이터만을 사용하여 blind time에서의 detection은 grount truth와 align 되어있지 않다. 

\noindent
\textbf{DSEC-3DOD.}
As shown in Table~\ref{tab:main_dsec}, we evaluate our model on the DSEC-3DOD real event dataset, which includes fast motion and challenging lighting conditions that significantly degrade performance during blind time. Our model effectively estimates the 3D motion of objects from events, achieving a substantial improvement over other online methods. Notably, it demonstrates performance comparable to offline oracle methods that utilize future information. The analysis of computational complexity is provided in the supplementary material.

Figure~\ref{fig:dsec_qual} shows blind time detection inference for fast-moving objects in the DSEC dataset. Under low-light conditions and rapid motion, the quality of the interpolated pseudo-sensor data may deteriorate, occasionally resulting in detection failures, even when future information is utilized. In contrast, the proposed Ev-3DOD accurately detects objects in blind time, predicting motion from events.

\begin{table}[t]
\begin{center}
\caption{Ablation study of the proposed components.}
% MCE denotes the motion confidence estimator (Sec.~\ref{sec:mce}).}
\label{tab:able_components}
\vspace{-6pt}
\renewcommand{\tabcolsep}{12pt}
\resizebox{.72\linewidth}{!}{
\begin{tabular}{cc|cc}
\Xhline{2.5\arrayrulewidth}
\multirow{2}{*}{V3D-EF} & \multirow{2}{*}{MCE} & \multicolumn{2}{c}{L2} \\
& & mAP & mAPH \\
\hline 
 & & 33.30 & 31.68 \\
\darkgreencheck & & \underline{46.55} & \underline{44.22} \\
% &  \darkgreencheck &  &  \\
\darkgreencheck & \darkgreencheck & \textbf{48.06} & \textbf{45.60} \\
% \bottomrule
% \hline
% \hline
\Xhline{2.5\arrayrulewidth}
\end{tabular}
}
\end{center}
\vspace{-13pt}
\end{table}

\begin{table}[t]
\begin{center}
\caption{Effectiveness of non-empty mask in the V3D-EF module.}
\label{tab:able_mask}
\vspace{-8pt}
\resizebox{.72\linewidth}{!}
{
\renewcommand{\tabcolsep}{10pt}
\begin{tabular}{lcc}
\Xhline{2.5\arrayrulewidth}
\multirow{2}{*}{Method} & \multicolumn{2}{c}{L2} \\
& mAP & mAPH \\
\hline
w/o Non-empty Mask & 42.57 & 40.50\\
w/ Non-empty Mask & \textbf{46.55} & \textbf{44.22}\\
\Xhline{2.5\arrayrulewidth}
\end{tabular}
}
\end{center}
\vspace{-18pt}
\end{table}

\subsection{Ablation Study and Analysis}
We present an ablation study on key components and various analyses on the Ev-Waymo dataset. In V3D-EF analyses, the motion confidence estimator (MCE) module is excluded for clear comparisons.

\noindent
\textbf{Effectiveness of Each Component.}
In Table~\ref{tab:able_components}, we conduct the ablation study on the key components of Ev-3DOD. We evaluated a baseline and progressively added modules to observe improvements. The motion confidence estimator is dependent on V3D-EF, so it was not evaluated separately. V3D-EF, which explicitly estimates motion, has a direct impact on performance, resulting in significant improvements of +13.25\% in mAP and +12.54\% in mAPH. Additionally, the motion confidence estimator refines prediction scores at each blind time, demonstrating its effectiveness with further gains of +1.51\% and +1.38\%, respectively.

\noindent
\textbf{Non-empty Mask in V3D-EF.}
In the proposed V3D-EF, we employ a non-empty mask to achieve accurate and efficient alignment between sparse events and voxel features. To demonstrate the effectiveness of this non-empty mask, we conduct a comparative experiment in which events are projected onto all voxels within the V3D-EF module. The results in Table~\ref{tab:able_mask} confirm that the non-empty mask enables effective alignment between the two modalities.

\noindent
\textbf{Variations of the V3D-EF module.}
Our core component, the V3D-EF module, offers several design choices. To analyze these, we designed various variants, presenting the results accordingly. Our core component, the V3D-EF module, provides multiple design options. To analyze these configurations, we developed and evaluated several variants, with the results summarized in Table~\ref{tab:able_variants}. We experimented with generating the implicit motion field using each modality, \ie,~LiDAR and event sensors, and found a performance drop compared to the multi-modal approach. This decline is likely due to the limitations of each modality: LiDAR alone cannot capture motion information during blind times, and event features cannot accurately estimate refined 3D motion. Additionally, multi-modal fusion before pooling decreased performance. We believe this is because fusing before pooling leads to integration between voxels and events outside of box proposal regions, resulting in the use of redundant and possibly misaligned information.

% Although we generated an implicit motion field based on a multi-modal approach using both LiDAR and event data, we conducted experiments by varying to a single modality.

\begin{figure}[t]
\begin{center}
\includegraphics[width=.96\linewidth]{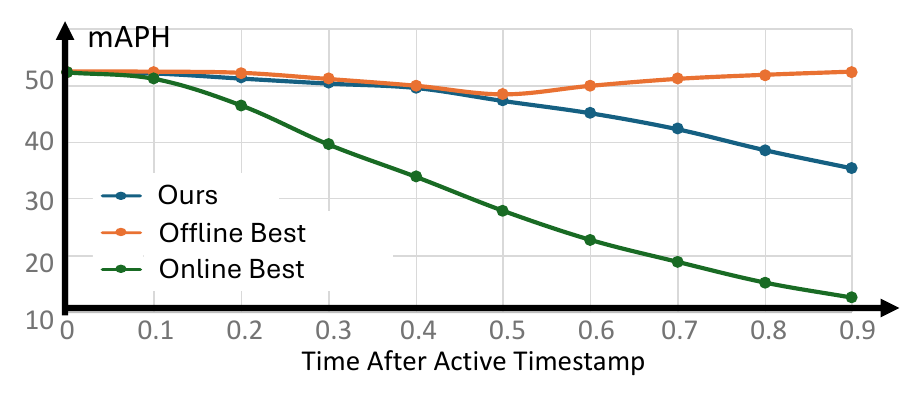}
\vspace{-10pt}
\caption{Detection performance over time elapsed since the active timestamp. We compare ours with offline and online approaches.}
\label{fig:frame_ablation}
\end{center}
\vspace{-20pt}
\end{figure}

\noindent
\textbf{Detection Performance per Elapsed Blind Time.} Figure~\ref{fig:frame_ablation} shows the detection performance over time elapsed since the active timestamp. Due to object motion, the performance of online detection deteriorates rapidly as the time distance from the active timestamp increases. 
In contrast, offline methods sustain robust performance throughout the blind time by leveraging future data, with only a slight performance drop at the center of the blind time as it moves further from the active timestamps located at both ends.
% , attributable to the limitations of interpolation. 
The proposed model, relying solely on event data, achieves a temporal degradation pattern comparable to that of the offline method.

% Object motion에 의해서 active timestamp에서 시간이 지날수록 online conventional detection 성능이 급격히 떨어짐. Offline은 future data를 사용해서 모든 blind time에 거쳐 좋은 성능을 유지했으며, interpolation의 nature에 의해 blind time의 중심에서 성능이 가장 낮음. Proposed model은 future data 없이도 event data만을 사용하여 offline method와 비슷한 정도의 성능 degradation이 있음을 확인함. 

\begin{table}[t]
\begin{center}
\caption{Comparison of results across different modalities and fusion variants of the V3D-EF module.}
\label{tab:able_variants}
\vspace{-8pt}
\setlength{\tabcolsep}{8.6pt}
\resizebox{.95\linewidth}{!}
{
\begin{tabular}{lcc}
\Xhline{2.5\arrayrulewidth}
\multirow{2}{*}{Method} & \multicolumn{2}{c}{L2} \\
& mAP & mAPH \\
\hline
Base Model & 33.30 & 31.68 \\
+ LiDAR & 37.04 & 35.25 \\
+ Event & 42.23 & 40.17 \\
+ LiDAR + Event (fusion before pooling) & \underline{43.95} & \underline{41.87} \\
+ LiDAR + Event (fusion after pooling) & \textbf{46.55} & \textbf{44.22}\\
\Xhline{2.5\arrayrulewidth}
\end{tabular}
}
\end{center}
\vspace{-18pt}
\end{table}

\section{Conclusions}
This paper proposes Ev-3DOD, a novel method to utilize an event camera in 3D detection for detecting objects in blind time. To transfer information from active times to the current blind time, we estimate 3D motion based on event data. To effectively fuse sparse LiDAR data with events, we propose a Virtual 3D Event Fusion (V3D-EF) and introduce a motion confidence estimator to define confidence for 3D detection during blind times. For this study, we introduce the first event-based 3D detection dataset, emphasizing its 100 FPS annotations. We hope this work showcases the potential of neuromorphic cameras, inspiring future research.

\clearpage
\setcounter{page}{1}
\maketitlesupplementary
\setlength{\aboverulesep}{-1.5pt}
\setlength{\belowrulesep}{0pt}
\setlength{\tabcolsep}{3.0pt}
\renewcommand{\arraystretch}{1.2}
\begin{table*}[!b]
\begin{center}
\caption{Performance and runtime comparison on the Ev-Waymo dataset. Evaluated at 100 FPS for $t = {0, 0.1, \ldots, 0.9}$. Offline results, which rely on sensor data from timestamp 1, future information, and additional interpolation algorithms, are excluded from evaluation.}
\vspace{-7pt}
\label{tab:inference_time}
\resizebox{.99\linewidth}{!}{
\begin{tabular}{c|c|c|cc|cc|cc|c}
\toprule
\multirow{3}{*}{Methods} & \multirow{3}{*}{\thead{3D Detection\\Modality}} &  ALL & \multicolumn{2}{c|}{\multirow{2}{*}{VEH (AP/APH)}} & \multicolumn{2}{c|}{\multirow{2}{*}{PED (AP/APH)}} & \multicolumn{2}{c|}{\multirow{2}{*}{CYC (AP/APH)}} & \multirow{3}{*}{FPS}\\
&  & (mAP/mAPH) & & & & & & & \\
\cline{3-9}
 & & L2 & L1 & L2 & L1 & L2 & L1 & L2 \\
\hline 
\noalign{\vskip 1pt}
VoxelNeXt~\cite{chen2023voxelnext} & L & 33.32/31.70 & 44.40/44.10 & 41.78/41.49 &	40.52/36.23 & 36.93/32.96 &	24.40/23.73 & 21.24/20.66 & 17.34\\
HEDNet~\cite{zhang2024hednet} & L & 31.57/29.90 & 42.03/41.71 & 39.32/39.02 &	38.86/34.43 & 35.67/31.53 &	22.64/21.99 & 19.72/19.14 & 12.84\\
Focals Conv~\cite{chen2022focal} & L + I & 26.27/25.01 & 37.31/37.01 & 36.60/36.31 &	29.20/26.16 & 28.41/25.44 &	14.30/13.78 & 13.79/13.29  & 6.08\\
LoGoNet~\cite{li2023logonet} & L + I  & 33.27/31.75 & 44.14/43.87 & 41.73/41.47 &	39.98/35.84 & 36.48/32.67 &	24.71/24.15 & 21.59/21.10 & 10.68\\
\hline
Ev-3DOD (Ours) &  L + I + E &\textbf{48.06}/\textbf{45.60} & \textbf{60.30}/\textbf{59.95} & \textbf{59.19}/\textbf{58.85} & \textbf{57.40}/\textbf{50.78} & \textbf{55.30}/\textbf{48.93} & \textbf{31.08}/\textbf{30.38} & \textbf{29.69}/\textbf{29.03} & \underline{27.09}\\
Ev-3DOD-\textit{Small} (Ours) &  L + I + E & \underline{44.21}/\underline{42.01} & \underline{57.95}/\underline{57.62} & \underline{56.89}/\underline{56.57} & \underline{51.87}/\underline{45.91} &	\underline{49.94}/\underline{44.21}	& \underline{27.01}/\underline{26.44} &	\underline{25.80}/\underline{25.25} & \textbf{54.14}
\\
\bottomrule
\end{tabular}
}
\end{center}
\vspace{-15pt}
\end{table*}

In this supplementary material, we offer more details of our work, Ev-3DOD. Specifically, we provide
\begin{itemize}
\item Implementation details in Section~\ref{sec:implementation};
\item Runtime analysis of our framework, Ev-3DOD, in Section~\ref{sec:inf_time};
\item Detailed processing and annotation processes about event-based 3D detection datasets in Section~\ref{sec:dataset};
\item Additional qualitative results and video demo in Section~\ref{sec:qual_video};
\item Hyper-parameter analysis in Section~\ref{sec:hyper};
\end{itemize}
\vspace{10pt}

\section{Implementation Details}
\label{sec:implementation}
The model is trained using a two-stage strategy inspired by previous works~\cite{li2023logonet, Yin2024ISFusionIC} that leverages pre-trained encoders. In the first stage, 10 FPS LiDAR and images are utilized to train the active timestamp RGB-LiDAR Region Proposal Network. In the second stage, all sensor data are incorporated to train the blind time modules with 100 FPS ground truth bounding boxes. In the first stage, the Region Proposal Network is trained for 15 epochs with a batch size of 4 and a learning rate of $0.001$, using the Adam optimizer~\cite{kingma2014adam} with the scheduling strategy from \cite{smith2019super}. In the second stage, the blind time modules are trained for 15 epochs with a batch size of 3, maintaining the same learning rate of $0.001$. The loss function incorporates weights $\lambda_1 = 1.0$ and $\lambda_2 = 1.0$.

Since only the front camera is used, we followed the KITTI~\cite{Geiger2012AreWR} methodology, utilizing only LiDAR point clouds and ground truth that fall within the camera's field of view. The point cloud spans $[0.0, 75.2m]$ along the $X$ axis, $[-75.2m, 75.2m]$ along the $Y$ axis, and $[-2m, 4m]$ along the $Z$ axis, with a voxel size of $(0.1m, 0.1m, 0.15m)$.
Ev-Waymo uses a resolution of $960 \times 640$, while DSEC-3DOD adopts $320 \times 240$ for both images and events using a downsample. The event stream is converted into a voxel grid with 5 bins. 

% The DSEC-3DOD dataset provides sparse LiDAR data, which is processed by merging the preceding 9 LiDAR sweeps to enhance density. In contrast, Ev-Waymo utilizes only the current sweep without merging.
Waymo provides dense LiDAR data with 64 channels, whereas DSEC has only 16 channels, making it significantly sparser. Due to this sparsity, we attempted to accumulate LiDAR frames, but previous methods still struggled to train stably. Consequently, following prior works~\cite{zhou2024bring, wan2023rpeflow}, we used disparity maps to generate 3D points instead of directly utilizing raw LiDAR data.
We acknowledge that these disparity-based 3D points are obtained through offline processing. However, the key focus of this work is not to achieve state-of-the-art performance using LiDAR but to demonstrate the feasibility of blind time object detection using event cameras. Therefore, using these 3D points does not pose an issue for our study. Nevertheless, to enhance the usability of DSEC-3DOD dataset for future research, we have conducted additional experiments using raw LiDAR data and have shared the results at the following link\footnote{\url{https://github.com/mickeykang16/Ev3DOD/tree/main/Benchmark}}.

The voxel data is encoded using a 3D backbone~\cite{zhou2018voxelnet}, while the image and event data are processed using a common image encoder~\cite{Liu2021SwinTH}. 
 The small version of our model is discussed in Section~\ref{sec:inf_time}. In the Virtual 3D Event Fusion module, each box proposal is divided into $6\times6\times6$ sub-voxels.

\begin{figure*}[t]
\begin{center}
\includegraphics[width=.95\linewidth]{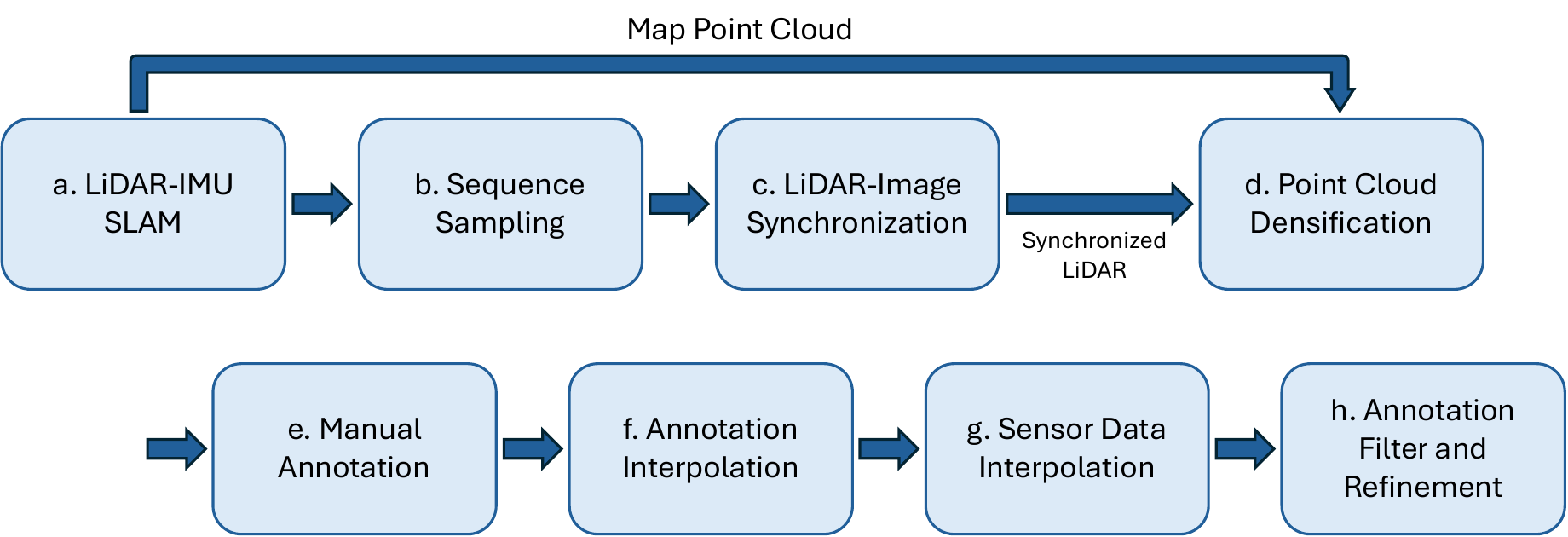}
\vspace{-3pt}
\caption{The overall pipeline for annotation. To enhance data quality, we perform software-based alignment and generate fine-grained 100 FPS ground-truths through additional annotation and post-processing.}
\label{fig:supple_ann_process}
\end{center}
% \vspace{-19pt}
\end{figure*}

\section{Inference Time}
\label{sec:inf_time}
To measure the inference time of our method and other approaches, we followed the speed measurement protocol from conventional event-based object detection~\cite{Gehrig2022RecurrentVT}, using the code provided at the given link\footnote{\url{https://github.com/uzh-rpg/RVT}}. We also performed GPU warm-up and ensured GPU-CPU synchronization using `torch.cuda.synchronize()' to accurately measure inference time. We measured inference time on a single NVIDIA A6000 GPU with a batch size of 1 and additionally designed a lightweight model, Ev-3DOD-\textit{Small}, to evaluate both performance and speed. Specifically, Ev-3DOD-\textit{Small} retains the overall structure of Ev-3DOD but replaces the event feature encoder with three simple convolution layers, reduces the number of pooling layers, and decreases the grid size in the Virtual 3D Event Fusion module. 

Table~\ref{tab:inference_time} compares performance and inference time among online methods. Looking at the performance metrics, our method, leveraging the event camera to infer during the blind time, achieves the best and second-best results across both approaches, with a significant margin over other methods. In terms of inference time, measured in FPS, even our full model (Ev-3DOD) achieves the fastest speed compared to other methods. This efficiency is attributed to our approach, which avoids recalculating point clouds and images during the blind time interval by explicitly leveraging events to update and reuse data at the present moment, making it highly cost-effective. Notably, when parameters are reduced, there is almost a twofold improvement in FPS with minimal performance degradation. This demonstrates that our method can effectively estimate 3D motion using event information without relying on a large number of parameters. We believe that the proposed Ev-3DOD, with its fast runtime using the high-frequency properties of an event camera, provides a promising direction for advancing future research in 3D object detection using event cameras.

\section{Event-based 3D Object Detection Datasets}
\label{sec:dataset}
In this section, we provide additional details about the dataset that may not have been fully covered in the main paper. Specifically, we delve into its structure, pre-processing steps, and unique characteristics critical for understanding the context and experimental results. By offering this comprehensive view, we aim to enhance the clarity and reproducibility of our work.

\subsection{DSEC-3DOD Dataset}

The DSEC~\cite{gehrig2021dsec} dataset provides LiDAR, stereo RGB images, and stereo events from diverse driving scenarios. To date, the DSEC dataset has been extensively studied for 2D perception tasks (\textit{e.g.} 2D object detection, semantic segmentation). In this study, we utilized this dataset for 3D perception for the first time and established the process of Fig.~\ref{fig:supple_ann_process} to provide fine-grained 100 FPS 3D detection ground truth. 

\noindent
\textbf{a. LiDAR-IMU SLAM}\\
For 3D detection annotation, a LiDAR sensor providing accurate depth information was designated as the reference sensor. The odometry of the reference sensor was estimated to synchronize LiDAR data with image timestamps, enabling accurate inter-modality alignment. Manual labeling was performed on a dense point cloud generated through pose-based LiDAR accumulation. To ensure precise LiDAR pose estimation, the LIO-Mapping~\cite{Ye2019TightlyC3} method was employed, consistent with the approach utilized in the DSEC dataset. The poses for the 10 FPS LiDAR data were subsequently obtained.

\begin{figure}[t]
\begin{center}
\includegraphics[width=.90\linewidth]{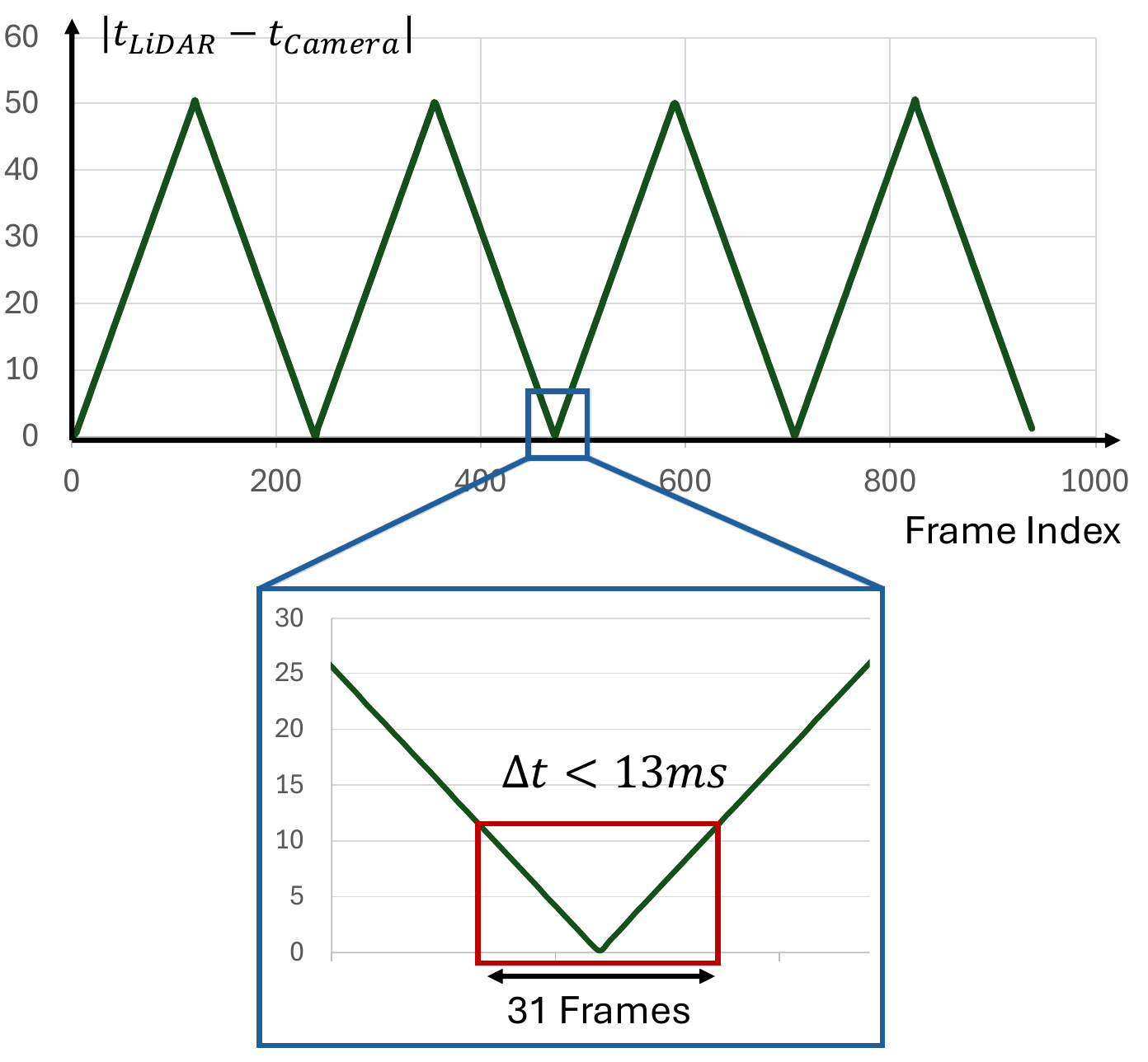}
\vspace{-8pt}
\caption{The time difference between images and closest LiDAR. To minimize temporal misalignment, 31 frames were sampled around the frame with the minimum time offset.}
\label{fig:supple_image_lidar_sync}
\end{center}
\vspace{-9pt}
\end{figure}

\noindent
\textbf{b. Sequence Sampling}\\
As mentioned in the main paper, we provide annotations for the ``zurich\_city" sequence. The DSEC provides images at 20 Hz and LiDAR data at 10 Hz. Images are sampled at 10 Hz following LiDAR. 
% The event data is aligned with the timestamps of the 10 Hz images. 
Although both the images and LiDAR data are sampled at the same 10 Hz rate, the lack of hardware time synchronization introduces temporal misalignment. To solve this problem, we utilize the sequence sampling strategy. Fig.~\ref{fig:supple_image_lidar_sync} illustrates the absolute time difference between the nearest image and LiDAR frames. Due to the periodic discrepancy between the two sensors, this misalignment repeats approximately every 237 frames, with a maximum time offset of half a sensor period (\textit{i.e.}, $50ms$). Therefore, we sampled 31 frames centered around the point of minimal time offset, reducing the maximum misalignment of $13 ms$.
As a result of sampling, the DSEC-3DOD dataset consists of 178 sequence chunks, each comprising 31 frames. Adjacent chunks are separated by a time gap of over 20 seconds, ensuring entirely different distributions in driving scenes. Table~\ref{tab:seq_train} and Table~\ref{tab:seq_test} show the train and test splits.

\noindent
\textbf{c. LiDAR-Image Synchronization}\\
Although the sequence sampling was designed to minimize time misalignment, synchronization errors still persist. Therefore, we further aligned the LiDAR data to the image timestamps using pose-based adjustments.

For an arbitrary RGB image $I_t$ at time $t$, the two closest-time LiDAR point clouds, $P_{t_0}$ and $P_{t_1}$, are identified, where $t_0 < t < t_1$. Assume the corresponding poses $X_{t_0}$ and $X_{t_1}$ are available from the mapping of Process \textbf{a}. Each pose is represented as a 3D coordinate and quaternion, denoted as $X = (x, y, z, Q)$, where $Q \in \mathbb{R}^4$. The image-aligned LiDAR pose $X_t$ is computed by interpolating $X_{t_0}$ and $X_{t_1}$. The position is interpolated linearly, while spherical linear interpolation (SLERP)~\cite{Shoemake1985AnimatingRW} is applied for the quaternions. 

The synchronized LiDAR point cloud $P_t$ is obtained using the transformation $P_t = T_t^{-1} T_{t'} P_{t'}$, where $T_t$ and $ T_{t'}$ are transformation matrices corresponding to the poses $X_t$ and $X_{t'}$, respectively. Here, $t'$ is the nearest time to $t$ as,
\begin{equation}
\begin{aligned}
t' =
\begin{cases} 
t_0 & \text{if } |t - t_0| < |t - t_1|, \\
t_1 & \text{otherwise.}
\end{cases}
\end{aligned}
\end{equation}
 
\begin{figure}[t]
\begin{center}
\includegraphics[width=.95\linewidth]{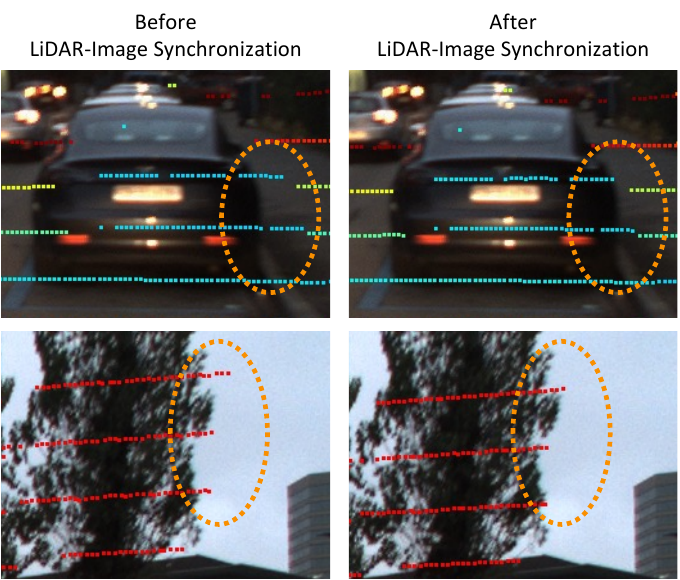}
\vspace{-8pt}
\caption{LiDAR projection on images before pose-based LiDAR synchronization (left) and after synchronization (right).}
\label{fig:supple_image_lidar_alignment_compare}
\end{center}
\vspace{-19pt}
\end{figure}

The effect of pose-based LiDAR synchronization is demonstrated in Fig.~\ref{fig:supple_image_lidar_alignment_compare}. LiDAR-image misalignment due to time offsets is most evident in scenes with large motions or significant rotations. By transforming the nearest LiDAR point cloud to the pose of the image timestamp, projection errors between sensors caused by time misalignment were minimized.

\noindent
\textbf{d. Point Cloud Densification}\\
Raw LiDAR data is inherently sparse, which can lead to reduced accuracy in ground truth bounding boxes if used directly for annotation. To address this, we utilized an accumulated point cloud created by combining multiple LiDAR scans with their relative poses. As noted in the DSEC~\cite{gehrig2021dsec}, LiDAR accumulation does not effectively handle occlusions or moving objects, both of which are critical considerations in the labeling process. To mitigate these issues, we employed filtered results (\textit{i.e.}, disparity) during the annotation process.

\noindent
\textbf{e. Manual Annotation}\\
Annotation experts labeled 3D bounding boxes on the densified 10 FPS LiDAR data, ensuring accuracy by considering both the LiDAR data and images. The ground truth consists of three classes: vehicle, pedestrian, and cyclist. Detailed annotation rules were derived from the guidelines provided by the Waymo Open Dataset~\cite{Sun2019ScalabilityIP}. To maintain high-quality annotations, bounding boxes containing fewer than two points or positioned beyond 50 meters were excluded from the ground truth.\\
\noindent
\textbf{f. Annotation Interpolation}\\
To generate 100 FPS annotations from the manually labeled 10 FPS annotations, bounding boxes were interpolated to create ground truth for blind times where neither LiDAR nor images were available. Linear interpolation was applied to the bounding box pose and size, while SLERP interpolation was applied for rotations. The interpolated annotations were subsequently refined through the following process.\\
\noindent
\textbf{g. Sensor Data Interpolation}\\
In process \textbf{f}, the automatically interpolated annotations are generally accurate but may not fully capture the dynamics of real-world motion. To enhance the quality of these annotations, sensor data was generated for the blind times. For images, realistic video frames were produced using a recent event-based video frame interpolation method~\cite{kim2023event}. Similarly, the latest techniques~\cite{zheng2023neuralpci} were utilized to generate accurate and realistic intermediate point cloud data.

\noindent
\textbf{h. Annotation Filter and Refinement}\\
The interpolated sensor data was employed to refine the annotations for the blind times. In most cases with minimal motion, bounding box interpolation alone yielded ground truth that aligned well with the sensor data. In such instances, refinements were avoided to preserve smooth bounding box poses and ensure temporal consistency. However, if the interpolated labels were misaligned with the sensor data or if sensor data was unavailable, the affected labels were filtered out.

\subsection{Ev-Waymo Dataset}
\noindent
\textbf{Event Synthesis.} The Waymo Open Dataset (WOD) provides 10 FPS synchronized images, LiDAR, and 3D bounding box labels. To generate the events, we utilize a widely adopted event simulation model~\cite{Gehrig_2020_CVPR} to synthesize the events from video data. This enables us to utilize temporally dense events between image and LiDAR active timestamps for training and testing.

\noindent
\textbf{Annotation Interpolation and Refinement.}
Since WOD provides dense 10 FPS annotations, we can obtain 100 FPS ground truth through annotation interpolation and refinement. A process similar to the \textbf{f}, \textbf{g}, and \textbf{h} steps in DSEC-3DOD dataset processing was employed. We interpolated the 10 FPS bounding box information, including pose, dimensions, and heading, provided by WOD to generate 100 FPS data. In addition, we synthesized blind-time data of camera and LiDAR using an interpolation algorithm for refinement and filtering, ensuring higher quality.

% The pose, dimensions, and heading of the 10 FPS bounding boxes were interpolated, while synthetic blind-time images and point clouds were used for refinement and filtering.

\section{More Qualitative Results and Videos}
\label{sec:qual_video}
\begin{figure*}[p]
\begin{center}
\includegraphics[width=.94\linewidth]{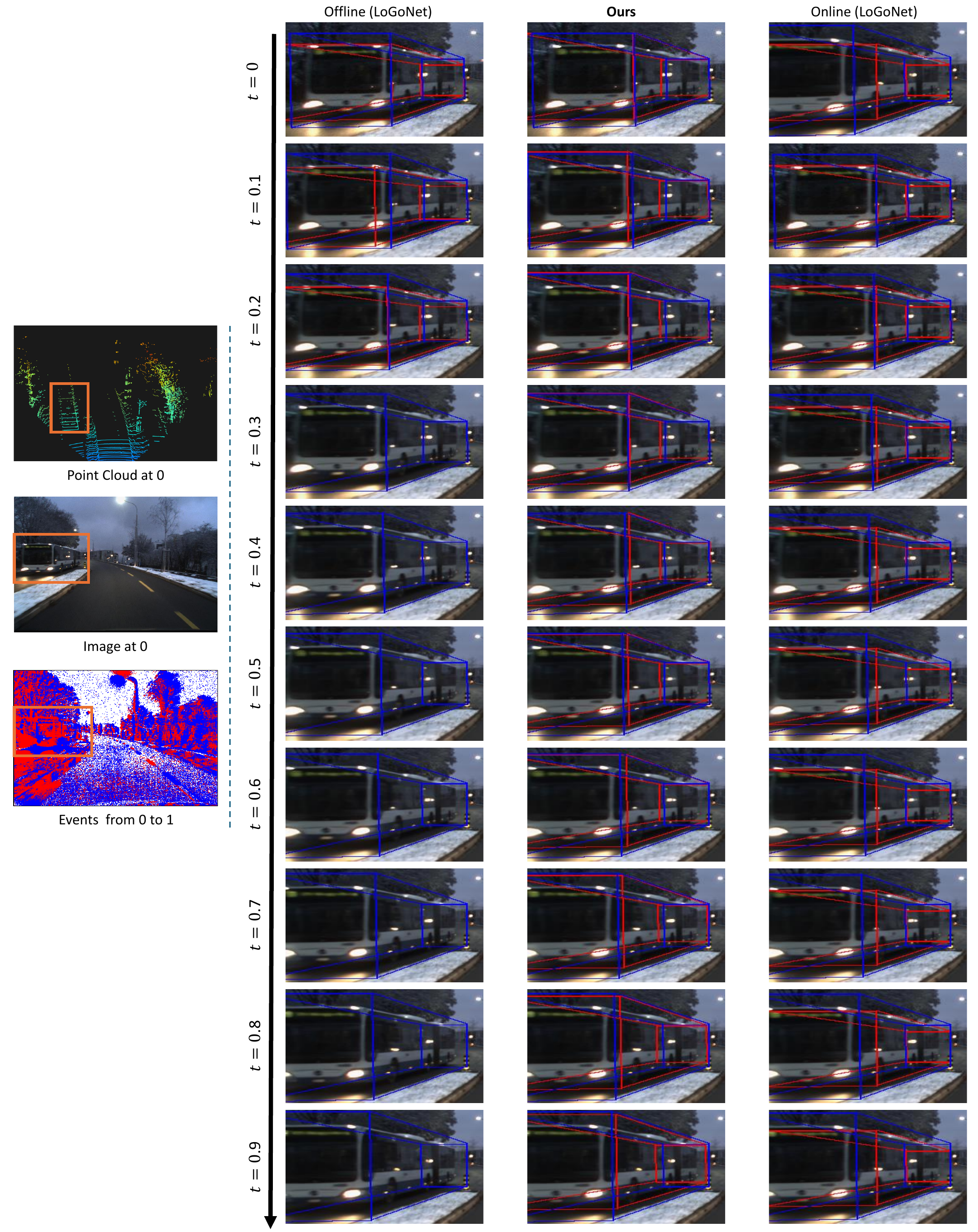}
\vspace{-11pt}
\caption{Qualitative comparisons with other offline and online methods on the DSEC-3DOD dataset. $t=0$ represents the active time, while $t = 0.1, 0.2, 0.3, 0.4, 0.5, 0.6, 0.7, 0.8, 0.9$ denote the blind times. The \textcolor{blue}{blue} box indicates ground truth, and the \textcolor{red}{red} box shows predictions. For better understanding, we overlaid the results onto the images generated by the interpolation method~\cite{kim2023event}.
}
\label{fig:dsec_qual_1}
\end{center}
\vspace{-19pt}
\end{figure*}

\begin{figure*}[p]
\begin{center}
\includegraphics[width=.94\linewidth]{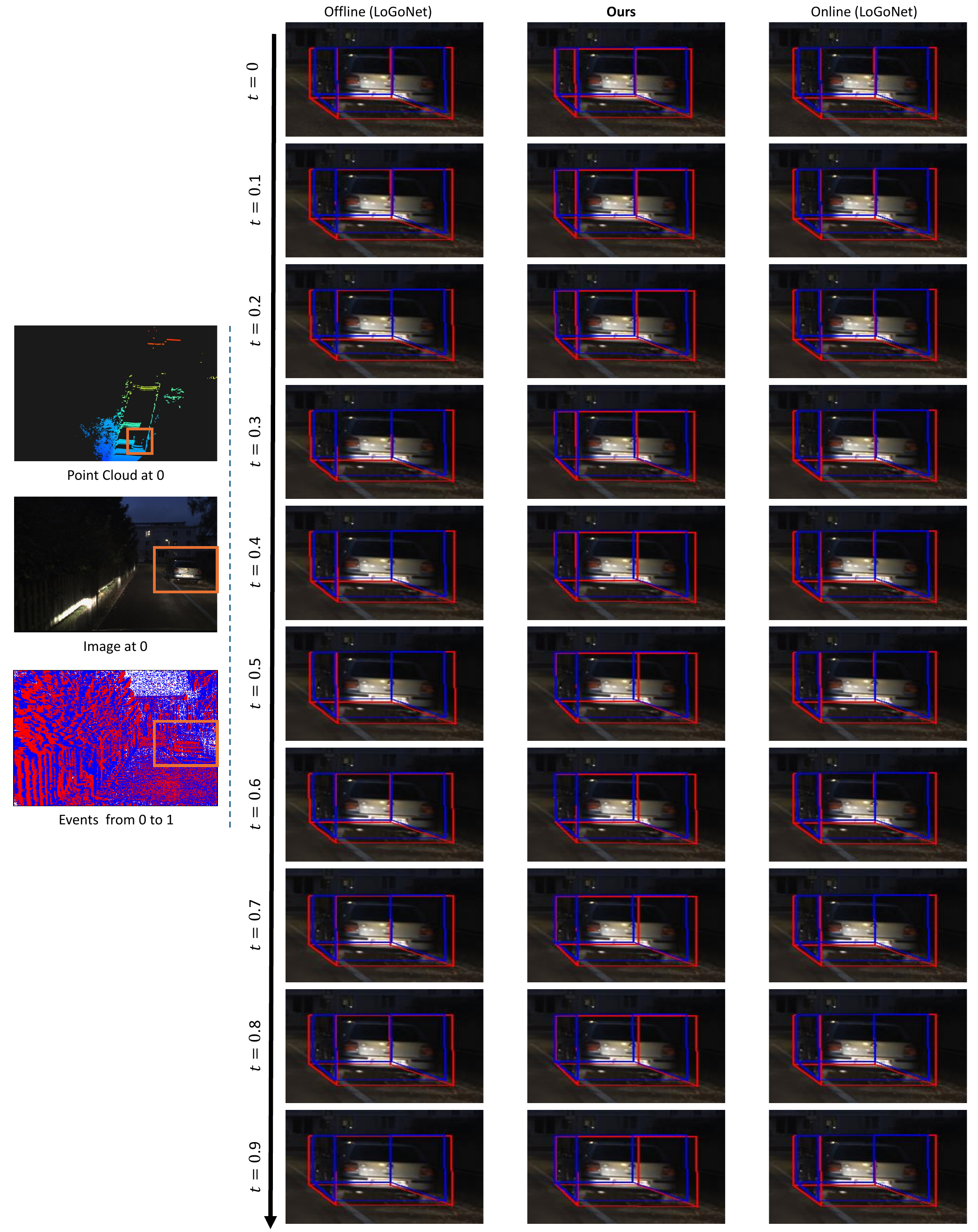}
\vspace{-11pt}
\caption{Qualitative comparisons with other offline and online methods on the DSEC-3DOD dataset. $t=0$ represents the active time, while $t = 0.1, 0.2, 0.3, 0.4, 0.5, 0.6, 0.7, 0.8, 0.9$ denote the blind times. The \textcolor{blue}{blue} box indicates ground truth, and the \textcolor{red}{red} box shows predictions. For better understanding, we overlaid the results onto the images generated by the interpolation method~\cite{kim2023event}.
}
\label{fig:dsec_qual_2}
\end{center}
\vspace{-19pt}
\end{figure*}

\begin{figure*}[p]
\begin{center}
\includegraphics[width=.94\linewidth]{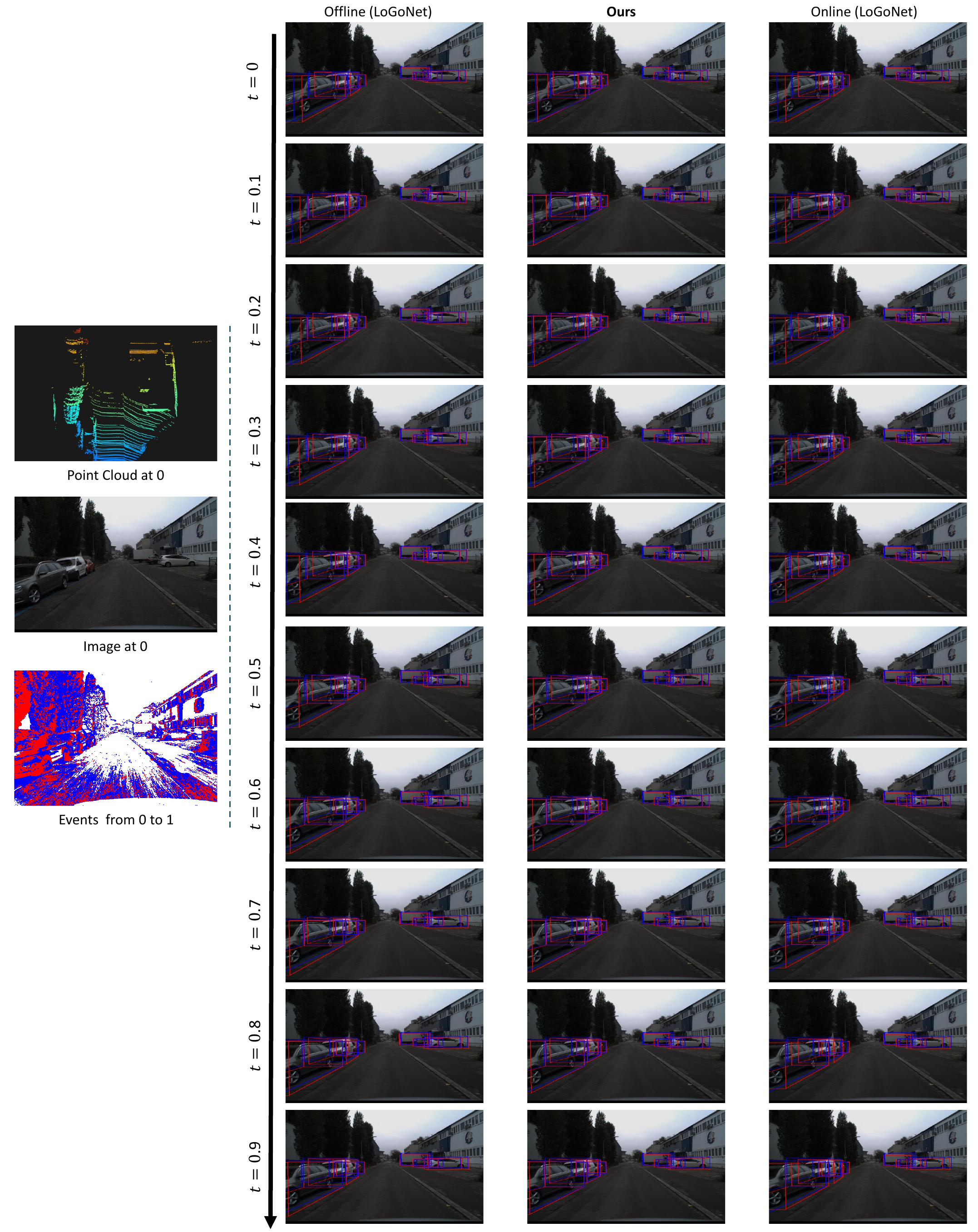}
\vspace{-11pt}
\caption{Qualitative comparisons with other offline and online methods on the DSEC-3DOD dataset. $t=0$ represents the active time, while $t = 0.1, 0.2, 0.3, 0.4, 0.5, 0.6, 0.7, 0.8, 0.9$ denote the blind times. The \textcolor{blue}{blue} box indicates ground truth, and the \textcolor{red}{red} box shows predictions. For better understanding, we overlaid the results onto the images generated by the interpolation method~\cite{kim2023event}.
}
\label{fig:dsec_qual_3}
\end{center}
\vspace{-19pt}
\end{figure*}

To provide a more comprehensive understanding of the proposed model, we present additional qualitative results in Figures~\ref{fig:dsec_qual_1}, \ref{fig:dsec_qual_2}, and \ref{fig:dsec_qual_3}, showcasing its performance on the DSEC-3DOD dataset. The proposed method consistently predicts bounding boxes that closely align with the ground truth across various challenging environments. 

Figure~\ref{fig:dsec_qual_1} illustrates a challenging scene involving a bus, where size estimation is particularly difficult. The offline method, even with access to future information, fails to detect certain instances. Likewise, the online method demonstrates increasing errors compared to the ground truth as time progresses beyond the active timestamp. In contrast, the proposed method shows robust performance, generating predictions that closely align with the ground truth labels.

The introduced DSEC-3DOD dataset features challenging scenarios, including night scenes, as shown in Fig.~\ref{fig:dsec_qual_2}. Unlike the Ev-Waymo dataset, which cannot leverage challenging illumination conditions to generate synthetic events, our proposed real event dataset enables validation in such scenarios. In the night scene, the proposed method exhibits accurate box predictions compared to both the offline and online methods.

\begin{figure*}[p]
\begin{center}
\includegraphics[width=.94\linewidth]{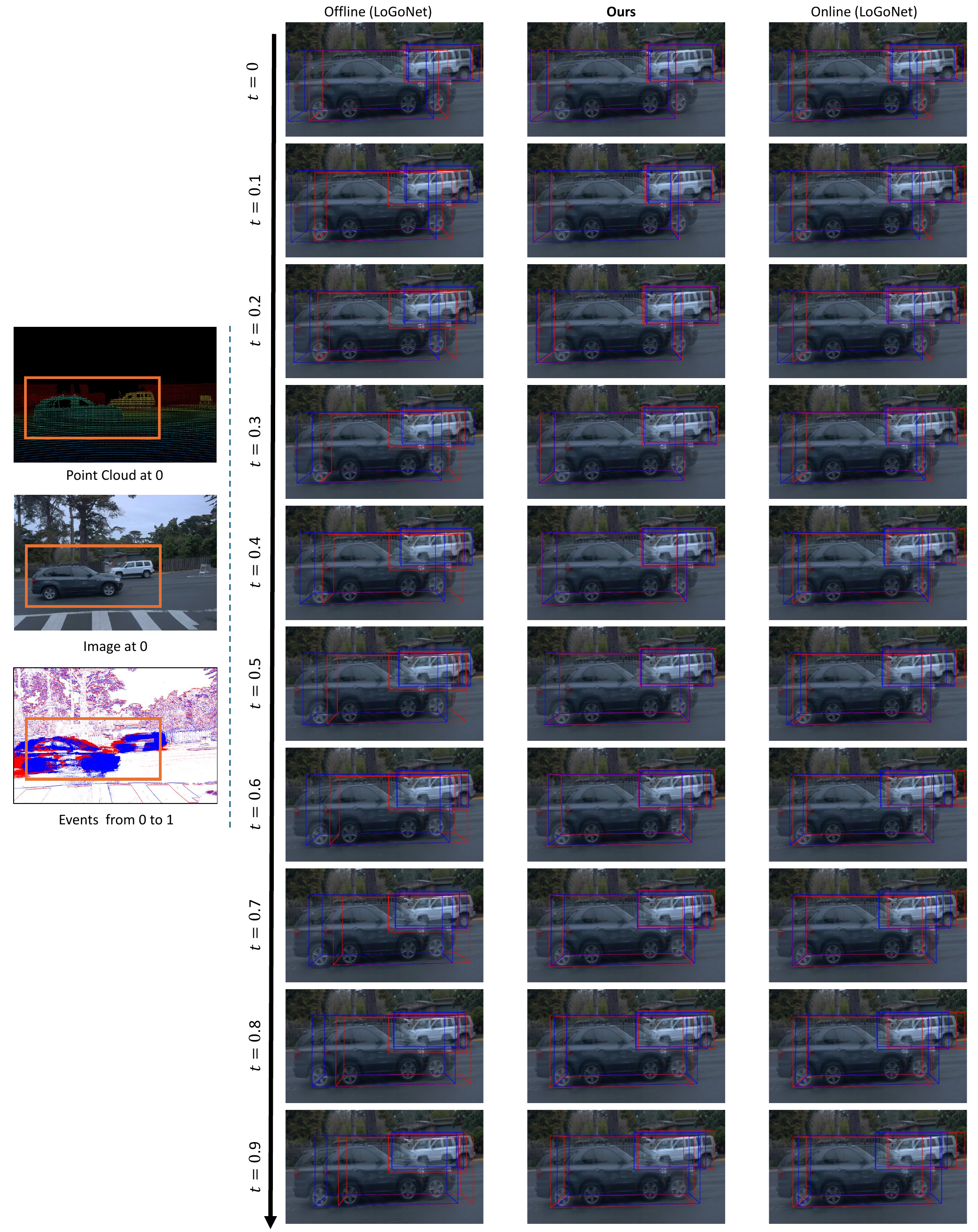}
\vspace{-11pt}
\caption{Qualitative comparisons of our method with other offline and online evaluations on the Ev-Waymo dataset. $t=0$ represents the active time, while $t = 0.1, 0.2, 0.3, 0.4, 0.5, 0.6, 0.7, 0.8, 0.9$ denote the blind times. The \textcolor{blue}{blue} box represents the ground truth, while the \textcolor{red}{red} box shows the prediction results of each method. For easier understanding, images at active timestamps 0 and 1 are overlaid.}
\label{fig:waymo_qual2}
\end{center}
\vspace{-19pt}
\end{figure*}

\begin{figure*}[p]
\begin{center}
\includegraphics[width=.94\linewidth]{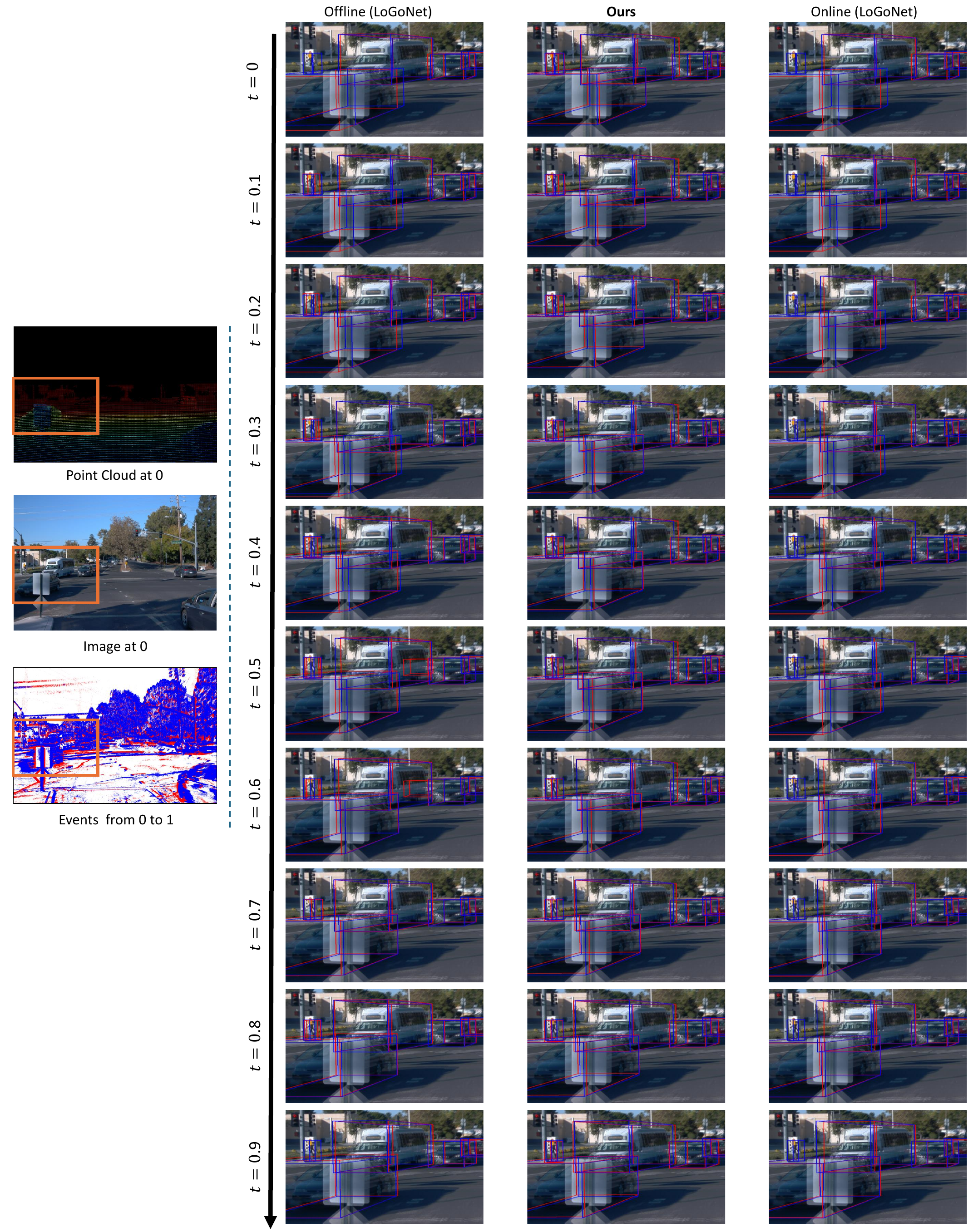}
\vspace{-11pt}
\caption{Qualitative comparisons of our method with other offline and online evaluations on the Ev-Waymo dataset. $t=0$ represents the active time, while $t = 0.1, 0.2, 0.3, 0.4, 0.5, 0.6, 0.7, 0.8, 0.9$ denote the blind times. The \textcolor{blue}{blue} box represents the ground truth, while the \textcolor{red}{red} box shows the prediction results of each method. For easier understanding, images at active timestamps 0 and 1 are overlaid.}
\label{fig:waymo_qual3}
\end{center}
\vspace{-19pt}
\end{figure*}

Figure~\ref{fig:waymo_qual}, \ref{fig:waymo_qual2}, and \ref{fig:waymo_qual3} present the results on the Ev-Waymo dataset, which features numerous complex sequences with high object density. The proposed model effectively predicts complex object motions, producing bounding boxes closely aligned with the ground truth. In such scenes, even with access to future data, interpolating sensor information during blind times remains challenging, which complicates precise bounding box predictions. Consequently, the qualitative results on Ev-Waymo demonstrate that the proposed method outperforms others by delivering more accurate bounding boxes.

We provide a short video to showcase the datasets used in the experiments and the results on sequential data. The proposed method demonstrates robust performance across various environments in both the DSEC-3DOD dataset and the Ev-Waymo dataset. Notably, it accurately estimates the motion of object bounding boxes even in challenging night scenes within the DSEC-3DOD dataset.

% \noindent
% \textbf{Ev-Waymo}

% \noindent
% \textbf{DSEC-3DOD}

\section{Hyper-parameter Analysis}
\label{sec:hyper}
As shown in Table~\ref{tab:hyper}, we conduct an ablation study on the loss weights. The box regression loss and confidence prediction loss weights were set to 0.1, 1.0, and 10.0, respectively, during model training. The results demonstrated that the model remained robust, producing consistent outcomes despite changes in the loss magnitudes.

\begin{table}[t]
\caption{The result according to hyper-paramter in Eq.~(3) on Ev-Waymo LEVEL 2 (L2). $\lambda_1$ and $\lambda_2$ refer to the weight of box regression loss and binary cross entropy loss, respectively.}
\label{tab:hyper}
\vspace{-12pt}
\begin{center}
\renewcommand{\arraystretch}{1.2}
\resizebox{0.99\linewidth}{!}{\renewcommand{\tabcolsep}{6.0pt}
\begin{tabular}{c|cc|cc|cc}
% \hline
%                      & \multicolumn{3}{c|}{N-Caltech101} & \multicolumn{3}{c}{N-ImageNet~(Mini)} \\ 
\hline
\multirow{2}{*}{$\lambda_1$ \textbackslash{} $\lambda_2$} & \multicolumn{2}{c|}{0.1}
& \multicolumn{2}{c|}{1.0} 
 & \multicolumn{2}{c}{10.0}   \\
\cline{2-7}
& mAP & mAPH & mAP & mAPH & mAP & mAPH\\ 
\hline
0.1 & \gradient{47.32} & \gradienttwo{44.87} & 
\gradient{47.70} & \gradienttwo{45.22} & 
\gradient{47.14} & \gradienttwo{44.70} 
\\
% 0.05                & \gradient{83.52}    & \gradient{82.02}   & \gradient{81.66}   
% \\
1.0 & \gradient{47.72} & \gradienttwo{45.25}  &
\gradient{48.06} & \gradienttwo{45.60}  &
\gradient{46.98} & \gradienttwo{44.53} 
\\
% 0.5                 & \gradient{81.99}    & \gradient{82.39}   & \gradient{82.61}   
% \\
10.0  & \gradient{47.46} & \gradienttwo{45.02}  & 
\gradient{47.20} & \gradienttwo{44.74}  & 
\gradient{47.32} & \gradienttwo{44.86} 
\\ 
\hline

\end{tabular}}
\end{center}
\vspace{-5pt}
\end{table}

\begin{table*}[p]
\begin{center}
\caption{The train/test sequence splits of Ev-Waymo dataset.}
\vspace{-5pt}
\resizebox{0.99\linewidth}{!}{\renewcommand{\tabcolsep}{5.6pt}
\begin{tabular}{c||c|c|c}
\hline
\textbf{Dataset}   & \multicolumn{2}{c}{\textbf{Ev-Waymo} }  \\ \hline\hline
\multirow{3}{*}{\textbf{Split}}  & \multirow{3}{*}{\textbf{Sequence Name}} & \multirow{3}{*}{\thead{ \textbf{No.}\\ \textbf{Seq.}}} & \textbf{No.}  \\
& & & \textbf{Labeled} \\
& & & \textbf{Scenes} \\
\hline
Train &  
\begin{tabular}[c]{@{}c@{}}
segment-207754730878135627\_1140\_000\_1160\_000, 
segment-13840133134545942567\_1060\_000\_1080\_000, \\
segment-8327447186504415549\_5200\_000\_5220\_000,
segment-10964956617027590844\_1584\_680\_1604\_680, \\
segment-11918003324473417938\_1400\_000\_1420\_000,
segment-15448466074775525292\_2920\_000\_2940\_000,\\
segment-14830022845193837364\_3488\_060\_3508\_060,
segment-11379226583756500423\_6230\_810\_6250\_810, \\
segment-7861168750216313148\_1305\_290\_1325\_290,
segment-13506499849906169066\_120\_000\_140\_000, \\
segment-6229371035421550389\_2220\_000\_2240\_000,
segment-15882343134097151256\_4820\_000\_4840\_000,\\
segment-14098605172844003779\_5084\_630\_5104\_630,
segment-8582923946352460474\_2360\_000\_2380\_000, \\
segment-16485056021060230344\_1576\_741\_1596\_741,
segment-915935412356143375\_1740\_030\_1760\_030, \\
segment-3002379261592154728\_2256\_691\_2276\_691,
segment-4348478035380346090\_1000\_000\_1020\_000,\\
segment-2036908808378190283\_4340\_000\_4360\_000,
segment-15844593126368860820\_3260\_000\_3280\_000, \\
segment-5835049423600303130\_180\_000\_200\_000,
segment-15696964848687303249\_4615\_200\_4635\_200, \\
segment-7543690094688232666\_4945\_350\_4965\_350,
segment-16372013171456210875\_5631\_040\_5651\_040,\\
segment-14193044537086402364\_534\_000\_554\_000,
segment-550171902340535682\_2640\_000\_2660\_000, \\
segment-4641822195449131669\_380\_000\_400\_000,
segment-7239123081683545077\_4044\_370\_4064\_370, \\
segment-11928449532664718059\_1200\_000\_1220\_000,
segment-5100136784230856773\_2517\_300\_2537\_300,\\
segment-13182548552824592684\_4160\_250\_4180\_250,
segment-14004546003548947884\_2331\_861\_2351\_861, \\
segment-2570264768774616538\_860\_000\_880\_000,
segment-7440437175443450101\_94\_000\_114\_000, \\
segment-15717839202171538526\_1124\_920\_1144\_920,
segment-8148053503558757176\_4240\_000\_4260\_000,\\
segment-16977844994272847523\_2140\_000\_2160\_000,
segment-5451442719480728410\_5660\_000\_5680\_000, \\
segment-7290499689576448085\_3960\_000\_3980\_000,
segment-16801666784196221098\_2480\_000\_2500\_000, \\
segment-4916527289027259239\_5180\_000\_5200\_000,
segment-16202688197024602345\_3818\_820\_3838\_820,\\
segment-9758342966297863572\_875\_230\_895\_230,
segment-12161824480686739258\_1813\_380\_1833\_380,\\
segment-14369250836076988112\_7249\_040\_7269\_040,
segment-2752216004511723012\_260\_000\_280\_000,\\
segment-10444454289801298640\_4360\_000\_4380\_000,
segment-17388121177218499911\_2520\_000\_2540\_000,\\
segment-7885161619764516373\_289\_280\_309\_280,
segment-16561295363965082313\_3720\_000\_3740\_000,\\
segment-11199484219241918646\_2810\_030\_2830\_030,
segment-4575961016807404107\_880\_000\_900\_000,\\
segment-7566697458525030390\_1440\_000\_1460\_000,
segment-10275144660749673822\_5755\_561\_5775\_561,\\
segment-6193696614129429757\_2420\_000\_2440\_000,
segment-12251442326766052580\_1840\_000\_1860\_000, \\
segment-13271285919570645382\_5320\_000\_5340\_000,
segment-9015546800913584551\_4431\_180\_4451\_180, \\
segment-10596949720463106554\_1933\_530\_1953\_530, 
segment-15942468615931009553\_1243\_190\_1263\_190, \\
segment-15125792363972595336\_4960\_000\_4980\_000, 
segment-1422926405879888210\_51\_310\_71\_310, \\
segment-5576800480528461086\_1000\_000\_1020\_000,
segment-1255991971750044803\_1700\_000\_1720\_000 \\
\end{tabular} 
&  64 & 126,330 
\\
\hline
Test & 
\begin{tabular}[c]{@{}c@{}}
segment-18446264979321894359\_3700\_000\_3720\_000,  
segment-17152649515605309595\_3440\_000\_3460\_000, \\
segment-16213317953898915772\_1597\_170\_1617\_170,
segment-5183174891274719570\_3464\_030\_3484\_030, \\
segment-3126522626440597519\_806\_440\_826\_440,
segment-3077229433993844199\_1080\_000\_1100\_000,\\ 
segment-10289507859301986274\_4200\_000\_4220\_000, 
segment-30779396576054160\_1880\_000\_1900\_000, \\
segment-9243656068381062947\_1297\_428\_1317\_428, 
segment-2834723872140855871\_1615\_000\_1635\_000,\\ 
segment-2736377008667623133\_2676\_410\_2696\_410,
segment-15948509588157321530\_7187\_290\_7207\_290,\\ 
segment-9231652062943496183\_1740\_000\_1760\_000, 
segment-4854173791890687260\_2880\_000\_2900\_000, \\
segment-6324079979569135086\_2372\_300\_2392\_300,
segment-6001094526418694294\_4609\_470\_4629\_470 \\
\end{tabular} 
& 16 & 31,550 \\
\hline
\end{tabular}}
\end{center}
\vspace{-5pt}
\label{tab:ev_waymo_split}
\vspace{-10pt}
\end{table*}

\begin{table}[p]
\centering
\caption{The train sequence splits of DSEC-3DOD dataset.}
\vspace{-7pt}
\renewcommand{\arraystretch}{1.1}
\setlength\tabcolsep{4.5pt}
\resizebox{.99\linewidth}{!}{
\begin{tabular}{l|l|ccc}
\hline
\textbf{Split} & \textbf{Time} & \textbf{Sequence} & \textbf{\# Frames} & \textbf{\# GT Scenes} \\ \hline
\textbf{Train} & \textbf{Day} & 
 zurich\_city\_00\_a & 31 & 301\\ 
& &zurich\_city\_00\_b & 155 & 1,505\\ 
& &zurich\_city\_01\_a & 62 & 602\\ 
& &zurich\_city\_01\_b & 155 & 1,505\\ 
& &zurich\_city\_01\_c & 124 & 1,204\\ 
& &zurich\_city\_01\_d & 93 & 903\\ 
& &zurich\_city\_01\_e & 217 & 2,107\\ 
& &zurich\_city\_01\_f & 155 & 1,505\\ 
& &zurich\_city\_02\_a & 31 & 301\\ 
& &zurich\_city\_02\_b & 124 & 1,204\\ 
& &zurich\_city\_02\_c & 279 & 2,709\\ 
& &zurich\_city\_02\_d & 155 & 1,505\\ 
& &zurich\_city\_02\_e & 186 & 1,806\\ 
& &zurich\_city\_04\_a & 93 & 903\\ 
& &zurich\_city\_04\_c & 93 & 903\\ 
& &zurich\_city\_04\_d & 93 & 903\\ 
& &zurich\_city\_04\_e & 31 & 301\\ 
& &zurich\_city\_04\_f & 124 & 1,204\\ 
& &zurich\_city\_05\_a & 217 & 2,107\\ 
& &zurich\_city\_05\_b & 124 & 1,204\\ 
& &zurich\_city\_06\_a & 186 & 1,806\\ 
& &zurich\_city\_07\_a & 124 & 1,204\\ 
& &zurich\_city\_08\_a & 62 & 602\\ 
& &zurich\_city\_11\_a & 31 & 301\\ 
& &zurich\_city\_11\_b & 93 & 903\\ 
& &zurich\_city\_11\_c & 155 & 1,505\\ 
\cline{2-5}
& \textbf{Day Total} & & \textbf{3,193} & \textbf{31,003} \\
\cline{2-5}
 & \textbf{Night} & 
 zurich\_city\_03\_a & 62 & 602\\ 
& &zurich\_city\_09\_a & 217 & 2,107\\ 
& &zurich\_city\_09\_b & 31 & 301\\ 
& &zurich\_city\_09\_c & 155 & 1,505\\ 
& &zurich\_city\_09\_d & 124 & 1,204\\ 
& &zurich\_city\_09\_e & 93 & 903\\ 
& &zurich\_city\_10\_a & 248 & 2,408\\ 
& &zurich\_city\_10\_b & 217 & 2,107\\ 
\cline{2-5}
& \textbf{Night Total} & & \textbf{1,147} & \textbf{11,137} \\
\cline{2-5} 
& \textbf{Train Total} & & \textbf{4,340} & \textbf{42,140} \\
\hline
\end{tabular}}
\label{tab:seq_train}
\end{table}

\begin{table}[p]
\centering
\caption{The test sequence splits of DSEC-3DOD dataset.}
\vspace{-7pt}
\renewcommand{\arraystretch}{1.1}
\setlength\tabcolsep{4.5pt}
\resizebox{.99\linewidth}{!}{
\begin{tabular}{l|l|ccc}
\hline
\textbf{Split} & \textbf{Time} & \textbf{Sequence} & \textbf{\# Frames} & \textbf{\# GT Scenes} \\ \hline
\textbf{Test} & \textbf{Day} & 
zurich\_city\_00\_a & 62 & 602\\ 
& &zurich\_city\_00\_b & 31 & 301\\ 
& &zurich\_city\_01\_e & 31 & 301\\ 
& &zurich\_city\_01\_f & 62 & 602\\ 
& &zurich\_city\_02\_b & 31 & 301\\ 
& &zurich\_city\_02\_c & 93 & 903\\ 
& &zurich\_city\_02\_d & 62 & 602\\ 
& &zurich\_city\_04\_c & 31 & 301\\ 
& &zurich\_city\_04\_d & 31 & 301\\ 
& &zurich\_city\_05\_b & 62 & 602\\ 
& &zurich\_city\_06\_a & 31 & 301\\ 
& &zurich\_city\_07\_a & 62 & 602\\ 
& &zurich\_city\_08\_a & 31 & 301\\ 
& &zurich\_city\_11\_b & 155 & 1,505\\ 
& &zurich\_city\_11\_c & 93 & 903\\ 
\cline{2-5}
& \textbf{Day Total} & & \textbf{868} & \textbf{8,428} \\
\cline{2-5} 
& \textbf{Night} &  
zurich\_city\_03\_a & 31 & 301\\ 
& &zurich\_city\_09\_a & 31 & 301\\ 
& &zurich\_city\_09\_c & 31 & 301\\ 
& &zurich\_city\_09\_d & 93 & 903\\ 
& &zurich\_city\_10\_a & 31 & 301\\ 
& &zurich\_city\_10\_b & 93 & 903\\ 
\cline{2-5}
& \textbf{Night Total} & & \textbf{310} & \textbf{3,010} \\
\cline{2-5}
& \textbf{Test Total} & & \textbf{1178} & \textbf{11,438} \\
\hline
\end{tabular}}
\label{tab:seq_test}
\end{table}

\begin{figure*}[p]
\begin{center}
\includegraphics[width=.94\linewidth]{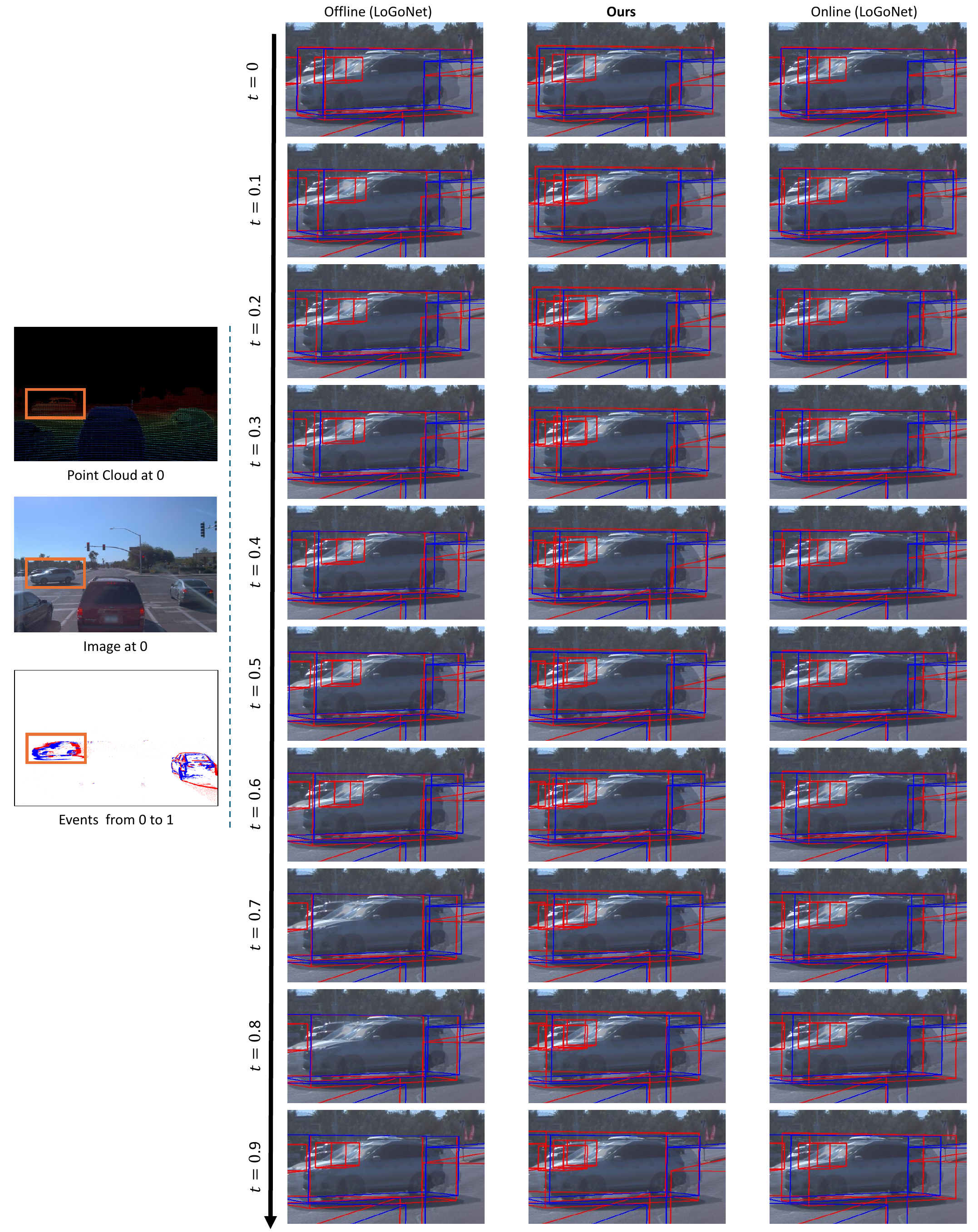}
\vspace{-11pt}
\caption{Qualitative comparisons of our method with other offline and online evaluations on the Ev-Waymo dataset. $t=0$ represents the active time, while $t = 0.1, 0.2, 0.3, 0.4, 0.5, 0.6, 0.7, 0.8, 0.9$ denote the blind times. The \textcolor{blue}{blue} box represents the ground truth, while the \textcolor{red}{red} box shows the prediction results of each method. For easier understanding, images at active timestamps 0 and 1 are overlaid.}
\label{fig:waymo_qual}
\end{center}
\vspace{-19pt}
\end{figure*}

{
    \small
    \bibliographystyle{ieeenat_fullname}
    \bibliography{main}

\begin{thebibliography}{130}
\providecommand{\natexlab}[1]{#1}
\providecommand{\url}[1]{\texttt{#1}}
\expandafter\ifx\csname urlstyle\endcsname\relax
  \providecommand{\doi}[1]{doi: #1}\else
  \providecommand{\doi}{doi: \begingroup \urlstyle{rm}\Url}\fi

\bibitem[Bai et~al.(2022)Bai, Hu, Zhu, Huang, Chen, Fu, and Tai]{bai2022transfusion}
Xuyang Bai, Zeyu Hu, Xinge Zhu, Qingqiu Huang, Yilun Chen, Hongbo Fu, and Chiew-Lan Tai.
\newblock Transfusion: Robust lidar-camera fusion for 3d object detection with transformers.
\newblock In \emph{Proceedings of the IEEE/CVF conference on computer vision and pattern recognition}, pages 1090--1099, 2022.

\bibitem[Bi et~al.(2019)Bi, Chadha, Abbas, Bourtsoulatze, and Andreopoulos]{Bi2019GraphBasedOC}
Yin Bi, Aaron Chadha, Alhabib Abbas, Eirina Bourtsoulatze, and Yiannis Andreopoulos.
\newblock Graph-based object classification for neuromorphic vision sensing.
\newblock \emph{2019 IEEE/CVF International Conference on Computer Vision (ICCV)}, pages 491--501, 2019.

\bibitem[Chen et~al.(2017)Chen, Ma, Wan, Li, and Xia]{chen2017multi}
Xiaozhi Chen, Huimin Ma, Ji Wan, Bo Li, and Tian Xia.
\newblock Multi-view 3d object detection network for autonomous driving.
\newblock In \emph{Proceedings of the IEEE conference on Computer Vision and Pattern Recognition}, pages 1907--1915, 2017.

\bibitem[Chen et~al.(2022{\natexlab{a}})Chen, Shi, Zhu, Cheung, Xu, and Li]{chen2022mppnet}
Xuesong Chen, Shaoshuai Shi, Benjin Zhu, Ka~Chun Cheung, Hang Xu, and Hongsheng Li.
\newblock Mppnet: Multi-frame feature intertwining with proxy points for 3d temporal object detection.
\newblock In \emph{European Conference on Computer Vision}, pages 680--697. Springer, 2022{\natexlab{a}}.

\bibitem[Chen et~al.(2022{\natexlab{b}})Chen, Zhang, Wang, Wang, and Zhao]{Chen2022FUTR3DAU}
Xuanyao Chen, Tianyuan Zhang, Yue Wang, Yilun Wang, and Hang Zhao.
\newblock Futr3d: A unified sensor fusion framework for 3d detection.
\newblock \emph{2023 IEEE/CVF Conference on Computer Vision and Pattern Recognition Workshops (CVPRW)}, pages 172--181, 2022{\natexlab{b}}.

\bibitem[Chen et~al.(2019)Chen, Liu, Shen, and Jia]{chen2019fast}
Yilun Chen, Shu Liu, Xiaoyong Shen, and Jiaya Jia.
\newblock Fast point r-cnn.
\newblock In \emph{Proceedings of the IEEE/CVF international conference on computer vision}, pages 9775--9784, 2019.

\bibitem[Chen et~al.(2020)Chen, Liu, Shen, and Jia]{chen2020dsgn}
Yilun Chen, Shu Liu, Xiaoyong Shen, and Jiaya Jia.
\newblock Dsgn: Deep stereo geometry network for 3d object detection.
\newblock In \emph{Proceedings of the IEEE/CVF conference on computer vision and pattern recognition}, pages 12536--12545, 2020.

\bibitem[Chen et~al.(2022{\natexlab{c}})Chen, Li, Zhang, Sun, and Jia]{chen2022focal}
Yukang Chen, Yanwei Li, Xiangyu Zhang, Jian Sun, and Jiaya Jia.
\newblock Focal sparse convolutional networks for 3d object detection.
\newblock In \emph{Proceedings of the IEEE/CVF Conference on Computer Vision and Pattern Recognition}, pages 5428--5437, 2022{\natexlab{c}}.

\bibitem[Chen et~al.(2023)Chen, Liu, Zhang, Qi, and Jia]{chen2023voxelnext}
Yukang Chen, Jianhui Liu, Xiangyu Zhang, Xiaojuan Qi, and Jiaya Jia.
\newblock Voxelnext: Fully sparse voxelnet for 3d object detection and tracking.
\newblock In \emph{Proceedings of the IEEE/CVF Conference on Computer Vision and Pattern Recognition}, pages 21674--21683, 2023.

\bibitem[Chen et~al.(2022{\natexlab{d}})Chen, Dai, and Ding]{chen2022pseudo}
Yi-Nan Chen, Hang Dai, and Yong Ding.
\newblock Pseudo-stereo for monocular 3d object detection in autonomous driving.
\newblock In \emph{Proceedings of the IEEE/CVF conference on computer vision and pattern recognition}, pages 887--897, 2022{\natexlab{d}}.

\bibitem[Chen et~al.(2022{\natexlab{e}})Chen, Li, Zhang, Fang, Jiang, and Zhao]{chen2022deformable}
Zehui Chen, Zhenyu Li, Shiquan Zhang, Liangji Fang, Qinhong Jiang, and Feng Zhao.
\newblock Deformable feature aggregation for dynamic multi-modal 3d object detection.
\newblock In \emph{European conference on computer vision}, pages 628--644. Springer, 2022{\natexlab{e}}.

\bibitem[Deng et~al.(2021)Deng, Shi, Li, Zhou, Zhang, and Li]{deng2021voxel}
Jiajun Deng, Shaoshuai Shi, Peiwei Li, Wengang Zhou, Yanyong Zhang, and Houqiang Li.
\newblock Voxel r-cnn: Towards high performance voxel-based 3d object detection.
\newblock In \emph{Proceedings of the AAAI conference on artificial intelligence}, pages 1201--1209, 2021.

\bibitem[Fan et~al.(2022)Fan, Pang, Zhang, Wang, Zhao, Wang, Wang, and Zhang]{fan2022embracing}
Lue Fan, Ziqi Pang, Tianyuan Zhang, Yu-Xiong Wang, Hang Zhao, Feng Wang, Naiyan Wang, and Zhaoxiang Zhang.
\newblock Embracing single stride 3d object detector with sparse transformer.
\newblock In \emph{Proceedings of the IEEE/CVF conference on computer vision and pattern recognition}, pages 8458--8468, 2022.

\bibitem[Feng et~al.(2024)Feng, Quan, Wang, Wang, and Yang]{feng2024interpretable3d}
Tuo Feng, Ruijie Quan, Xiaohan Wang, Wenguan Wang, and Yi Yang.
\newblock Interpretable3d: An ad-hoc interpretable classifier for 3d point clouds.
\newblock In \emph{Proceedings of the AAAI Conference on Artificial Intelligence}, pages 1761--1769, 2024.

\bibitem[Gallego et~al.(2020)Gallego, Delbr{\"u}ck, Orchard, Bartolozzi, Taba, Censi, Leutenegger, Davison, Conradt, Daniilidis, et~al.]{gallego2020event}
Guillermo Gallego, Tobi Delbr{\"u}ck, Garrick Orchard, Chiara Bartolozzi, Brian Taba, Andrea Censi, Stefan Leutenegger, Andrew~J Davison, J{\"o}rg Conradt, Kostas Daniilidis, et~al.
\newblock Event-based vision: A survey.
\newblock \emph{IEEE transactions on pattern analysis and machine intelligence}, 44\penalty0 (1):\penalty0 154--180, 2020.

\bibitem[Gehrig and Scaramuzza(2022{\natexlab{a}})]{Gehrig2022PushingTL}
Daniel Gehrig and Davide Scaramuzza.
\newblock Pushing the limits of asynchronous graph-based object detection with event cameras.
\newblock \emph{ArXiv}, abs/2211.12324, 2022{\natexlab{a}}.

\bibitem[Gehrig and Scaramuzza(2024)]{Gehrig2024LowlatencyAV}
Daniel Gehrig and Davide Scaramuzza.
\newblock Low-latency automotive vision with event cameras.
\newblock \emph{Nature}, 629:\penalty0 1034 -- 1040, 2024.

\bibitem[Gehrig et~al.(2020)Gehrig, Gehrig, Hidalgo-Carri\'o, and Scaramuzza]{Gehrig_2020_CVPR}
Daniel Gehrig, Mathias Gehrig, Javier Hidalgo-Carri\'o, and Davide Scaramuzza.
\newblock Video to events: Recycling video datasets for event cameras.
\newblock In \emph{{IEEE} Conf. Comput. Vis. Pattern Recog. (CVPR)}, 2020.

\bibitem[Gehrig and Scaramuzza(2022{\natexlab{b}})]{Gehrig2022RecurrentVT}
Mathias Gehrig and Davide Scaramuzza.
\newblock Recurrent vision transformers for object detection with event cameras.
\newblock \emph{2023 IEEE/CVF Conference on Computer Vision and Pattern Recognition (CVPR)}, pages 13884--13893, 2022{\natexlab{b}}.

\bibitem[Gehrig et~al.(2021)Gehrig, Aarents, Gehrig, and Scaramuzza]{gehrig2021dsec}
Mathias Gehrig, Willem Aarents, Daniel Gehrig, and Davide Scaramuzza.
\newblock Dsec: A stereo event camera dataset for driving scenarios.
\newblock \emph{IEEE Robotics and Automation Letters}, 6\penalty0 (3):\penalty0 4947--4954, 2021.

\bibitem[Gehrig et~al.(2024)Gehrig, Muglikar, and Scaramuzza]{gehrig2024dense}
Mathias Gehrig, Manasi Muglikar, and Davide Scaramuzza.
\newblock Dense continuous-time optical flow from event cameras.
\newblock \emph{IEEE Transactions on Pattern Analysis and Machine Intelligence}, 2024.

\bibitem[Geiger et~al.(2012)Geiger, Lenz, and Urtasun]{Geiger2012AreWR}
Andreas Geiger, Philip Lenz, and Raquel Urtasun.
\newblock Are we ready for autonomous driving? the kitti vision benchmark suite.
\newblock \emph{2012 IEEE Conference on Computer Vision and Pattern Recognition}, pages 3354--3361, 2012.

\bibitem[Guan et~al.(2022)Guan, Wang, Lan, Chandra, Wu, Davis, and Manocha]{guan2022m3detr}
Tianrui Guan, Jun Wang, Shiyi Lan, Rohan Chandra, Zuxuan Wu, Larry Davis, and Dinesh Manocha.
\newblock M3detr: Multi-representation, multi-scale, mutual-relation 3d object detection with transformers.
\newblock In \emph{Proceedings of the IEEE/CVF winter conference on applications of computer vision}, pages 772--782, 2022.

\bibitem[Guo et~al.(2021)Guo, Shi, Wang, and Li]{guo2021liga}
Xiaoyang Guo, Shaoshuai Shi, Xiaogang Wang, and Hongsheng Li.
\newblock Liga-stereo: Learning lidar geometry aware representations for stereo-based 3d detector.
\newblock In \emph{Proceedings of the IEEE/CVF international conference on computer vision}, pages 3153--3163, 2021.

\bibitem[Hamaguchi et~al.(2023)Hamaguchi, Furukawa, Onishi, and Sakurada]{Hamaguchi2023HierarchicalNM}
Ryuhei Hamaguchi, Yasutaka Furukawa, Masaki Onishi, and Ken Sakurada.
\newblock Hierarchical neural memory network for low latency event processing.
\newblock \emph{2023 IEEE/CVF Conference on Computer Vision and Pattern Recognition (CVPR)}, pages 22867--22876, 2023.

\bibitem[He et~al.(2020)He, Zeng, Huang, Hua, and Zhang]{he2020structure}
Chenhang He, Hui Zeng, Jianqiang Huang, Xian-Sheng Hua, and Lei Zhang.
\newblock Structure aware single-stage 3d object detection from point cloud.
\newblock In \emph{Proceedings of the IEEE/CVF conference on computer vision and pattern recognition}, pages 11873--11882, 2020.

\bibitem[He et~al.(2022)He, Li, Li, and Zhang]{he2022voxel}
Chenhang He, Ruihuang Li, Shuai Li, and Lei Zhang.
\newblock Voxel set transformer: A set-to-set approach to 3d object detection from point clouds.
\newblock In \emph{Proceedings of the IEEE/CVF conference on computer vision and pattern recognition}, pages 8417--8427, 2022.

\bibitem[He and Soatto(2019)]{he2019mono3d++}
Tong He and Stefano Soatto.
\newblock Mono3d++: Monocular 3d vehicle detection with two-scale 3d hypotheses and task priors.
\newblock In \emph{Proceedings of the AAAI Conference on Artificial Intelligence}, pages 8409--8416, 2019.

\bibitem[Ho et~al.(2023)Ho, Tai, Lin, Yang, and Tsai]{ho2023diffusion}
Cheng-Ju Ho, Chen-Hsuan Tai, Yen-Yu Lin, Ming-Hsuan Yang, and Yi-Hsuan Tsai.
\newblock Diffusion-ss3d: Diffusion model for semi-supervised 3d object detection.
\newblock \emph{Advances in Neural Information Processing Systems}, 36:\penalty0 49100--49112, 2023.

\bibitem[Hu et~al.(2022)Hu, Kuai, and Waslander]{hu2022point}
Jordan~SK Hu, Tianshu Kuai, and Steven~L Waslander.
\newblock Point density-aware voxels for lidar 3d object detection.
\newblock In \emph{Proceedings of the IEEE/CVF conference on computer vision and pattern recognition}, pages 8469--8478, 2022.

\bibitem[Huang et~al.(2021{\natexlab{a}})Huang, Huang, Zhu, Ye, and Du]{huang2021bevdet}
Junjie Huang, Guan Huang, Zheng Zhu, Yun Ye, and Dalong Du.
\newblock Bevdet: High-performance multi-camera 3d object detection in bird-eye-view.
\newblock \emph{arXiv preprint arXiv:2112.11790}, 2021{\natexlab{a}}.

\bibitem[Huang et~al.(2025)Huang, Ye, Liang, Shan, and Du]{huang2025detecting}
Junjie Huang, Yun Ye, Zhujin Liang, Yi Shan, and Dalong Du.
\newblock Detecting as labeling: Rethinking lidar-camera fusion in 3d object detection.
\newblock In \emph{European Conference on Computer Vision}, pages 439--455. Springer, 2025.

\bibitem[Huang et~al.(2022)Huang, Wu, Su, and Hsu]{huang2022monodtr}
Kuan-Chih Huang, Tsung-Han Wu, Hung-Ting Su, and Winston~H Hsu.
\newblock Monodtr: Monocular 3d object detection with depth-aware transformer.
\newblock In \emph{Proceedings of the IEEE/CVF conference on computer vision and pattern recognition}, pages 4012--4021, 2022.

\bibitem[Huang et~al.(2021{\natexlab{b}})Huang, Du, Xue, Chen, Zhao, and Huang]{huang2021makes}
Yu Huang, Chenzhuang Du, Zihui Xue, Xuanyao Chen, Hang Zhao, and Longbo Huang.
\newblock What makes multi-modal learning better than single (provably).
\newblock \emph{Advances in Neural Information Processing Systems}, 34:\penalty0 10944--10956, 2021{\natexlab{b}}.

\bibitem[Iacono et~al.(2018)Iacono, Weber, Glover, and Bartolozzi]{Iacono2018TowardsEO}
Massimiliano Iacono, Stefan Weber, Arren~J. Glover, and Chiara Bartolozzi.
\newblock Towards event-driven object detection with off-the-shelf deep learning.
\newblock \emph{2018 IEEE/RSJ International Conference on Intelligent Robots and Systems (IROS)}, pages 1--9, 2018.

\bibitem[Jiang et~al.(2023)Jiang, Zhang, Miao, Zhu, Gao, Hu, and Jiang]{jiang2023polarformer}
Yanqin Jiang, Li Zhang, Zhenwei Miao, Xiatian Zhu, Jin Gao, Weiming Hu, and Yu-Gang Jiang.
\newblock Polarformer: Multi-camera 3d object detection with polar transformer.
\newblock In \emph{Proceedings of the AAAI conference on Artificial Intelligence}, pages 1042--1050, 2023.

\bibitem[Kim et~al.(2023)Kim, Chae, Jang, and Yoon]{kim2023event}
Taewoo Kim, Yujeong Chae, Hyun-Kurl Jang, and Kuk-Jin Yoon.
\newblock Event-based video frame interpolation with cross-modal asymmetric bidirectional motion fields.
\newblock In \emph{Proceedings of the IEEE/CVF Conference on Computer Vision and Pattern Recognition}, pages 18032--18042, 2023.

\bibitem[Kim et~al.(2024)Kim, Cho, and Yoon]{kim2024cross}
Taewoo Kim, Hoonhee Cho, and Kuk-Jin Yoon.
\newblock Cross-modal temporal alignment for event-guided video deblurring.
\newblock \emph{arXiv preprint arXiv:2408.14930}, 2024.

\bibitem[Kingma(2014)]{kingma2014adam}
Diederik~P Kingma.
\newblock Adam: A method for stochastic optimization.
\newblock \emph{arXiv preprint arXiv:1412.6980}, 2014.

\bibitem[Ku et~al.(2018)Ku, Mozifian, Lee, Harakeh, and Waslander]{ku2018joint}
Jason Ku, Melissa Mozifian, Jungwook Lee, Ali Harakeh, and Steven~L Waslander.
\newblock Joint 3d proposal generation and object detection from view aggregation.
\newblock In \emph{2018 IEEE/RSJ International Conference on Intelligent Robots and Systems (IROS)}, pages 1--8. IEEE, 2018.

\bibitem[Lang et~al.(2019)Lang, Vora, Caesar, Zhou, Yang, and Beijbom]{lang2019pointpillars}
Alex~H Lang, Sourabh Vora, Holger Caesar, Lubing Zhou, Jiong Yang, and Oscar Beijbom.
\newblock Pointpillars: Fast encoders for object detection from point clouds.
\newblock In \emph{Proceedings of the IEEE/CVF conference on computer vision and pattern recognition}, pages 12697--12705, 2019.

\bibitem[Li et~al.(2023{\natexlab{a}})Li, Li, and Tian]{Li2023SODFormerSO}
Dianze Li, Jianing Li, and Yonghong Tian.
\newblock Sodformer: Streaming object detection with transformer using events and frames.
\newblock \emph{IEEE Transactions on Pattern Analysis and Machine Intelligence}, 45:\penalty0 14020--14037, 2023{\natexlab{a}}.

\bibitem[Li et~al.(2022{\natexlab{a}})Li, Li, Zhu, Xiang, Huang, and Tian]{Li2022AsynchronousSM}
Jianing Li, Jia Li, Lin Zhu, Xijie Xiang, Tiejun Huang, and Yonghong Tian.
\newblock Asynchronous spatio-temporal memory network for continuous event-based object detection.
\newblock \emph{IEEE Transactions on Image Processing}, 31:\penalty0 2975--2987, 2022{\natexlab{a}}.

\bibitem[Li et~al.(2019)Li, Chen, and Shen]{li2019stereo}
Peiliang Li, Xiaozhi Chen, and Shaojie Shen.
\newblock Stereo r-cnn based 3d object detection for autonomous driving.
\newblock In \emph{Proceedings of the IEEE/CVF Conference on Computer Vision and Pattern Recognition}, pages 7644--7652, 2019.

\bibitem[Li et~al.(2023{\natexlab{b}})Li, Ma, Hou, Shi, Yang, Liu, Wu, Chen, Li, Qiao, et~al.]{li2023logonet}
Xin Li, Tao Ma, Yuenan Hou, Botian Shi, Yuchen Yang, Youquan Liu, Xingjiao Wu, Qin Chen, Yikang Li, Yu Qiao, et~al.
\newblock Logonet: Towards accurate 3d object detection with local-to-global cross-modal fusion.
\newblock In \emph{Proceedings of the IEEE/CVF Conference on Computer Vision and Pattern Recognition}, pages 17524--17534, 2023{\natexlab{b}}.

\bibitem[Li et~al.(2024{\natexlab{a}})Li, Fan, Tian, and Fan]{li2024gafusion}
Xiaotian Li, Baojie Fan, Jiandong Tian, and Huijie Fan.
\newblock Gafusion: Adaptive fusing lidar and camera with multiple guidance for 3d object detection.
\newblock In \emph{Proceedings of the IEEE/CVF Conference on Computer Vision and Pattern Recognition}, pages 21209--21218, 2024{\natexlab{a}}.

\bibitem[Li et~al.(2024{\natexlab{b}})Li, Yin, Li, Xu, Yang, and Shen]{li2024di}
Xiang Li, Junbo Yin, Wei Li, Chengzhong Xu, Ruigang Yang, and Jianbing Shen.
\newblock Di-v2x: Learning domain-invariant representation for vehicle-infrastructure collaborative 3d object detection.
\newblock In \emph{Proceedings of the AAAI Conference on Artificial Intelligence}, pages 3208--3215, 2024{\natexlab{b}}.

\bibitem[Li et~al.(2021{\natexlab{a}})Li, Zhou, Yang, Zhang, Cui, Bao, and Zhang]{Li2021GraphbasedAE}
Yijin Li, Han Zhou, Bangbang Yang, Yexin Zhang, Zhaopeng Cui, Hujun Bao, and Guofeng Zhang.
\newblock Graph-based asynchronous event processing for rapid object recognition.
\newblock \emph{2021 IEEE/CVF International Conference on Computer Vision (ICCV)}, pages 914--923, 2021{\natexlab{a}}.

\bibitem[Li et~al.(2022{\natexlab{b}})Li, Chen, Qi, Li, Sun, and Jia]{li2022unifying}
Yanwei Li, Yilun Chen, Xiaojuan Qi, Zeming Li, Jian Sun, and Jiaya Jia.
\newblock Unifying voxel-based representation with transformer for 3d object detection.
\newblock \emph{Advances in Neural Information Processing Systems}, 35:\penalty0 18442--18455, 2022{\natexlab{b}}.

\bibitem[Li et~al.(2022{\natexlab{c}})Li, Qi, Chen, Wang, Li, Sun, and Jia]{li2022voxel}
Yanwei Li, Xiaojuan Qi, Yukang Chen, Liwei Wang, Zeming Li, Jian Sun, and Jiaya Jia.
\newblock Voxel field fusion for 3d object detection.
\newblock In \emph{Proceedings of the IEEE/CVF Conference on Computer Vision and Pattern Recognition}, pages 1120--1129, 2022{\natexlab{c}}.

\bibitem[Li et~al.(2022{\natexlab{d}})Li, Yu, Meng, Caine, Ngiam, Peng, Shen, Lu, Zhou, Le, et~al.]{li2022deepfusion}
Yingwei Li, Adams~Wei Yu, Tianjian Meng, Ben Caine, Jiquan Ngiam, Daiyi Peng, Junyang Shen, Yifeng Lu, Denny Zhou, Quoc~V Le, et~al.
\newblock Deepfusion: Lidar-camera deep fusion for multi-modal 3d object detection.
\newblock In \emph{Proceedings of the IEEE/CVF conference on computer vision and pattern recognition}, pages 17182--17191, 2022{\natexlab{d}}.

\bibitem[Li et~al.(2023{\natexlab{c}})Li, Ge, Yu, Yang, Wang, Shi, Sun, and Li]{li2023bevdepth}
Yinhao Li, Zheng Ge, Guanyi Yu, Jinrong Yang, Zengran Wang, Yukang Shi, Jianjian Sun, and Zeming Li.
\newblock Bevdepth: Acquisition of reliable depth for multi-view 3d object detection.
\newblock In \emph{Proceedings of the AAAI Conference on Artificial Intelligence}, pages 1477--1485, 2023{\natexlab{c}}.

\bibitem[Li et~al.(2021{\natexlab{b}})Li, Wang, and Wang]{li2021lidar}
Zhichao Li, Feng Wang, and Naiyan Wang.
\newblock Lidar r-cnn: An efficient and universal 3d object detector.
\newblock In \emph{Proceedings of the IEEE/CVF conference on computer vision and pattern recognition}, pages 7546--7555, 2021{\natexlab{b}}.

\bibitem[Liang et~al.(2018)Liang, Yang, Wang, and Urtasun]{liang2018deep}
Ming Liang, Bin Yang, Shenlong Wang, and Raquel Urtasun.
\newblock Deep continuous fusion for multi-sensor 3d object detection.
\newblock In \emph{Proceedings of the European conference on computer vision (ECCV)}, pages 641--656, 2018.

\bibitem[Liang et~al.(2019)Liang, Yang, Chen, Hu, and Urtasun]{liang2019multi}
Ming Liang, Bin Yang, Yun Chen, Rui Hu, and Raquel Urtasun.
\newblock Multi-task multi-sensor fusion for 3d object detection.
\newblock In \emph{Proceedings of the IEEE/CVF conference on computer vision and pattern recognition}, pages 7345--7353, 2019.

\bibitem[Liang et~al.(2022)Liang, Xie, Yu, Xia, Lin, Wang, Tang, Wang, and Tang]{liang2022bevfusion}
Tingting Liang, Hongwei Xie, Kaicheng Yu, Zhongyu Xia, Zhiwei Lin, Yongtao Wang, Tao Tang, Bing Wang, and Zhi Tang.
\newblock Bevfusion: A simple and robust lidar-camera fusion framework.
\newblock \emph{Advances in Neural Information Processing Systems}, 35:\penalty0 10421--10434, 2022.

\bibitem[Lin et~al.(2024)Lin, Liu, Xia, Wang, Wang, Qi, Dong, Dong, Zhang, and Zhu]{lin2024rcbevdet}
Zhiwei Lin, Zhe Liu, Zhongyu Xia, Xinhao Wang, Yongtao Wang, Shengxiang Qi, Yang Dong, Nan Dong, Le Zhang, and Ce Zhu.
\newblock Rcbevdet: Radar-camera fusion in bird's eye view for 3d object detection.
\newblock In \emph{Proceedings of the IEEE/CVF Conference on Computer Vision and Pattern Recognition}, pages 14928--14937, 2024.

\bibitem[Liu et~al.(2024)Liu, Zheng, Qian, Xue, Chen, Zhang, Li, and Wu]{Liu_2024_CVPR}
Xianpeng Liu, Ce Zheng, Ming Qian, Nan Xue, Chen Chen, Zhebin Zhang, Chen Li, and Tianfu Wu.
\newblock Multi-view attentive contextualization for multi-view 3d object detection.
\newblock In \emph{Proceedings of the IEEE/CVF Conference on Computer Vision and Pattern Recognition (CVPR)}, pages 16688--16698, 2024.

\bibitem[Liu et~al.(2022)Liu, Wang, Zhang, and Sun]{liu2022petr}
Yingfei Liu, Tiancai Wang, Xiangyu Zhang, and Jian Sun.
\newblock Petr: Position embedding transformation for multi-view 3d object detection.
\newblock In \emph{European Conference on Computer Vision}, pages 531--548. Springer, 2022.

\bibitem[Liu et~al.(2020)Liu, Wu, and T{\'o}th]{liu2020smoke}
Zechen Liu, Zizhang Wu, and Roland T{\'o}th.
\newblock Smoke: Single-stage monocular 3d object detection via keypoint estimation.
\newblock In \emph{Proceedings of the IEEE/CVF conference on computer vision and pattern recognition workshops}, pages 996--997, 2020.

\bibitem[Liu et~al.(2021)Liu, Lin, Cao, Hu, Wei, Zhang, Lin, and Guo]{Liu2021SwinTH}
Ze Liu, Yutong Lin, Yue Cao, Han Hu, Yixuan Wei, Zheng Zhang, Stephen Lin, and Baining Guo.
\newblock Swin transformer: Hierarchical vision transformer using shifted windows.
\newblock \emph{2021 IEEE/CVF International Conference on Computer Vision (ICCV)}, pages 9992--10002, 2021.

\bibitem[Liu et~al.(2023)Liu, Tang, Amini, Yang, Mao, Rus, and Han]{liu2023bevfusion}
Zhijian Liu, Haotian Tang, Alexander Amini, Xinyu Yang, Huizi Mao, Daniela~L Rus, and Song Han.
\newblock Bevfusion: Multi-task multi-sensor fusion with unified bird's-eye view representation.
\newblock In \emph{2023 IEEE international conference on robotics and automation (ICRA)}, pages 2774--2781. IEEE, 2023.

\bibitem[Lu et~al.(2021)Lu, Ma, Yang, Zhang, Liu, Chu, Yan, and Ouyang]{lu2021geometry}
Yan Lu, Xinzhu Ma, Lei Yang, Tianzhu Zhang, Yating Liu, Qi Chu, Junjie Yan, and Wanli Ouyang.
\newblock Geometry uncertainty projection network for monocular 3d object detection.
\newblock In \emph{Proceedings of the IEEE/CVF International Conference on Computer Vision}, pages 3111--3121, 2021.

\bibitem[Mao et~al.(2021{\natexlab{a}})Mao, Niu, Bai, Liang, Xu, and Xu]{mao2021pyramid}
Jiageng Mao, Minzhe Niu, Haoyue Bai, Xiaodan Liang, Hang Xu, and Chunjing Xu.
\newblock Pyramid r-cnn: Towards better performance and adaptability for 3d object detection.
\newblock In \emph{Proceedings of the IEEE/CVF International Conference on Computer Vision}, pages 2723--2732, 2021{\natexlab{a}}.

\bibitem[Mao et~al.(2021{\natexlab{b}})Mao, Xue, Niu, Bai, Feng, Liang, Xu, and Xu]{mao2021voxel}
Jiageng Mao, Yujing Xue, Minzhe Niu, Haoyue Bai, Jiashi Feng, Xiaodan Liang, Hang Xu, and Chunjing Xu.
\newblock Voxel transformer for 3d object detection.
\newblock In \emph{Proceedings of the IEEE/CVF international conference on computer vision}, pages 3164--3173, 2021{\natexlab{b}}.

\bibitem[Messikommer et~al.(2020)Messikommer, Gehrig, Loquercio, and Scaramuzza]{Messikommer2020EventbasedAS}
Nico Messikommer, Daniel Gehrig, Antonio Loquercio, and Davide Scaramuzza.
\newblock Event-based asynchronous sparse convolutional networks.
\newblock \emph{ArXiv}, abs/2003.09148, 2020.

\bibitem[Miao et~al.(2021)Miao, Chen, Pan, Zhang, Liu, Hao, Zhu, Wang, and Zhan]{miao2021pvgnet}
Zhenwei Miao, Jikai Chen, Hongyu Pan, Ruiwen Zhang, Kaixuan Liu, Peihan Hao, Jun Zhu, Yang Wang, and Xin Zhan.
\newblock Pvgnet: A bottom-up one-stage 3d object detector with integrated multi-level features.
\newblock In \emph{Proceedings of the IEEE/CVF Conference on Computer Vision and Pattern Recognition}, pages 3279--3288, 2021.

\bibitem[Pang et~al.(2020)Pang, Morris, and Radha]{pang2020clocs}
Su Pang, Daniel Morris, and Hayder Radha.
\newblock Clocs: Camera-lidar object candidates fusion for 3d object detection.
\newblock In \emph{2020 IEEE/RSJ International Conference on Intelligent Robots and Systems (IROS)}, pages 10386--10393. IEEE, 2020.

\bibitem[Park et~al.(2021)Park, Ambrus, Guizilini, Li, and Gaidon]{park2021pseudo}
Dennis Park, Rares Ambrus, Vitor Guizilini, Jie Li, and Adrien Gaidon.
\newblock Is pseudo-lidar needed for monocular 3d object detection?
\newblock In \emph{Proceedings of the IEEE/CVF International Conference on Computer Vision}, pages 3142--3152, 2021.

\bibitem[Peng et~al.(2020)Peng, Pan, Liu, and Sun]{peng2020ida}
Wanli Peng, Hao Pan, He Liu, and Yi Sun.
\newblock Ida-3d: Instance-depth-aware 3d object detection from stereo vision for autonomous driving.
\newblock In \emph{Proceedings of the IEEE/CVF conference on computer vision and pattern recognition}, pages 13015--13024, 2020.

\bibitem[Peng et~al.(2023{\natexlab{a}})Peng, Zhang, Xiao, Sun, and Wu]{Peng2023BetterAF}
Yan Peng, Yueyi Zhang, Peilin Xiao, Xiaoyan Sun, and Feng Wu.
\newblock Better and faster: Adaptive event conversion for event-based object detection.
\newblock In \emph{AAAI Conference on Artificial Intelligence}, 2023{\natexlab{a}}.

\bibitem[Peng et~al.(2023{\natexlab{b}})Peng, Zhang, Xiong, Sun, and Wu]{Peng2023GETGE}
Yansong Peng, Yueyi Zhang, Zhiwei Xiong, Xiaoyan Sun, and Feng Wu.
\newblock Get: Group event transformer for event-based vision.
\newblock \emph{2023 IEEE/CVF International Conference on Computer Vision (ICCV)}, pages 6015--6025, 2023{\natexlab{b}}.

\bibitem[Peng et~al.(2024)Peng, Li, Zhang, Sun, and Wu]{Peng2024SceneAS}
Yansong Peng, Hebei Li, Yueyi Zhang, Xiaoyan Sun, and Feng Wu.
\newblock Scene adaptive sparse transformer for event-based object detection.
\newblock \emph{2024 IEEE/CVF Conference on Computer Vision and Pattern Recognition (CVPR)}, pages 16794--16804, 2024.

\bibitem[Perot et~al.(2020)Perot, de~Tournemire, Nitti, Masci, and Sironi]{Perot2020LearningTD}
Etienne Perot, Pierre de Tournemire, Davide~Oscar Nitti, Jonathan Masci, and Amos Sironi.
\newblock Learning to detect objects with a 1 megapixel event camera.
\newblock \emph{ArXiv}, abs/2009.13436, 2020.

\bibitem[Philion and Fidler(2020)]{philion2020lift}
Jonah Philion and Sanja Fidler.
\newblock Lift, splat, shoot: Encoding images from arbitrary camera rigs by implicitly unprojecting to 3d.
\newblock In \emph{Computer Vision--ECCV 2020: 16th European Conference, Glasgow, UK, August 23--28, 2020, Proceedings, Part XIV 16}, pages 194--210. Springer, 2020.

\bibitem[Piergiovanni et~al.(2021)Piergiovanni, Casser, Ryoo, and Angelova]{piergiovanni20214d}
AJ Piergiovanni, Vincent Casser, Michael~S Ryoo, and Anelia Angelova.
\newblock 4d-net for learned multi-modal alignment.
\newblock In \emph{Proceedings of the IEEE/CVF International Conference on Computer Vision}, pages 15435--15445, 2021.

\bibitem[Prakash et~al.(2021)Prakash, Chitta, and Geiger]{prakash2021multi}
Aditya Prakash, Kashyap Chitta, and Andreas Geiger.
\newblock Multi-modal fusion transformer for end-to-end autonomous driving.
\newblock In \emph{Proceedings of the IEEE/CVF conference on computer vision and pattern recognition}, pages 7077--7087, 2021.

\bibitem[Qi et~al.(2017{\natexlab{a}})Qi, Su, Mo, and Guibas]{qi2017pointnet}
Charles~R Qi, Hao Su, Kaichun Mo, and Leonidas~J Guibas.
\newblock Pointnet: Deep learning on point sets for 3d classification and segmentation.
\newblock In \emph{Proceedings of the IEEE conference on computer vision and pattern recognition}, pages 652--660, 2017{\natexlab{a}}.

\bibitem[Qi et~al.(2017{\natexlab{b}})Qi, Yi, Su, and Guibas]{qi2017pointnet++}
Charles~Ruizhongtai Qi, Li Yi, Hao Su, and Leonidas~J Guibas.
\newblock Pointnet++: Deep hierarchical feature learning on point sets in a metric space.
\newblock \emph{Advances in neural information processing systems}, 30, 2017{\natexlab{b}}.

\bibitem[Qi et~al.(2018)Qi, Liu, Wu, Su, and Guibas]{qi2018frustum}
Charles~R Qi, Wei Liu, Chenxia Wu, Hao Su, and Leonidas~J Guibas.
\newblock Frustum pointnets for 3d object detection from rgb-d data.
\newblock In \emph{Proceedings of the IEEE conference on computer vision and pattern recognition}, pages 918--927, 2018.

\bibitem[Qi et~al.(2019)Qi, Litany, He, and Guibas]{qi2019deep}
Charles~R Qi, Or Litany, Kaiming He, and Leonidas~J Guibas.
\newblock Deep hough voting for 3d object detection in point clouds.
\newblock In \emph{proceedings of the IEEE/CVF International Conference on Computer Vision}, pages 9277--9286, 2019.

\bibitem[Reading et~al.(2021)Reading, Harakeh, Chae, and Waslander]{reading2021categorical}
Cody Reading, Ali Harakeh, Julia Chae, and Steven~L Waslander.
\newblock Categorical depth distribution network for monocular 3d object detection.
\newblock In \emph{Proceedings of the IEEE/CVF Conference on Computer Vision and Pattern Recognition}, pages 8555--8564, 2021.

\bibitem[Schaefer et~al.(2022)Schaefer, Gehrig, and Scaramuzza]{Schaefer22cvpr}
Simon Schaefer, Daniel Gehrig, and Davide Scaramuzza.
\newblock Aegnn: Asynchronous event-based graph neural networks.
\newblock In \emph{IEEE Conference on Computer Vision and Pattern Recognition}, 2022.

\bibitem[Sheng et~al.(2021)Sheng, Cai, Liu, Deng, Huang, Hua, and Zhao]{sheng2021improving}
Hualian Sheng, Sijia Cai, Yuan Liu, Bing Deng, Jianqiang Huang, Xian-Sheng Hua, and Min-Jian Zhao.
\newblock Improving 3d object detection with channel-wise transformer.
\newblock In \emph{Proceedings of the IEEE/CVF international conference on computer vision}, pages 2743--2752, 2021.

\bibitem[Shi et~al.(2019)Shi, Wang, and Li]{shi2019pointrcnn}
Shaoshuai Shi, Xiaogang Wang, and Hongsheng Li.
\newblock Pointrcnn: 3d object proposal generation and detection from point cloud.
\newblock In \emph{Proceedings of the IEEE/CVF conference on computer vision and pattern recognition}, pages 770--779, 2019.

\bibitem[Shi et~al.(2020)Shi, Guo, Jiang, Wang, Shi, Wang, and Li]{shi2020pv}
Shaoshuai Shi, Chaoxu Guo, Li Jiang, Zhe Wang, Jianping Shi, Xiaogang Wang, and Hongsheng Li.
\newblock Pv-rcnn: Point-voxel feature set abstraction for 3d object detection.
\newblock In \emph{Proceedings of the IEEE/CVF conference on computer vision and pattern recognition}, pages 10529--10538, 2020.

\bibitem[Shi and Rajkumar(2020)]{shi2020point}
Weijing Shi and Raj Rajkumar.
\newblock Point-gnn: Graph neural network for 3d object detection in a point cloud.
\newblock In \emph{Proceedings of the IEEE/CVF conference on computer vision and pattern recognition}, pages 1711--1719, 2020.

\bibitem[Shiba et~al.(2022)Shiba, Aoki, and Gallego]{shiba2022secrets}
Shintaro Shiba, Yoshimitsu Aoki, and Guillermo Gallego.
\newblock Secrets of event-based optical flow.
\newblock In \emph{European Conference on Computer Vision}, pages 628--645. Springer, 2022.

\bibitem[Shoemake(1985)]{Shoemake1985AnimatingRW}
Ken Shoemake.
\newblock Animating rotation with quaternion curves.
\newblock \emph{Proceedings of the 12th annual conference on Computer graphics and interactive techniques}, 1985.

\bibitem[Smith and Topin(2019)]{smith2019super}
Leslie~N Smith and Nicholay Topin.
\newblock Super-convergence: Very fast training of neural networks using large learning rates.
\newblock In \emph{Artificial intelligence and machine learning for multi-domain operations applications}, pages 369--386. SPIE, 2019.

\bibitem[Song et~al.(2024)Song, Yang, Xu, Liu, Xu, Jia, Jia, and Wang]{song2024graphbev}
Ziying Song, Lei Yang, Shaoqing Xu, Lin Liu, Dongyang Xu, Caiyan Jia, Feiyang Jia, and Li Wang.
\newblock Graphbev: Towards robust bev feature alignment for multi-modal 3d object detection.
\newblock \emph{arXiv preprint arXiv:2403.11848}, 2024.

\bibitem[Sun et~al.(2020)Sun, Chen, Xie, Zhang, Jiang, Zhou, and Bao]{sun2020disp}
Jiaming Sun, Linghao Chen, Yiming Xie, Siyu Zhang, Qinhong Jiang, Xiaowei Zhou, and Hujun Bao.
\newblock Disp r-cnn: Stereo 3d object detection via shape prior guided instance disparity estimation.
\newblock In \emph{Proceedings of the IEEE/CVF conference on computer vision and pattern recognition}, pages 10548--10557, 2020.

\bibitem[Sun et~al.(2019)Sun, Kretzschmar, Dotiwalla, Chouard, Patnaik, Tsui, Guo, Zhou, Chai, Caine, Vasudevan, Han, Ngiam, Zhao, Timofeev, Ettinger, Krivokon, Gao, Joshi, Zhang, Shlens, Chen, and Anguelov]{Sun2019ScalabilityIP}
Pei Sun, Henrik Kretzschmar, Xerxes Dotiwalla, Aurelien Chouard, Vijaysai Patnaik, Paul Tsui, James Guo, Yin Zhou, Yuning Chai, Benjamin Caine, Vijay Vasudevan, Wei Han, Jiquan Ngiam, Hang Zhao, Aleksei Timofeev, Scott~M. Ettinger, Maxim Krivokon, Amy Gao, Aditya Joshi, Yu Zhang, Jonathon Shlens, Zhifeng Chen, and Dragomir Anguelov.
\newblock Scalability in perception for autonomous driving: Waymo open dataset.
\newblock \emph{2020 IEEE/CVF Conference on Computer Vision and Pattern Recognition (CVPR)}, pages 2443--2451, 2019.

\bibitem[Tulyakov et~al.(2022)Tulyakov, Bochicchio, Gehrig, Georgoulis, Li, and Scaramuzza]{tulyakov2022time}
Stepan Tulyakov, Alfredo Bochicchio, Daniel Gehrig, Stamatios Georgoulis, Yuanyou Li, and Davide Scaramuzza.
\newblock Time lens++: Event-based frame interpolation with parametric non-linear flow and multi-scale fusion.
\newblock In \emph{Proceedings of the IEEE/CVF Conference on Computer Vision and Pattern Recognition}, pages 17755--17764, 2022.

\bibitem[Vora et~al.(2020)Vora, Lang, Helou, and Beijbom]{vora2020pointpainting}
Sourabh Vora, Alex~H Lang, Bassam Helou, and Oscar Beijbom.
\newblock Pointpainting: Sequential fusion for 3d object detection.
\newblock In \emph{Proceedings of the IEEE/CVF conference on computer vision and pattern recognition}, pages 4604--4612, 2020.

\bibitem[Wan et~al.(2023)Wan, Mao, Zhang, and Dai]{wan2023rpeflow}
Zhexiong Wan, Yuxin Mao, Jing Zhang, and Yuchao Dai.
\newblock Rpeflow: Multimodal fusion of rgb-pointcloud-event for joint optical flow and scene flow estimation.
\newblock In \emph{Proceedings of the IEEE/CVF International Conference on Computer Vision}, pages 10030--10040, 2023.

\bibitem[Wang et~al.(2023{\natexlab{a}})Wang, Jia, Zhang, Zhang, Wang, Zhang, Wang, and Lu]{Wang2023DualMA}
Dongsheng Wang, Xu Jia, Yang Zhang, Xinyu Zhang, Yaoyuan Wang, Ziyang Zhang, D. Wang, and Huchuan Lu.
\newblock Dual memory aggregation network for event-based object detection with learnable representation.
\newblock In \emph{AAAI Conference on Artificial Intelligence}, 2023{\natexlab{a}}.

\bibitem[Wang et~al.(2024)Wang, Lu, Zheng, Zhan, Ye, Tan, Wang, Wang, and Li]{wang2024bevspread}
Wenjie Wang, Yehao Lu, Guangcong Zheng, Shuigen Zhan, Xiaoqing Ye, Zichang Tan, Jingdong Wang, Gaoang Wang, and Xi Li.
\newblock Bevspread: Spread voxel pooling for bird's-eye-view representation in vision-based roadside 3d object detection.
\newblock In \emph{Proceedings of the IEEE/CVF Conference on Computer Vision and Pattern Recognition}, pages 14718--14727, 2024.

\bibitem[Wang et~al.(2019)Wang, Chao, Garg, Hariharan, Campbell, and Weinberger]{wang2019pseudo}
Yan Wang, Wei-Lun Chao, Divyansh Garg, Bharath Hariharan, Mark Campbell, and Kilian~Q Weinberger.
\newblock Pseudo-lidar from visual depth estimation: Bridging the gap in 3d object detection for autonomous driving.
\newblock In \emph{Proceedings of the IEEE/CVF conference on computer vision and pattern recognition}, pages 8445--8453, 2019.

\bibitem[Wang et~al.(2022)Wang, Guizilini, Zhang, Wang, Zhao, and Solomon]{wang2022detr3d}
Yue Wang, Vitor~Campagnolo Guizilini, Tianyuan Zhang, Yilun Wang, Hang Zhao, and Justin Solomon.
\newblock Detr3d: 3d object detection from multi-view images via 3d-to-2d queries.
\newblock In \emph{Conference on Robot Learning}, pages 180--191. PMLR, 2022.

\bibitem[Wang et~al.(2023{\natexlab{b}})Wang, Yin, Li, Frossard, Yang, and Shen]{wang2023ssda3d}
Yan Wang, Junbo Yin, Wei Li, Pascal Frossard, Ruigang Yang, and Jianbing Shen.
\newblock Ssda3d: Semi-supervised domain adaptation for 3d object detection from point cloud.
\newblock In \emph{Proceedings of the AAAI Conference on Artificial Intelligence}, pages 2707--2715, 2023{\natexlab{b}}.

\bibitem[Xie et~al.(2023)Xie, Xu, Rakotosaona, Rim, Tombari, Keutzer, Tomizuka, and Zhan]{Xie2023SparseFusionFM}
Yichen Xie, Chenfeng Xu, Marie-Julie Rakotosaona, Patrick Rim, Federico Tombari, Kurt Keutzer, Masayoshi Tomizuka, and Wei Zhan.
\newblock Sparsefusion: Fusing multi-modal sparse representations for multi-sensor 3d object detection.
\newblock \emph{2023 IEEE/CVF International Conference on Computer Vision (ICCV)}, pages 17545--17556, 2023.

\bibitem[Xu et~al.(2018)Xu, Anguelov, and Jain]{xu2018pointfusion}
Danfei Xu, Dragomir Anguelov, and Ashesh Jain.
\newblock Pointfusion: Deep sensor fusion for 3d bounding box estimation.
\newblock In \emph{Proceedings of the IEEE conference on computer vision and pattern recognition}, pages 244--253, 2018.

\bibitem[Xu et~al.(2022)Xu, Miao, Zhang, Pan, Liu, Hao, Zhu, Sun, Li, and Zhan]{xu2022int}
Jianyun Xu, Zhenwei Miao, Da Zhang, Hongyu Pan, Kaixuan Liu, Peihan Hao, Jun Zhu, Zhengyang Sun, Hongmin Li, and Xin Zhan.
\newblock Int: Towards infinite-frames 3d detection with an efficient framework.
\newblock In \emph{European Conference on Computer Vision}, pages 193--209. Springer, 2022.

\bibitem[Xu et~al.(2021)Xu, Zhou, Fang, Yin, Bin, and Zhang]{xu2021fusionpainting}
Shaoqing Xu, Dingfu Zhou, Jin Fang, Junbo Yin, Zhou Bin, and Liangjun Zhang.
\newblock Fusionpainting: Multimodal fusion with adaptive attention for 3d object detection.
\newblock In \emph{2021 IEEE International Intelligent Transportation Systems Conference (ITSC)}, pages 3047--3054. IEEE, 2021.

\bibitem[Yan et~al.(2018)Yan, Mao, and Li]{Yan2018SECONDSE}
Yan Yan, Yuxing Mao, and Bo Li.
\newblock Second: Sparsely embedded convolutional detection.
\newblock \emph{Sensors (Basel, Switzerland)}, 18, 2018.

\bibitem[Yang et~al.(2024)Yang, Liang, Yu, Chen, Ren, and Shi]{yang2024latency}
Yixin Yang, Jinxiu Liang, Bohan Yu, Yan Chen, Jimmy~S Ren, and Boxin Shi.
\newblock Latency correction for event-guided deblurring and frame interpolation.
\newblock In \emph{Proceedings of the IEEE/CVF Conference on Computer Vision and Pattern Recognition}, pages 24977--24986, 2024.

\bibitem[Yang et~al.(2019)Yang, Sun, Liu, Shen, and Jia]{yang2019std}
Zetong Yang, Yanan Sun, Shu Liu, Xiaoyong Shen, and Jiaya Jia.
\newblock Std: Sparse-to-dense 3d object detector for point cloud.
\newblock In \emph{Proceedings of the IEEE/CVF international conference on computer vision}, pages 1951--1960, 2019.

\bibitem[Yang et~al.(2020)Yang, Sun, Liu, and Jia]{yang20203dssd}
Zetong Yang, Yanan Sun, Shu Liu, and Jiaya Jia.
\newblock 3dssd: Point-based 3d single stage object detector.
\newblock In \emph{Proceedings of the IEEE/CVF conference on computer vision and pattern recognition}, pages 11040--11048, 2020.

\bibitem[Yang et~al.(2022)Yang, Chen, Miao, Li, Zhu, and Zhang]{yang2022deepinteraction}
Zeyu Yang, Jiaqi Chen, Zhenwei Miao, Wei Li, Xiatian Zhu, and Li Zhang.
\newblock Deepinteraction: 3d object detection via modality interaction.
\newblock \emph{Advances in Neural Information Processing Systems}, 35:\penalty0 1992--2005, 2022.

\bibitem[Yao et~al.(2021)Yao, Gao, Zhao, Wang, Lin, Yang, and Li]{Yao2021TemporalwiseAS}
Man Yao, Huanhuan Gao, Guangshe Zhao, Dingheng Wang, Yihan Lin, Zhao-Xu Yang, and Guoqi Li.
\newblock Temporal-wise attention spiking neural networks for event streams classification.
\newblock \emph{2021 IEEE/CVF International Conference on Computer Vision (ICCV)}, pages 10201--10210, 2021.

\bibitem[Ye et~al.(2019)Ye, Chen, and Liu]{Ye2019TightlyC3}
Haoyang Ye, Yuying Chen, and Ming Liu.
\newblock Tightly coupled 3d lidar inertial odometry and mapping.
\newblock \emph{2019 International Conference on Robotics and Automation (ICRA)}, pages 3144--3150, 2019.

\bibitem[Yin et~al.(2022{\natexlab{a}})Yin, Fang, Zhou, Zhang, Xu, Shen, and Wang]{yin2022semi}
Junbo Yin, Jin Fang, Dingfu Zhou, Liangjun Zhang, Cheng-Zhong Xu, Jianbing Shen, and Wenguan Wang.
\newblock Semi-supervised 3d object detection with proficient teachers.
\newblock In \emph{European Conference on Computer Vision}, pages 727--743. Springer, 2022{\natexlab{a}}.

\bibitem[Yin et~al.(2022{\natexlab{b}})Yin, Zhou, Zhang, Fang, Xu, Shen, and Wang]{yin2022proposalcontrast}
Junbo Yin, Dingfu Zhou, Liangjun Zhang, Jin Fang, Cheng-Zhong Xu, Jianbing Shen, and Wenguan Wang.
\newblock Proposalcontrast: Unsupervised pre-training for lidar-based 3d object detection.
\newblock In \emph{European conference on computer vision}, pages 17--33. Springer, 2022{\natexlab{b}}.

\bibitem[Yin et~al.(2024{\natexlab{a}})Yin, Shen, Chen, Li, Yang, Frossard, and Wang]{Yin2024ISFusionIC}
Junbo Yin, Jianbing Shen, Runnan Chen, Wei Li, Ruigang Yang, Pascal Frossard, and Wenguan Wang.
\newblock Is-fusion: Instance-scene collaborative fusion for multimodal 3d object detection.
\newblock \emph{2024 IEEE/CVF Conference on Computer Vision and Pattern Recognition (CVPR)}, pages 14905--14915, 2024{\natexlab{a}}.

\bibitem[Yin et~al.(2024{\natexlab{b}})Yin, Shen, Chen, Li, Yang, Frossard, and Wang]{yin2024fusion}
Junbo Yin, Jianbing Shen, Runnan Chen, Wei Li, Ruigang Yang, Pascal Frossard, and Wenguan Wang.
\newblock Is-fusion: Instance-scene collaborative fusion for multimodal 3d object detection.
\newblock In \emph{Proceedings of the IEEE/CVF Conference on Computer Vision and Pattern Recognition}, pages 14905--14915, 2024{\natexlab{b}}.

\bibitem[Yin et~al.(2021)Yin, Zhou, and Krahenbuhl]{yin2021center}
Tianwei Yin, Xingyi Zhou, and Philipp Krahenbuhl.
\newblock Center-based 3d object detection and tracking.
\newblock In \emph{Proceedings of the IEEE/CVF conference on computer vision and pattern recognition}, pages 11784--11793, 2021.

\bibitem[You et~al.(2019)You, Wang, Chao, Garg, Pleiss, Hariharan, Campbell, and Weinberger]{you2019pseudo}
Yurong You, Yan Wang, Wei-Lun Chao, Divyansh Garg, Geoff Pleiss, Bharath Hariharan, Mark Campbell, and Kilian~Q Weinberger.
\newblock Pseudo-lidar++: Accurate depth for 3d object detection in autonomous driving.
\newblock \emph{arXiv preprint arXiv:1906.06310}, 2019.

\bibitem[Zhang et~al.(2023{\natexlab{a}})Zhang, Zhu, Wang, Chen, Wu, and Wang]{zhang2023extracting}
Guozhen Zhang, Yuhan Zhu, Haonan Wang, Youxin Chen, Gangshan Wu, and Limin Wang.
\newblock Extracting motion and appearance via inter-frame attention for efficient video frame interpolation.
\newblock In \emph{Proceedings of the IEEE/CVF Conference on Computer Vision and Pattern Recognition}, pages 5682--5692, 2023{\natexlab{a}}.

\bibitem[Zhang et~al.(2024{\natexlab{a}})Zhang, Chen, Gao, Li, Liu, and Hu]{Zhang_2024_CVPR}
Gang Zhang, Junnan Chen, Guohuan Gao, Jianmin Li, Si Liu, and Xiaolin Hu.
\newblock Safdnet: A simple and effective network for fully sparse 3d object detection.
\newblock In \emph{Proceedings of the IEEE/CVF Conference on Computer Vision and Pattern Recognition (CVPR)}, pages 14477--14486, 2024{\natexlab{a}}.

\bibitem[Zhang et~al.(2024{\natexlab{b}})Zhang, Junnan, Gao, Li, and Hu]{zhang2024hednet}
Gang Zhang, Chen Junnan, Guohuan Gao, Jianmin Li, and Xiaolin Hu.
\newblock Hednet: A hierarchical encoder-decoder network for 3d object detection in point clouds.
\newblock \emph{Advances in Neural Information Processing Systems}, 36, 2024{\natexlab{b}}.

\bibitem[Zhang et~al.(2022{\natexlab{a}})Zhang, Dong, Zhang, Ding, Heide, Yin, and Yang]{Zhang2022SpikingTF}
Jiqing Zhang, B. Dong, Haiwei Zhang, Jianchuan Ding, Felix Heide, Baocai Yin, and Xin Yang.
\newblock Spiking transformers for event-based single object tracking.
\newblock \emph{2022 IEEE/CVF Conference on Computer Vision and Pattern Recognition (CVPR)}, pages 8791--8800, 2022{\natexlab{a}}.

\bibitem[Zhang et~al.(2023{\natexlab{b}})Zhang, Qiu, Wang, Guo, Cui, Qiao, Li, and Gao]{zhang2023monodetr}
Renrui Zhang, Han Qiu, Tai Wang, Ziyu Guo, Ziteng Cui, Yu Qiao, Hongsheng Li, and Peng Gao.
\newblock Monodetr: Depth-guided transformer for monocular 3d object detection.
\newblock In \emph{Proceedings of the IEEE/CVF International Conference on Computer Vision}, pages 9155--9166, 2023{\natexlab{b}}.

\bibitem[Zhang et~al.(2022{\natexlab{b}})Zhang, Chen, and Huang]{zhang2022cat}
Yanan Zhang, Jiaxin Chen, and Di Huang.
\newblock Cat-det: Contrastively augmented transformer for multi-modal 3d object detection.
\newblock In \emph{Proceedings of the IEEE/CVF Conference on Computer Vision and Pattern Recognition}, pages 908--917, 2022{\natexlab{b}}.

\bibitem[Zheng et~al.(2023)Zheng, Wu, Lu, Lu, Chen, and Jiang]{zheng2023neuralpci}
Zehan Zheng, Danni Wu, Ruisi Lu, Fan Lu, Guang Chen, and Changjun Jiang.
\newblock Neuralpci: Spatio-temporal neural field for 3d point cloud multi-frame non-linear interpolation.
\newblock In \emph{Proceedings of the IEEE/CVF Conference on Computer Vision and Pattern Recognition}, pages 909--918, 2023.

\bibitem[Zhou et~al.(2020)Zhou, Fang, Song, Liu, Yin, Dai, Li, and Yang]{zhou2020joint}
Dingfu Zhou, Jin Fang, Xibin Song, Liu Liu, Junbo Yin, Yuchao Dai, Hongdong Li, and Ruigang Yang.
\newblock Joint 3d instance segmentation and object detection for autonomous driving.
\newblock In \emph{Proceedings of the IEEE/CVF Conference on Computer Vision and Pattern Recognition}, pages 1839--1849, 2020.

\bibitem[Zhou et~al.(2024)Zhou, Chang, and Shi]{zhou2024bring}
Hanyu Zhou, Yi Chang, and Zhiwei Shi.
\newblock Bring event into rgb and lidar: Hierarchical visual-motion fusion for scene flow.
\newblock In \emph{Proceedings of the IEEE/CVF Conference on Computer Vision and Pattern Recognition}, pages 26477--26486, 2024.

\bibitem[Zhou and Tuzel(2018)]{zhou2018voxelnet}
Yin Zhou and Oncel Tuzel.
\newblock Voxelnet: End-to-end learning for point cloud based 3d object detection.
\newblock In \emph{Proceedings of the IEEE conference on computer vision and pattern recognition}, pages 4490--4499, 2018.

\bibitem[Zhu et~al.(2019)Zhu, Yuan, Chaney, and Daniilidis]{zhu2019unsupervised}
Alex~Zihao Zhu, Liangzhe Yuan, Kenneth Chaney, and Kostas Daniilidis.
\newblock Unsupervised event-based learning of optical flow, depth, and egomotion.
\newblock In \emph{Proceedings of the IEEE/CVF Conference on Computer Vision and Pattern Recognition}, pages 989--997, 2019.

\bibitem[Zubic et~al.(2023)Zubic, Gehrig, Gehrig, and Scaramuzza]{Zubic2023FromCC}
Nikola Zubic, Daniel Gehrig, Mathias Gehrig, and Davide Scaramuzza.
\newblock From chaos comes order: Ordering event representations for object recognition and detection.
\newblock \emph{2023 IEEE/CVF International Conference on Computer Vision (ICCV)}, pages 12800--12810, 2023.

\end{thebibliography}
}


\end{document}